\documentclass[11pt]{article}
\usepackage[preprint]{acl}
\usepackage{times,latexsym}

\usepackage[utf8]{inputenc}
\usepackage[T1]{fontenc}
\usepackage{hyperref}
\usepackage{url}
\usepackage{booktabs}
\usepackage{amsfonts}
\usepackage{amssymb}
\usepackage{amsmath}
\usepackage{nicefrac}
\usepackage{microtype}
\usepackage{graphicx}
\usepackage{xcolor}
\usepackage{multirow}
\usepackage{makecell}
\usepackage{enumitem}
\usepackage{subcaption}
\usepackage{algorithm}
\usepackage{algorithmic}
\usepackage{fvextra}
\usepackage{etoolbox}
\usepackage{placeins}

\title{LADDER: Diagnosing Domain-Level Self-Assessment Under Noisy Feedback}

\author{
\parbox{\textwidth}{\centering
Zhenshuai Yin\textsuperscript{1,*}\quad
Jianshu Liang\textsuperscript{1,*}\quad
Xuyue Xiong\textsuperscript{1}\quad
Yongxin Guo\textsuperscript{3}\\[4pt]
Mingda Zhang\textsuperscript{1}\quad
Zhongze Luo\textsuperscript{1}\quad
Haoyue Liu\textsuperscript{1}\quad
Lianmin Zhang\textsuperscript{2}\\[4pt]
Xiaoying Tang\textsuperscript{1,$\dagger$}\\[6pt]
\normalfont\normalsize
\textsuperscript{1}School of Science and Engineering,\\
The Chinese University of Hong Kong (Shenzhen),\\
Longgang, Shenzhen, Guangdong, 518172, P.R. China\\[3pt]
\textsuperscript{2}Shenzhen Research Institute of Big Data, Shenzhen, China\\[3pt]
\textsuperscript{3}ByteDance\\[4pt]
\small\textsuperscript{*}Equal contribution.\quad
\textsuperscript{$\dagger$}Corresponding author: Xiaoying Tang.
}
}

\newsavebox{\laddertablebox}
\newcommand{\fitTable}[1]{%
  \sbox{\laddertablebox}{#1}%
  \ifdim\wd\laddertablebox>\linewidth
    \resizebox{\linewidth}{!}{\usebox{\laddertablebox}}%
  \else\usebox{\laddertablebox}\fi}

\hypersetup{pdfauthor={Zhenshuai Yin, Jianshu Liang, Xuyue Xiong, Yongxin Guo, Mingda Zhang, Zhongze Luo, Haoyue Liu, Lianmin Zhang, Xiaoying Tang},pdftitle={LADDER: Diagnosing Domain-Level Self-Assessment Under Noisy Feedback}}

\begin{document}
\raggedbottom
\setlength{\textfloatsep}{12pt}

\maketitle

\begin{abstract}
Knowing when they are reliable is essential for language models to support human--AI collaboration. Yet interaction feedback is not always complete or correct, and may mislead models' understanding of their own capabilities. We introduce \textbf{LADDER} to study how language models assess themselves under unreliable feedback. Our findings show that models can accurately report supplied performance information while still struggling to form trustworthy self-assessments from interaction experience. Apparently accurate assessments can obscure capability differences and do not necessarily predict future task performance. These findings suggest that trustworthy self-assessment should be evaluated against observed task performance and its predictive value. Our study provides empirical evidence of the limits of model self-knowledge and a basis for evaluating capability awareness in reliable human--AI collaboration.
\end{abstract}

\begin{figure}[t]
\centering
\includegraphics[width=\columnwidth]{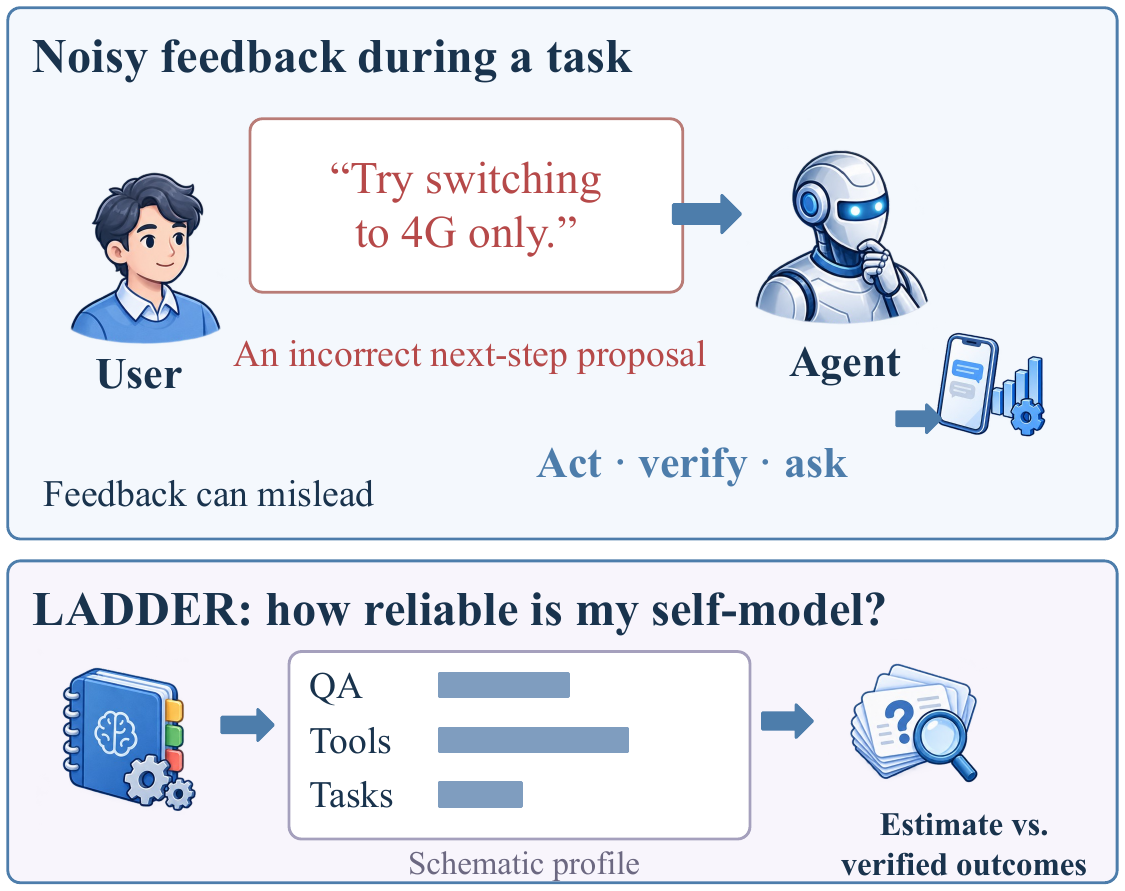}
\caption{\textbf{The LADDER research question.} The upper panel illustrates an application in which user feedback can be incorrect. The evaluated setting in the lower panel reconstructs a domain profile from QA records with simulated feedback; profile estimates are scored against recorded answer outcomes.}
\label{fig:problem}
\end{figure}

\section{Introduction}
\label{sec:intro}

Repeated use of language models exposes differences in their reliability across domains. A useful self-model would summarize these differences and provide evidence for deciding when additional verification is needed. Yet an interaction history is an imperfect source of such evidence: users may accept an incorrect answer, question a correct one, or provide no usable judgment. Before studying how a capability profile changes an agent's decisions, we need to establish whether that profile accurately summarizes its observed performance. We investigate this prerequisite through domain-level reconstruction from question-answering histories.

We organize the question into three diagnostic contrasts. H1 asks whether a model can report supplied numerical statistics. H2 measures the accuracy of its prior domain estimates. H3 asks how well observed feedback supports reconstruction of performance. These contrasts motivate controls with different information access, from no history to verified outcomes. Work on self-knowledge~\citep{barkan2026capable}, calibration~\citep{kadavath2022language,guo2017calibration,xiong2024can}, and memory~\citep{packer2023memgpt,mem02025,tan2025rmm,xu2025amem,li2025memos,hu2026memoryagent} motivates these questions. Our goal is to distinguish errors associated with the available information and its aggregation, rather than infer a unique internal cause from a benchmark score.

\begin{figure*}[t]
\centering
\includegraphics[width=\textwidth,trim=0 175bp 0 0,clip]{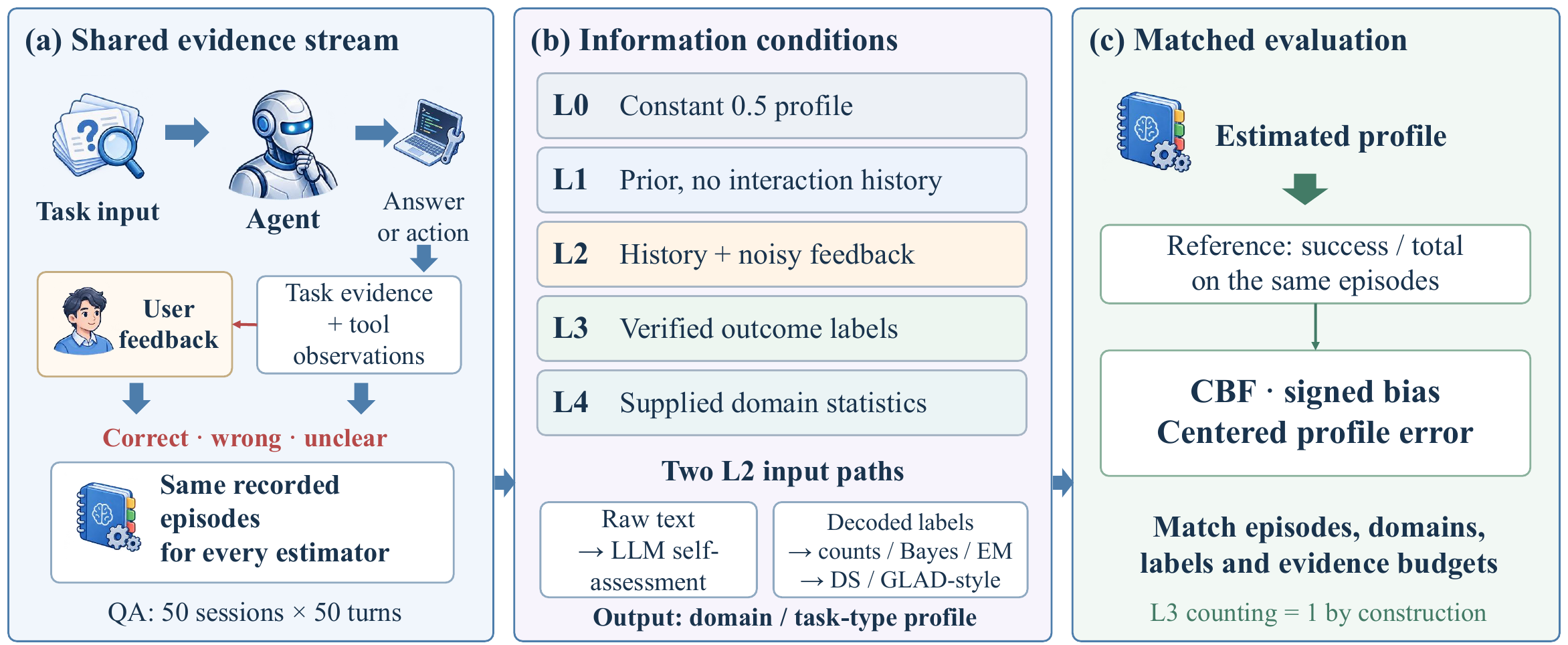}
\caption{\textbf{LADDER's diagnostic architecture.} Interaction records connect tasks, agent responses, and user feedback to a domain-level self-model. The L0--L4 ladder varies access to prior knowledge, noisy feedback, verified outcomes, and supplied statistics. Estimated profiles are evaluated against empirical domain performance.}
\label{fig:overview}
\end{figure*}

We study this question with LADDER. The benchmark is organized around an oracle-bracketed \emph{five-level baseline ladder}: L0 a constant no-memory profile, L1 a zero-shot per-domain prior, L2 noisy-feedback learners (LLM self-assessment, classical recalibration, EM-Joint, and Dawid--Skene/GLAD-style estimators), L3 exact counting of verified outcomes, and L4 model reporting of supplied statistics. L4--L3 examines numerical reporting (H1), L1--L3 measures prior error (H2), and L2--L3 measures reconstruction from feedback (H3). Each agent is paired with a domain-heterogeneous simulated user. Sessions last 50 turns, and \textbf{Capability Boundary Fidelity (CBF)} evaluates the resulting domain profile alongside per-query calibration measures such as ECE.

Across 18 models from seven families, the ladder reveals a distinction between recovering overall accuracy and learning differences across domains. A global feedback mean outperforms domain-specific Count, Beta, DS, and GLAD-style estimators on 14 of 18 models at the session endpoint. Prior self-assessments and LSA can improve on these references, but their advantage varies by model. A matched-input control on GPT-5.5, DeepSeek-V4-Pro, and Qwen2.5-7B shows that clean-label aggregation and reporting supplied statistics are also different tasks: LSA reaches CBF .989, .995, and .835, respectively, while all three reproduce the supplied statistics exactly. Figure~\ref{fig:overview} links these contrasts to the diagnostic ladder. Centered domain-shape error $e_{\rm shape}$ complements CBF by measuring domain-profile error after removing overall bias.

\paragraph{Contributions.} (i) A five-level diagnostic protocol for comparing prior knowledge, feedback-based estimation, and numerical reporting; (ii) domain-level CBF with baseline adjustment and complementary signed and centered errors; and (iii) an 18-model empirical study, with matched-input, noise, domain-partition, and cross-task controls that identify when feedback improves the estimated capability profile.

\section{Related Work}
\label{sec:related}

LADDER connects model self-knowledge with learning from interaction feedback. We organize related work around capability assessment, interactive agents, memory and feedback learning, and statistical estimation. Table~\ref{tab:related_axes} compares representative benchmarks by feedback channel, interaction horizon, accumulated self-knowledge, and user-reliability inference.

\paragraph{Model self-knowledge and calibration.} Self-knowledge research examines whether models can recognize and report the correctness of their own answers. Prior work examines answer confidence~\citep{kadavath2022language}, activation monitoring~\citep{jian2025metacognition}, and behavioral self-prediction~\citep{binder2025introspection}. Behavioral evaluations find context-dependent metacognition~\citep{ackerman2026metacognition}, while position work argues for its role in self-improvement~\citep{liu2025metacognitive_learning}. SAD~\citep{laine2024sad} and SelfAware~\citep{yin2023selfaware} assess self-knowledge without an accumulated feedback history, providing a reference for prior self-assessment.

Knowledge-boundary studies and task-performance benchmarks~\citep{li2025knowledge_boundary,yin2024benchmarking_boundary,wei2024simpleqa,hendrycks2021measuring,lin2022truthfulqa,liang2022helm,geng2024survey} characterize models' capabilities on scored questions. Selective prediction~\citep{kapoor2024taught,diao2024rtuning,traub2024augrc,li2025abstention,kirichenko2025abstentionbench,ahdritz2024knowable} and confidence calibration~\citep{xu2024sayself,shen2024thermometer,chen2024reconfidencing,manggala2025qacal} connect confidence to query-level judgments. LADDER extends these questions to capability profiles estimated from repeated interactions with imperfect feedback, evaluating whether reported domain performance agrees with recorded answer outcomes. Appendix~\ref{app:related_extended} provides further comparisons.

\begin{table*}[t]
\centering\fontsize{9}{10.8}\selectfont

\setlength{\tabcolsep}{4pt}
\fitTable{\begin{tabular}{@{}>{\raggedright\arraybackslash}p{0.3\textwidth}>{\raggedright\arraybackslash}p{0.22\textwidth}>{\raggedright\arraybackslash}p{0.12\textwidth}>{\raggedright\arraybackslash}p{0.18\textwidth}>{\raggedright\arraybackslash}p{0.08\textwidth}@{}}
\toprule
\textbf{Benchmark} & \textbf{Feedback} & \textbf{Horizon} & \textbf{Self-knowledge} & \textbf{2nd-order} \\
\midrule
SAD~\citep{laine2024sad}, SelfAware~\citep{yin2023selfaware} & none & single-shot & static probe & --- \\
Barkan~\citep{barkan2026capable} & execution traces & multi-step exec & per-task & --- \\
MemoryAgentBench~\citep{hu2026memoryagent} & implicit (mem ops) & long-horizon & exogenous mem only & --- \\
AgentBoard~\citep{ma2024agentboard}, $\tau$-bench~\citep{yao2024taubench} & progress / task reward & multi-turn & none & --- \\
SimpleQA~\citep{wei2024simpleqa} & none & single-shot & static & --- \\
\midrule
\textbf{LADDER (ours)} & \textbf{noisy domain-dependent NL} & \textbf{50-turn session} & \textbf{per-domain accumulated} & \textbf{\checkmark} \\
\bottomrule
\end{tabular}}
\caption{\textbf{Related benchmarks along four design axes.} LADDER combines domain-dependent feedback, repeated observations, and explicit reliability-estimation controls.}
\label{tab:related_axes}
\end{table*}

\paragraph{Interactive agent evaluation.} Multi-step and multi-turn benchmarks relate model reliability to behavior during task execution. \citet{barkan2026capable} study models' predictions of their own success on BigCodeBench and SWE-Bench. AgentBoard~\citep{ma2024agentboard} evaluates intermediate progress and task completion, while $\tau$-bench~\citep{yao2024taubench} measures performance in interactive tasks. These evaluations motivate a complementary target: whether interaction experience yields an accurate account of a model's capabilities. LADDER studies this target by comparing estimated domain profiles with observed performance under imperfect feedback.

\paragraph{Agent memory and feedback learning.} Memory systems study how information from previous interactions can be retained and used in subsequent reasoning. MemGPT~\citep{packer2023memgpt}, Mem0~\citep{mem02025}, RMM~\citep{tan2025rmm}, A-MEM~\citep{xu2025amem}, MemOS~\citep{li2025memos}, and MEM1~\citep{mem1_2026} explore this capability, while MemoryAgentBench~\citep{hu2026memoryagent} evaluates memory use over extended interactions. RLHF~\citep{ouyang2022training} and DPO~\citep{rafailov2024direct} use feedback to update model parameters. More broadly, \citet{wu2025postllm} review how knowledge sharing and complementary model strengths support collaboration and co-evolution. LADDER examines the accuracy of capability estimates formed from interaction records with fixed model parameters, connecting accumulated experience to an explicit self-assessment target.

\paragraph{Imperfect feedback and statistical estimation.} Research on implicit feedback~\citep{liu2025implicit} and confidence after feedback training~\citep{leng2025taming} motivates examining how feedback informs capability judgments. Statistical methods for noisy labels provide complementary references: Dawid--Skene~\citep{dawid1979maximum} models annotator confusion, while GLAD~\citep{whitehill2009glad} relates annotator ability to item difficulty. LADDER adapts these estimation ideas to domain performance and compares them with direct counting and a global feedback mean, examining how feedback quality and aggregation affect the resulting capability profile.

\section{LADDER}
\label{sec:benchmark}

\paragraph{Evaluation objective.} Reliable self-assessment should agree with observed task performance. Yet interaction feedback may be missing or incorrect, so user approval or criticism alone does not establish whether a model accurately understands its capabilities. LADDER therefore asks whether models can use imperfect feedback to accurately summarize their performance across domains. By relating reported capabilities to recorded answer outcomes, the benchmark examines whether interaction experience supports trustworthy self-knowledge.

\paragraph{Design rationale.} We record model answers and natural-language feedback from simulated users, and use scored answer outcomes to establish a domain-performance reference. Against this common target, LADDER compares capability estimates under different information conditions. Prior self-assessments and feedback histories examine existing capability knowledge and reconstruction from interaction experience, while verified outcomes and supplied statistics provide references with more complete information. This ladder allows us to examine whether prior assessments are accurate, whether feedback improves them, and whether models can report known performance (Figure~\ref{fig:overview}).

\paragraph{Task definition.} A session contains $T$ interactions. Turn $t$ records a question $q_t$, an answer $a_t$, reported confidence $c_t$, and natural-language feedback $f_t$. Let $\mathcal D$ denote the domain set. A profile estimator uses its available information to produce domain-accuracy estimates $\hat p:\mathcal D\to[0,1]$, evaluated against empirical domain accuracy $p^*(d)$ from scored answer outcomes. The answering model sees the current question; profile estimators receive the records permitted by their information condition after the session. Domain identifiers are supplied to the estimators.

The benchmark uses 50 recorded sessions of 50 attempted turns per model over 20 MMLU subjects~\citep{hendrycks2021measuring}. Strong/Medium/Weak tiers summarize each model's subject accuracy. Every simulated persona has three assigned expertise subjects. This assignment controls the requested feedback behavior; realized label agreement, coverage, confirmation, and correction are measured separately (Appendix~\ref{app:feedback_channel}).

We additionally construct three extension tasks---GSM8K~\citep{cobbe2021gsm8k}, HumanEval~\citep{chen2021humaneval}, and BFCL tool-calling~\citep{bfcl2024}, collectively called the \emph{extension suite}---to test cross-task generalisation (Appendix~\ref{app:fourtask}). Full task formalisation, session protocol, question-pool partitioning, and persona parameterisation are deferred to Appendix~\ref{app:bench_details}.

\begin{figure}[t]
\centering
\includegraphics[width=\columnwidth]{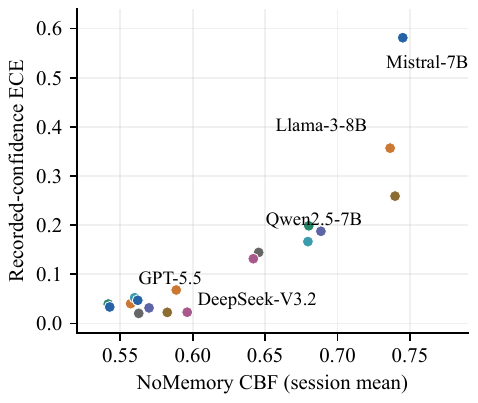}
\caption{\textbf{CBF and ECE describe different targets.} Constant-profile CBF and recorded-confidence ECE for all 18 models.}
\label{fig:cbf_ece}
\end{figure}

\paragraph{Capability Boundary Fidelity (CBF).} Our central metric scores how closely the agent's self-assessed per-domain accuracy tracks the truth:
\begin{equation}
  \begin{gathered}
  \text{CBF} = 1 - \frac{1}{|\mathcal{D}|} \sum_{d \in \mathcal{D}} \left| \hat{p}(d) - p^*(d) \right|,\\
  \quad \text{CBF} \in [0, 1].
  \end{gathered}
  \label{eq:cbf}
\end{equation} CBF is computed at the session endpoint over domains containing at least two valid answers. All estimators share this target and support. ECE uses individual confidence--outcome pairs, whereas CBF uses domain-profile pairs (Figure~\ref{fig:cbf_ece}). A constant profile can achieve high CBF when domain accuracies are close to its value; CBF therefore measures profile error, not task competence.

\paragraph{Baseline-adjusted CBF: relative to a fixed reference.} To express improvement relative to the no-memory reference, we define a normalized variant that measures the fraction of the oracle gap a method closes relative to the constant baseline:
\begin{equation}
  \begin{gathered}
  \widetilde{\text{CBF}}(m, \text{method}) =\\
  \frac{\text{CBF}_{\text{method}}(m) - \text{CBF}_{\text{NoMem}}(m)}{\text{CBF}_{\text{Oracle}}(m) - \text{CBF}_{\text{NoMem}}(m)},
  \end{gathered}
  \label{eq:adj_cbf}
\end{equation} Here NoMem is the constant .5 profile and Oracle is exact counting with CBF 1. A value of 1 closes the reference error, 0 ties NoMem, and a negative value performs worse. Raw CBF supports within-model comparisons; adjusted CBF reports relative error reduction on the same target.
\paragraph{Direction and domain shape.} To distinguish overall miscalibration from domain discrimination, we also report
\begin{align}
 b&=|\mathcal D|^{-1}\sum_d(\hat p_d-p_d^*),\\
 e_{\rm shape}&=|\mathcal D|^{-1}\sum_d| (\hat p_d-\bar{\hat p})-(p_d^*-\bar p^*)|.
\end{align}
Positive $b$ denotes overestimation; lower $e_{\rm shape}$ denotes better recovery of domain contrasts. These complement absolute error rather than form an additive decomposition.

\paragraph{Auxiliary metrics and domain structure.} We report error rate, ECE, and Brier score as auxiliary measures; their definitions appear in Appendix~\ref{app:bench_details}. ECE follows \citet{guo2017calibration}; Brier measures squared confidence--outcome error. CBF requires multiple domains. For the single-domain extension tasks (GSM8K, HumanEval, and the BFCL name/key control), we report task scores and confidence--score gaps (Appendix~\ref{app:fourtask}).

\paragraph{Cross-model comparability.} The 20 subjects are grouped into Strong/Medium/Weak tiers relative to each model's measured subject accuracy. CBF compares methods against the same model-specific target $p^*(d)$. Means across models summarize profile-estimation error; task accuracy is reported separately. Appendix~\ref{app:bench_details} gives the subject inventory and scoring units.

\paragraph{Baselines: a five-level diagnostic ladder.} The baselines and reliability controls share five information levels.
\begin{itemize}[nosep,leftmargin=1em]
\item \textbf{L0:} NoMemory assigns $\hat p_d=.5$; per-answer confidence is evaluated separately.
\item \textbf{L1:} ZS-SK asks the model for domain accuracies without a session history.
\item \textbf{L2:} LSA reads question and feedback history. UF-PDC, Platt, HistBin, Bayes, EM-Joint, Global, DS, and GLAD-style aggregate the same strictly decoded positive/negative label stream; ambiguous labels are missing. Platt and HistBin recalibrate per-answer confidence but use domain counts for their CBF profiles.
\item \textbf{L3:} OraclePDC counts verified outcomes on the same scored turns. Its CBF of 1 is an exact reconstruction reference.
\item \textbf{L4:} GT-SelfStats asks the model to report supplied empirical domain accuracies.
\end{itemize}
LSA's natural-language history and the algorithmic label stream are distinct input representations. The matched-input control fixes the supplied label history to examine aggregation directly. Baseline implementations appear in Appendix~\ref{app:baselines}.

\section{Diagnostic Results: Where Does Agent Self-Knowledge Fail?}
\label{sec:experiments}

We evaluate the five information levels across 18 models, then examine matched controls, cross-task variation, and held-out prediction.

\subsection{Setup}
\label{sec:setup}

The evaluation covers 18 models from seven families (Appendix~\ref{app:api_versions}), with 50 recorded sessions of 50 turns per model. Accuracy is the fraction of correct answers among parsable A--D responses with A--D references. All estimators use the same scored answers and corresponding feedback; missing feedback supplies no label but does not remove an answer from the target. Each session profile covers domains with at least two scored answers. Session means compare estimators on the same interaction histories, while pooled scores assess estimation after combining a model\textquotesingle s histories. The five levels compare different information conditions against these shared targets.

\subsection{Main Results}
\label{sec:results}

\begin{table*}[t]\centering\fontsize{9}{10.8}\selectfont\setlength{\tabcolsep}{4pt}

\fitTable{\begin{tabular}{lrrrrrrrrrrr}\toprule
Model (Acc) & $N$ & L0 & L1 & \multicolumn{4}{c}{L2} & L3 & L4 & \multicolumn{2}{c}{L1 errors} \\\cmidrule(lr){5-8}\cmidrule(lr){11-12} & & NoMem & ZS-SK & LSA & Count* & Bayes & EM & OraPDC & GT-Stats & $b$ & $e_{\rm shape}$ \\\midrule
Mistral-7B (41.6\%) & 2263 & 0.745 & 0.677 & 0.638 & 0.734 & 0.748 & \textbf{0.748} & 1.000 & 1.000 & +0.202 & 0.269 \\
Llama-3-8B (57.7\%) & 2268 & 0.736 & 0.734 & 0.549 & 0.739 & 0.747 & \textbf{0.748} & 1.000 & 1.000 & -0.028 & 0.259 \\
GLM-5 (95.8\%) & 1605 & 0.542 & \textbf{0.904} & 0.568 & 0.595 & 0.575 & 0.572 & 1.000 & 1.000 & -0.034 & 0.071 \\
DeepSeek-V3.2 (88.6\%) & 1831 & 0.596 & \textbf{0.834} & 0.761 & 0.638 & 0.628 & 0.626 & 1.000 & 1.000 & -0.016 & 0.149 \\
Qwen-Turbo (75.5\%) & 2358 & 0.688 & \textbf{0.772} & 0.745 & 0.698 & 0.707 & 0.711 & 1.000 & 1.000 & +0.159 & 0.212 \\
Qwen2.5-7B (60.6\%) & 2462 & 0.740 & 0.670 & 0.732 & 0.741 & 0.751 & \textbf{0.753} & 1.000 & 1.000 & +0.251 & 0.254 \\
Qwen-Max (78.8\%) & 2368 & 0.680 & \textbf{0.789} & 0.757 & 0.666 & 0.679 & 0.680 & 1.000 & 1.000 & +0.063 & 0.204 \\
Qwen-Plus (82.2\%) & 2238 & 0.646 & \textbf{0.806} & 0.558 & 0.672 & 0.667 & 0.667 & 1.000 & 1.000 & +0.108 & 0.200 \\
Qwen3.5-27B (95.9\%) & 1671 & 0.543 & \textbf{0.886} & 0.571 & 0.595 & 0.576 & 0.573 & 1.000 & 1.000 & -0.046 & 0.083 \\
GLM-5.1 (94.2\%) & 1556 & 0.557 & \textbf{0.608} & 0.473 & 0.579 & 0.577 & 0.578 & 1.000 & 1.000 & -0.374 & 0.148 \\
GPT-5.4-mini (75.0\%) & 2467 & 0.680 & \textbf{0.764} & 0.747 & 0.686 & 0.688 & 0.686 & 1.000 & 1.000 & +0.053 & 0.229 \\
GPT-5.4 (82.8\%) & 2472 & 0.642 & \textbf{0.797} & 0.762 & 0.647 & 0.650 & 0.651 & 1.000 & 1.000 & +0.008 & 0.195 \\
GPT-5.5 (92.8\%) & 2500 & 0.570 & \textbf{0.858} & 0.844 & 0.612 & 0.603 & 0.605 & 1.000 & 1.000 & -0.059 & 0.106 \\
Claude-Opus-4.7 (90.4\%) & 1167 & 0.582 & \textbf{0.742} & 0.706 & 0.624 & 0.613 & 0.609 & 1.000 & 1.000 & -0.225 & 0.129 \\
Claude-Sonnet-4.6 (93.0\%) & 914 & 0.560 & \textbf{0.801} & 0.739 & 0.592 & 0.583 & 0.579 & 1.000 & 1.000 & -0.154 & 0.099 \\
Claude-Opus-4.6 (93.0\%) & 949 & 0.563 & \textbf{0.771} & 0.753 & 0.620 & 0.601 & 0.592 & 1.000 & 1.000 & -0.187 & 0.110 \\
DeepSeek-V4-Pro (92.8\%) & 2452 & 0.562 & 0.316 & 0.414 & \textbf{0.598} & 0.590 & 0.592 & 1.000 & 1.000 & -0.682 & 0.103 \\
Qwen3.6-Plus (90.7\%) & 2357 & 0.589 & 0.719 & \textbf{0.821} & 0.616 & 0.614 & 0.616 & 1.000 & 1.000 & -0.233 & 0.146 \\\bottomrule\end{tabular}}
\caption{\textbf{Main results across 18 models and all five levels.} Mean session CBF on shared domain targets. $N$ is the number of scored answers; parentheses give accuracy. Count* is the shared domain output of UF-PDC, Platt, and HistBin. Bold marks the best displayed L0--L2 score. The last columns report L1 signed bias and centered domain-shape error on the same session targets. Additional L2 controls appear in Table~\ref{tab:matched_l2}.}
\label{tab:main}
\end{table*}

\begin{table*}[t]\centering\fontsize{9}{10.8}\selectfont\setlength{\tabcolsep}{4pt}

\fitTable{\begin{tabular}{lccccc}\toprule
Model & Global & Count & Beta & DS & GLAD-style \\\midrule
Mistral-7B & 0.691/0.812 & 0.734/0.809 & 0.748/0.822 & 0.742/0.799 & 0.731/0.712 \\
Llama-3-8B & 0.717/0.835 & 0.739/0.852 & 0.747/0.876 & 0.748/0.868 & 0.734/0.730 \\
GLM-5 & 0.774/0.934 & 0.595/0.862 & 0.575/0.842 & 0.575/0.883 & 0.594/0.908 \\
DeepSeek-V3.2 & 0.769/0.925 & 0.638/0.849 & 0.628/0.852 & 0.627/0.845 & 0.636/0.885 \\
Qwen-Turbo & 0.751/0.897 & 0.698/0.892 & 0.707/0.911 & 0.703/0.813 & 0.694/0.793 \\
Qwen2.5-7B & 0.715/0.857 & 0.741/0.846 & 0.751/0.874 & 0.748/0.791 & 0.738/0.724 \\
Qwen-Max & 0.660/0.732 & 0.666/0.710 & 0.679/0.731 & 0.672/0.676 & 0.667/0.643 \\
Qwen-Plus & 0.750/0.887 & 0.672/0.878 & 0.667/0.883 & 0.670/0.871 & 0.670/0.849 \\
Qwen3.5-27B & 0.775/0.922 & 0.595/0.888 & 0.576/0.847 & 0.577/0.820 & 0.594/0.940 \\
GLM-5.1 & 0.711/0.878 & 0.579/0.864 & 0.577/0.794 & 0.580/0.799 & 0.578/0.919 \\
GPT-5.4-mini & 0.700/0.861 & 0.686/0.820 & 0.688/0.859 & 0.686/0.849 & 0.685/0.811 \\
GPT-5.4 & 0.689/0.908 & 0.647/0.839 & 0.650/0.851 & 0.648/0.781 & 0.647/0.857 \\
GPT-5.5 & 0.800/0.931 & 0.612/0.888 & 0.603/0.873 & 0.601/0.818 & 0.609/0.909 \\
Claude-Opus-4.7 & 0.776/0.908 & 0.624/0.847 & 0.613/0.837 & 0.613/0.830 & 0.621/0.890 \\
Claude-Sonnet-4.6 & 0.772/0.902 & 0.592/0.872 & 0.583/0.783 & 0.589/0.838 & 0.590/0.915 \\
Claude-Opus-4.6 & 0.779/0.907 & 0.620/0.877 & 0.601/0.830 & 0.599/0.833 & 0.616/0.917 \\
DeepSeek-V4-Pro & 0.780/0.904 & 0.598/0.870 & 0.590/0.860 & 0.592/0.819 & 0.596/0.913 \\
Qwen3.6-Plus & 0.777/0.893 & 0.616/0.868 & 0.614/0.859 & 0.613/0.847 & 0.612/0.903 \\\bottomrule\end{tabular}}
\caption{\textbf{Additional matched L2 comparisons for all 18 models.} Each cell reports session/pooled CBF. All five methods receive identical labels and use identical scoring domains.}
\label{tab:matched_l2}
\end{table*}

Table~\ref{tab:main} retains the complete model panel and the original five-level comparison. Three results organize the diagnosis: prior self-assessment differs from history-based estimation; pooled aggregation can change the ordering of noisy-feedback estimators; and numerical reporting can succeed even when aggregation is inaccurate.

\paragraph{Overall accuracy and domain structure.} Global leads the five matched algorithmic controls on 14/18 session-level rows (Table~\ref{tab:matched_l2}). On GPT-5.5, its CBF is 0.800, compared with 0.603 for Beta and 0.609 for GLAD-style. Global assigns every domain one value: its score measures recovery of the overall level. A domain profile must additionally reduce centered shape error $e_{\rm shape}$ to demonstrate useful differentiation. This contrast is a benchmark finding: high CBF alone can reward overall calibration without learning domain differences. The paired level and shape measures identify which improvement an estimator provides.

\paragraph{H1: reporting supplied statistics.} All 18 complete main-panel L4 evaluations reproduce the supplied domain statistics, giving CBF 1.000. The three-model matched-input experiment gives the same result. In contrast, clean-label LSA on Qwen2.5-7B reaches .835. This contrast distinguishes successful numerical reporting from accurate aggregation of a history.

\paragraph{H2: prior domain assessment.} L1 improves on the constant L0 reference for 14 of the 18 models, but the size and direction of the prior error vary. For example, Qwen2.5-7B scores 0.670 at L1, below its L0 score of 0.740. The signed and centered errors in Table~\ref{tab:main} distinguish a misplaced overall confidence level from incorrect differences among domains. A useful prior must capture both: a high CBF alone is insufficient evidence of accurate domain discrimination.

\paragraph{H3: estimating performance from feedback.} Reliability-aware estimation does not uniformly improve on simple aggregation. Session and pooled fits use different evidence budgets: for Qwen2.5-7B, pooled Beta reaches 0.874 and GLAD-style 0.724. All main and sensitivity comparisons use the same positive-ability initialization for GLAD-style. The L2--L3 gap combines label errors, missing feedback, and estimator behavior; Appendix~\ref{app:decoder_ablation} separates label correction from coverage. The comparisons describe the evaluated adaptations under frozen feedback rules; Tables~\ref{tab:convergence} and~\ref{tab:decoder_sensitivity} examine optimization and decoding sensitivity.

\paragraph{Input-matched aggregation.} The common-pool experiment holds the 398 scored questions and feedback labels fixed for GPT-5.5, DeepSeek-V4-Pro, and Qwen2.5-7B. At full coverage, changing from clean labels to 30\% symmetric label noise changes LSA CBF from .989 to .715, .995 to .760, and .835 to .886. The Qwen result shows that aggregation error and label noise need not affect CBF monotonically. Full results for all label channels and estimators appear in Appendix~\ref{app:matched_control}.

\paragraph{Label quality and coverage.} Across the 18 models, mean pooled Count CBF rises from 0.852 with decoded feedback to 0.886 after correcting labels at the same observed positions, and to 1.000 with complete correct labels. The larger gain from complete coverage shows that missing feedback, in addition to incorrect feedback, limits profile reconstruction (Appendix~\ref{app:decoder_ablation}).

\subsection{Diagnosis via the Ladder Visualization}
\label{sec:analysis}
\label{sec:bottleneck}

Figure~\ref{fig:ladder} displays the five levels for all 18 models. Each row follows the same target across all levels, with LSA representing L2. The matrix exposes differences among L0--L2 while showing the L3 and L4 references explicitly. Algorithmic alternatives appear in Tables~\ref{tab:main} and~\ref{tab:matched_l2}.

\begingroup
\setlength{\intextsep}{6pt}
\setlength{\abovecaptionskip}{5pt}
\begin{figure}[!t]
\centering
\includegraphics[width=.93\columnwidth]{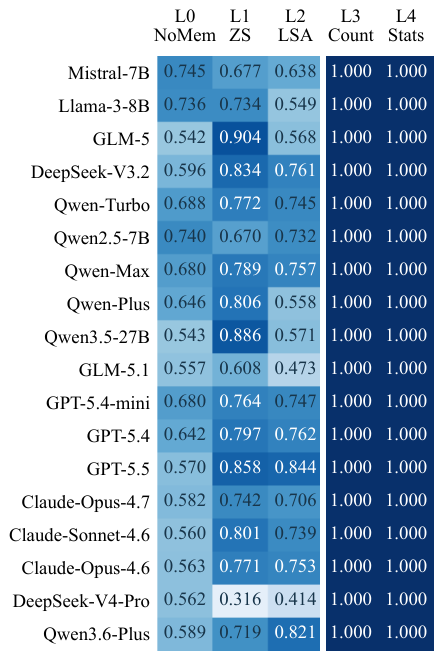}
\caption{\textbf{Five-level CBF for all 18 models.} Each cell gives mean session CBF. L2 uses LSA; L3 counts verified outcomes and L4 reports supplied statistics.}
\label{fig:ladder}
\end{figure}
\endgroup

\begin{figure*}[t]
\centering
\includegraphics[width=.9\textwidth]{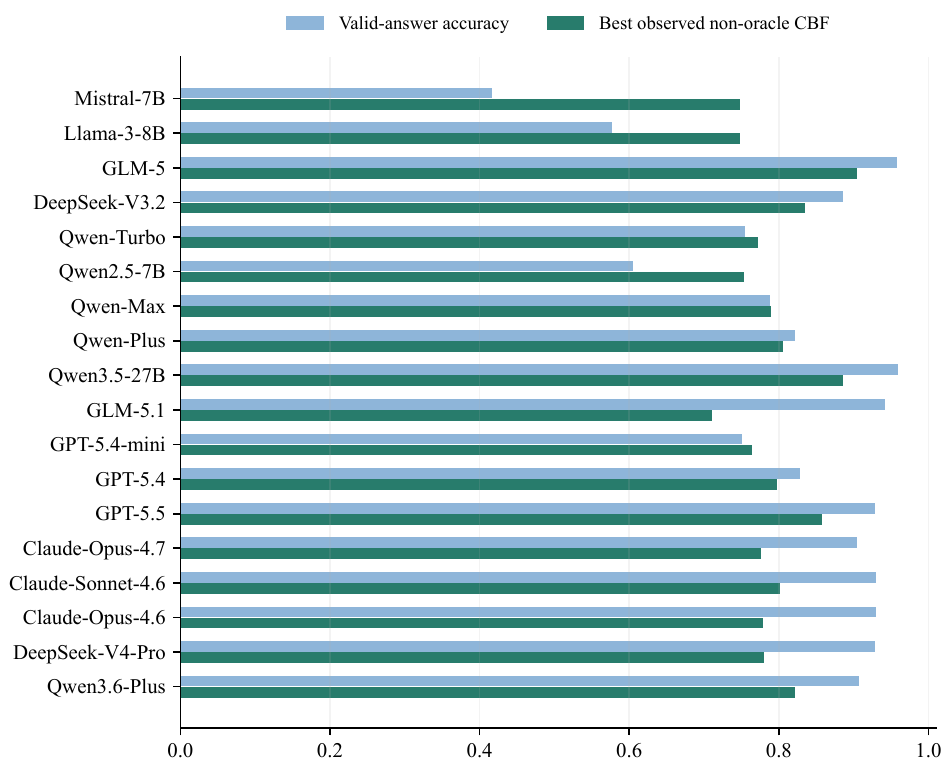}
\caption{\textbf{Capability and profile-estimation spectrum.} Answer accuracy and the highest observed non-oracle CBF across the 18-model panel.}
\label{fig:spectrum}
\end{figure*}

\paragraph{Task competence and profile accuracy.} Figure~\ref{fig:spectrum} places answer accuracy beside the highest observed non-oracle CBF for each model. These quantities describe different capabilities: solving the questions and estimating the resulting performance profile. Their different model orderings show why task accuracy cannot substitute for self-assessment evaluation. The spectrum also places the five-level comparison in context: the value of a profile estimate depends on how closely it captures the model's own outcomes, rather than on how often that model answers correctly.

The adjacent contrasts answer different questions. L4--L3 asks whether supplied statistics can be reported; L1--L3 quantifies prior error; L2--L1 measures the change from using a history; and L2--L3 measures the reconstruction gap. A large reconstruction gap motivates interventions on decoding, coverage, and aggregation. The matched-input and noise experiments distinguish these interventions without equating the performance of an external estimator with the answering model's online learning ability.

\paragraph{A diagnostic per-model selection rule (R4).} R4 is a descriptive comparison: select ZS if its CBF exceeds NoMem, and LSA otherwise. This rule uses a scored target to choose an estimator. It measures the potential value of choosing between prior and history on observed tasks; prediction on new tasks requires a separate held-out test (Appendix~\ref{app:r4}).

\subsection{Cross-Task Heterogeneity and Profile Transfer}\label{sec:ablation}

Across MMLU, GSM8K, HumanEval, and BFCL, task accuracy and reported confidence can differ substantially for the same model. The 18-model extension tables compare confidence minus accuracy where model-produced confidence is available (Appendix~\ref{app:fourtask}). MMLU-Pro and BBH add subject-level prior comparisons for all eight frontier models. To examine prediction directly, a separate ReAct analysis uses the first four questions per subject to predict performance on the next four. Across six configurations and 960 records, empirical ReAct Count profiles improve on a global training-accuracy predictor in two configurations, while zero-shot profiles improve in one. Domain structure is therefore a candidate prediction signal whose benefit needs to be established on held-out tasks. Estimator rankings also change across threshold, random, and semantic domain partitions; Global leads four of five models after merging subjects into four MMLU categories (Appendix~\ref{app:partition_sensitivity}). Recorded tool-use rates across the six configurations range from 37.5\% to 100\%; this is tool-assisted QA.

\section{Conclusion}
\label{sec:conclusion}

LADDER evaluates domain-level self-assessment under noisy feedback through five information conditions. Across 18 models, the results separate overall calibration from domain discrimination and numerical reporting from history aggregation. Global feedback averages provide a strong reference, while the paired level and shape metrics reveal whether domain-specific estimates add useful information. Matched-input controls identify feedback quality, coverage, and aggregation as distinct sources of profile error.

The held-out ReAct analysis shows that accurate reconstruction need not imply predictive benefit. LADDER connects self-knowledge to recorded performance and tests its value on new tasks. Further evaluation should compare richer memory representations under matched targets and feedback budgets, using larger agentic task sets and documented human-feedback studies.

\par\label{sec:main_end}
\clearpage
\section*{Limitations}
\label{sec:limitations}

LADDER studies self-assessment primarily through structured question answering. Extending its conclusions to multilingual, open-ended, and long-horizon interactive tasks requires broader evaluation, where task success and useful feedback may be harder to define.

The feedback in this study comes from simulated users. Real human collaboration involves changing goals, implicit judgments, and more diverse response patterns. A central direction for future work is to test whether the resulting self-assessments can improve decisions in sustained human--agent interaction.

\section*{Ethical considerations}
The study uses public benchmark questions and LLM-simulated feedback. No human participants are involved in the reported experiments. Benchmark materials remain subject to their respective licenses. Domain-profile estimates can help identify mismatches between reported and observed performance, but their use in consequential decisions requires validation on the intended task and user population.

\bibliography{references}

\begin{thebibliography}{53}
\providecommand{\natexlab}[1]{#1}

\bibitem[{Ackerman(2026)}]{ackerman2026metacognition}
Christopher Ackerman. 2026.
\newblock \href {https://arxiv.org/abs/2509.21545} {Evidence for limited
  metacognition in llms}.
\newblock \emph{Proceedings of ICLR}.

\bibitem[{Ahdritz et~al.(2024)Ahdritz, Qin, Vyas, Barak, and
  Edelman}]{ahdritz2024knowable}
Gustaf Ahdritz, Tian Qin, Nikhil Vyas, Boaz Barak, and Benjamin~L. Edelman.
  2024.
\newblock Distinguishing the knowable from the unknowable with language models.
\newblock \emph{Proceedings of ICML}.

\bibitem[{Barkan et~al.(2026)Barkan, Black, and Sourbut}]{barkan2026capable}
Casey~O. Barkan, Sid Black, and Oliver Sourbut. 2026.
\newblock \href {https://arxiv.org/abs/2512.24661} {Do large language models
  know what they are capable of?}
\newblock \emph{Proceedings of ICLR}.

\bibitem[{Binder et~al.(2025)Binder, Chua, Korbak, Sleight, Hughes, Long,
  Perez, Turpin, and Evans}]{binder2025introspection}
Felix~J Binder, James Chua, Tomek Korbak, Henry Sleight, John Hughes, Robert
  Long, Ethan Perez, Miles Turpin, and Owain Evans. 2025.
\newblock Looking inward: Language models can learn about themselves by
  introspection.
\newblock \emph{Proceedings of ICLR}.

\bibitem[{Chen et~al.(2024)Chen, Perez-Lebel, Suchanek, and
  Varoquaux}]{chen2024reconfidencing}
Lihu Chen, Alexandre Perez-Lebel, Fabian~M Suchanek, and Ga{\"e}l Varoquaux.
  2024.
\newblock Reconfidencing llms from the grouping loss perspective.
\newblock \emph{Findings of EMNLP}.

\bibitem[{Chen et~al.(2021)Chen, Tworek, Jun, Yuan et~al.}]{chen2021humaneval}
Mark Chen, Jerry Tworek, Heewoo Jun, Qiming Yuan, et~al. 2021.
\newblock Evaluating large language models trained on code.
\newblock \emph{arXiv preprint arXiv:2107.03374}.

\bibitem[{Chhikara et~al.(2025)Chhikara, Khant, Aryan, Singh, and
  Yadav}]{mem02025}
Prateek Chhikara, Dev Khant, Saket Aryan, Taranjeet Singh, and Deshraj Yadav.
  2025.
\newblock \href {https://arxiv.org/abs/2504.19413} {Mem0: Building
  production-ready ai agents with scalable long-term memory}.
\newblock \emph{arXiv preprint arXiv:2504.19413}.

\bibitem[{Cobbe et~al.(2021)Cobbe, Kosaraju, Bavarian, Chen, Jun, Kaiser,
  Plappert, Tworek, Hilton, Nakano et~al.}]{cobbe2021gsm8k}
Karl Cobbe, Vineet Kosaraju, Mohammad Bavarian, Mark Chen, Heewoo Jun, Lukasz
  Kaiser, Matthias Plappert, Jerry Tworek, Jacob Hilton, Reiichiro Nakano,
  et~al. 2021.
\newblock Training verifiers to solve math word problems.
\newblock \emph{arXiv preprint arXiv:2110.14168}.

\bibitem[{Dawid and Skene(1979)}]{dawid1979maximum}
A~Philip Dawid and Allan~M Skene. 1979.
\newblock Maximum likelihood estimation of observer error-rates using the em
  algorithm.
\newblock \emph{Journal of the Royal Statistical Society Series C: Applied
  Statistics}, 28(1):20--28.

\bibitem[{Dong et~al.(2025)Dong, Jiang et~al.}]{dong2025rlplus}
Y.~Dong, X.~Jiang, et~al. 2025.
\newblock \href {https://arxiv.org/abs/2508.00222} {Rl-plus: Countering
  capability boundary collapse of llms in reinforcement learning with
  hybrid-policy optimization}.
\newblock \emph{arXiv preprint arXiv:2508.00222}.

\bibitem[{Fang et~al.(2025)Fang, Zhao, and Cheng}]{fang2025ccg}
Ke~Fang, Tianyi Zhao, and Lu~Cheng. 2025.
\newblock \href {https://arxiv.org/abs/2508.14390v1} {Credence calibration
  game? calibrating large language models through structured play}.
\newblock \emph{arXiv preprint arXiv:2508.14390v1}.

\bibitem[{Geng et~al.(2024)Geng, Cai, Wang, Koeppl, Nakov, and
  Gurevych}]{geng2024survey}
Jiahui Geng, Fengyu Cai, Yuxia Wang, Heinz Koeppl, Preslav Nakov, and Iryna
  Gurevych. 2024.
\newblock \href {https://aclanthology.org/2024.naacl-long.366/} {A survey of
  confidence estimation and calibration in large language models}.
\newblock In \emph{Proceedings of the 2024 Conference of the North American
  Chapter of the Association for Computational Linguistics: Human Language
  Technologies (Volume 1: Long Papers)}, pages 6577--6595.

\bibitem[{Guo et~al.(2017)Guo, Pleiss, Sun, and
  Weinberger}]{guo2017calibration}
Chuan Guo, Geoff Pleiss, Yu~Sun, and Kilian~Q Weinberger. 2017.
\newblock On calibration of modern neural networks.
\newblock In \emph{ICML}.

\bibitem[{Han et~al.(2024)Han, Li, Chen, Shi, Du, Xiao, Liang, and
  Lin}]{han2024lepe}
Haixia Han, Tingyun Li, Shisong Chen, Jie Shi, Chengyu Du, Yanghua Xiao,
  Jiaqing Liang, and Xin Lin. 2024.
\newblock Enhancing confidence expression in large language models through
  learning from past experience.
\newblock \emph{arXiv preprint arXiv:2404.10315}.

\bibitem[{Hendrycks et~al.(2021)}]{hendrycks2021measuring}
Dan Hendrycks et~al. 2021.
\newblock Measuring massive multitask language understanding.
\newblock In \emph{ICLR}.

\bibitem[{Hu et~al.(2026)Hu, Wang, and McAuley}]{hu2026memoryagent}
Yuanzhe Hu, Yu~Wang, and Julian McAuley. 2026.
\newblock \href {https://arxiv.org/abs/2507.05257} {Evaluating memory in llm
  agents via incremental multi-turn interactions}.
\newblock \emph{Proceedings of ICLR}.

\bibitem[{Kadavath et~al.(2022)}]{kadavath2022language}
Saurav Kadavath et~al. 2022.
\newblock Language models (mostly) know what they know.
\newblock \emph{arXiv preprint arXiv:2207.05221}.

\bibitem[{Kapoor et~al.(2024)Kapoor, Gruver, Roberts, Collins, Pal, Bhatt,
  Weller, Dooley, Goldblum, and Wilson}]{kapoor2024taught}
Sanyam Kapoor, Nate Gruver, Manley Roberts, Katherine Collins, Arka Pal, Umang
  Bhatt, Adrian Weller, Samuel Dooley, Micah Goldblum, and Andrew~Gordon
  Wilson. 2024.
\newblock Large language models must be taught to know what they don't know.
\newblock \emph{Proceedings of NeurIPS}.

\bibitem[{Kirichenko et~al.(2025)Kirichenko, Ibrahim, Chaudhuri, and
  Bell}]{kirichenko2025abstentionbench}
Polina Kirichenko, Mark Ibrahim, Kamalika Chaudhuri, and Samuel~J. Bell. 2025.
\newblock \href {https://arxiv.org/abs/2506.09038} {Abstentionbench: Reasoning
  llms fail on unanswerable questions}.
\newblock \emph{Proceedings of NeurIPS}, 38.

\bibitem[{Laine et~al.(2024)Laine, Chughtai, Betley, Hariharan, Scheurer,
  Balesni, Hobbhahn, Meinke, and Evans}]{laine2024sad}
Rudolf Laine, Bilal Chughtai, Jan Betley, Kaivalya Hariharan, Jérémy
  Scheurer, Mikita Balesni, Marius Hobbhahn, Alexander Meinke, and Owain Evans.
  2024.
\newblock \href {https://arxiv.org/abs/2407.04694} {Me, myself, and ai: The
  situational awareness dataset (sad) for llms}.
\newblock \emph{arXiv preprint arXiv:2407.04694}.

\bibitem[{Leng et~al.(2025)Leng, Huang, Zhu, and Huang}]{leng2025taming}
Jixuan Leng, Chengsong Huang, Banghua Zhu, and Jiaxin Huang. 2025.
\newblock Taming overconfidence in llms: Reward calibration in rlhf.
\newblock \emph{Proceedings of ICLR}.

\bibitem[{Li et~al.(2025{\natexlab{a}})Li, Xiong, Wilson, Mattar, and
  Benna}]{jian2025metacognition}
Ji-An Li, Huadong Xiong, Robert Wilson, Marcelo~G. Mattar, and Marcus~K. Benna.
  2025{\natexlab{a}}.
\newblock \href {https://arxiv.org/abs/2505.13763} {Language models are capable
  of metacognitive monitoring and control of their internal activations}.
\newblock \emph{Proceedings of NeurIPS}, 38.

\bibitem[{Li et~al.(2025{\natexlab{b}})Li, Zhao, Zhang, Li, Xie, Ng, Chua, and
  Deng}]{li2025knowledge_boundary}
Moxin Li, Yong Zhao, Wenxuan Zhang, Shuaiyi Li, Wenya Xie, See-Kiong Ng,
  Tat-Seng Chua, and Yang Deng. 2025{\natexlab{b}}.
\newblock \href {https://doi.org/10.18653/v1/2025.acl-long.256} {Knowledge
  boundary of large language models: A survey}.
\newblock \emph{Proceedings of ACL}, pages 5131--5157.

\bibitem[{Li et~al.(2025{\natexlab{c}})}]{li2025memos}
Zhiyu Li et~al. 2025{\natexlab{c}}.
\newblock \href {https://arxiv.org/abs/2507.03724} {Memos: A memory os for ai
  system}.
\newblock \emph{arXiv preprint arXiv:2507.03724}.

\bibitem[{Liang et~al.(2022)Liang, Bommasani, Lee, Tsipras, Soylu, Yasunaga,
  Zhang, Narayanan, Wu, Kumar et~al.}]{liang2022helm}
Percy Liang, Rishi Bommasani, Tony Lee, Dimitris Tsipras, Dilara Soylu,
  Michihiro Yasunaga, Yian Zhang, Deepak Narayanan, Yuhuai Wu, Ananya Kumar,
  et~al. 2022.
\newblock Holistic evaluation of language models.
\newblock \emph{arXiv preprint arXiv:2211.09110}.

\bibitem[{Lin et~al.(2022)Lin, Hilton, and Evans}]{lin2022truthfulqa}
Stephanie Lin, Jacob Hilton, and Owain Evans. 2022.
\newblock Truthfulqa: Measuring how models mimic human falsehoods.
\newblock \emph{Proceedings of ACL}.

\bibitem[{Liu and van~der Schaar(2025)}]{liu2025metacognitive_learning}
Tennison Liu and Mihaela van~der Schaar. 2025.
\newblock Position: Truly self-improving agents require intrinsic metacognitive
  learning.
\newblock \emph{Proceedings of ICML}.

\bibitem[{Liu et~al.(2025)Liu, Zhang, and Choi}]{liu2025implicit}
Yuhan Liu, Michael~JQ Zhang, and Eunsol Choi. 2025.
\newblock \href {https://doi.org/10.18653/v1/2025.emnlp-main.133} {User
  feedback in human-llm dialogues: A lens to understand users but noisy as a
  learning signal}.
\newblock \emph{Proceedings of EMNLP}, pages 2666--2681.

\bibitem[{Ma et~al.(2024)Ma, Zhang, Zhu, Yang, Yang, Jin, Lan, Kong, and
  He}]{ma2024agentboard}
Chang Ma, Junlei Zhang, Zhihao Zhu, Cheng Yang, Yujiu Yang, Yaohui Jin,
  Zhenzhong Lan, Lingpeng Kong, and Junxian He. 2024.
\newblock Agentboard: An analytical evaluation board of multi-turn llm agents.
\newblock In \emph{Advances in Neural Information Processing Systems
  (NeurIPS)}.

\bibitem[{Manggala et~al.(2025)Manggala, Mastakouri, Kirschbaum,
  Kasiviswanathan, and Ramdas}]{manggala2025qacal}
Putra Manggala, Atalanti Mastakouri, Elke Kirschbaum, Shiva~Prasad
  Kasiviswanathan, and Aaditya Ramdas. 2025.
\newblock Qa-calibration of language model confidence scores.
\newblock \emph{Proceedings of ICLR}.

\bibitem[{Ouyang et~al.(2022)Ouyang, Wu, Jiang, Almeida, Wainwright, Mishkin,
  Zhang, Agarwal, Slama, Ray et~al.}]{ouyang2022training}
Long Ouyang, Jeffrey Wu, Xu~Jiang, Diogo Almeida, Carroll Wainwright, Pamela
  Mishkin, Chong Zhang, Sandhini Agarwal, Katarina Slama, Alex Ray, et~al.
  2022.
\newblock Training language models to follow instructions with human feedback.
\newblock \emph{Advances in Neural Information Processing Systems},
  35:27730--27744.

\bibitem[{Packer et~al.(2023)Packer, Wooders, Lin et~al.}]{packer2023memgpt}
Charles Packer, Sarah Wooders, Kevin Lin, et~al. 2023.
\newblock Memgpt: Towards llms as operating systems.
\newblock \emph{arXiv preprint arXiv:2310.08560}.

\bibitem[{Platt(2000)}]{platt1999probabilistic}
John~C. Platt. 2000.
\newblock \href {https://doi.org/10.7551/mitpress/1113.003.0008} {Probabilities
  for sv machines}.
\newblock In \emph{Advances in Large-Margin Classifiers}, pages 61--74. MIT
  Press.

\bibitem[{Rafailov et~al.(2023)Rafailov, Sharma, Mitchell, Manning, Ermon, and
  Finn}]{rafailov2024direct}
Rafael Rafailov, Archit Sharma, Eric Mitchell, Christopher~D Manning, Stefano
  Ermon, and Chelsea Finn. 2023.
\newblock \href {https://arxiv.org/abs/2305.18290} {Direct preference
  optimization: Your language model is secretly a reward model}.
\newblock \emph{Advances in Neural Information Processing Systems},
  36:53728--53741.

\bibitem[{Shen et~al.(2024)Shen, Das, Greenewald, Sattigeri, Wornell, and
  Ghosh}]{shen2024thermometer}
Maohao Shen, Subhro Das, Kristjan Greenewald, Prasanna Sattigeri, Gregory
  Wornell, and Soumya Ghosh. 2024.
\newblock Thermometer: Towards universal calibration for large language models.
\newblock \emph{Proceedings of ICML}.

\bibitem[{Suzgun et~al.(2022)Suzgun, Scales, Sch{"a}rli, Gehrmann, Tay, Chung,
  Chowdhery, Le, Chi, Zhou, and Wei}]{suzgun2022challenging}
Mirac Suzgun, Nathan Scales, Nathanael Sch{"a}rli, Sebastian Gehrmann, Yi~Tay,
  Hyung~Won Chung, Aakanksha Chowdhery, Quoc~V. Le, Ed~H. Chi, Denny Zhou, and
  Jason Wei. 2022.
\newblock \href {https://arxiv.org/abs/2210.09261} {Challenging big-bench tasks
  and whether chain-of-thought can solve them}.
\newblock \emph{arXiv preprint arXiv:2210.09261}.

\bibitem[{Tan et~al.(2025)}]{tan2025rmm}
Zhen Tan et~al. 2025.
\newblock \href {https://doi.org/10.18653/v1/2025.acl-long.413} {In prospect
  and retrospect: Reflective memory management for long-term personalized
  dialogue agents}.
\newblock \emph{Proceedings of ACL}, pages 8416--8439.

\bibitem[{Traub et~al.(2024)Traub, Bungert, Lüth, Baumgartner, Maier-Hein,
  Maier-Hein, and Jäger}]{traub2024augrc}
Jeremias Traub, Till~J. Bungert, Carsten~T. Lüth, Michael Baumgartner,
  Klaus~H. Maier-Hein, Lena Maier-Hein, and Paul~F. Jäger. 2024.
\newblock \href {https://arxiv.org/abs/2407.01032} {Overcoming common flaws in
  the evaluation of selective classification systems}.
\newblock \emph{Proceedings of NeurIPS}, 37.

\bibitem[{Wei et~al.(2024)}]{wei2024simpleqa}
Jason Wei et~al. 2024.
\newblock \href {https://arxiv.org/abs/2411.04368} {Measuring short-form
  factuality in large language models}.
\newblock \emph{OpenAI Technical Report}.

\bibitem[{Wen et~al.(2025)Wen, Yao, Feng, Xu, Tsvetkov, Howe, and
  Wang}]{li2025abstention}
Bingbing Wen, Jihan Yao, Shangbin Feng, Chenjun Xu, Yulia Tsvetkov, Bill Howe,
  and Lucy~Lu Wang. 2025.
\newblock \href {https://doi.org/10.1162/tacl_a_00754} {Know your limits: A
  survey of abstention in large language models}.
\newblock \emph{Transactions of the Association for Computational Linguistics},
  13:529--556.

\bibitem[{Whitehill et~al.(2009)Whitehill, Ruvolo, Wu, Bergsma, and
  Movellan}]{whitehill2009glad}
Jacob Whitehill, Paul Ruvolo, Tingfan Wu, Jacob Bergsma, and Javier Movellan.
  2009.
\newblock Whose vote should count more: Optimal integration of labels from
  labelers of unknown expertise.
\newblock In \emph{Advances in Neural Information Processing Systems},
  volume~22, pages 2035--2043.

\bibitem[{Wu et~al.(2025)Wu, Shen, B{\"a}ck, Chen, Huang, Jin, Kuang, Li, Lu,
  Miao, Wang, Wei, Wu, Yan, Yang, Yang, Zhang, Zhao, Zhuang, and
  Pan}]{wu2025postllm}
Fei Wu, Tao Shen, Thomas B{\"a}ck, Jingyuan Chen, Gang Huang, Yaochu Jin, Kun
  Kuang, Mengze Li, Cewu Lu, Jiaxu Miao, Yongwei Wang, Ying Wei, Fan Wu, Junchi
  Yan, Hongxia Yang, Yi~Yang, Shengyu Zhang, Zhou Zhao, Yueting Zhuang, and
  Yunhe Pan. 2025.
\newblock \href {https://doi.org/10.1016/j.eng.2024.12.008}
  {Knowledge-empowered, collaborative, and co-evolving {AI} models: The
  post-{LLM} roadmap}.
\newblock \emph{Engineering}, 44:87--100.

\bibitem[{Xiong et~al.(2024)Xiong, Hu, Lu, Li, Fu, He, and Hooi}]{xiong2024can}
Miao Xiong, Zhiyuan Hu, Xinyang Lu, Yifei Li, Jie Fu, Junxian He, and Bryan
  Hooi. 2024.
\newblock Can llms express their uncertainty? an empirical evaluation of
  confidence elicitation in llms.
\newblock In \emph{ICLR}.

\bibitem[{Xu et~al.(2024)Xu, Wu, Diao, Liu, Wang, Chen, and
  Gao}]{xu2024sayself}
Tianyang Xu, Shujin Wu, Shizhe Diao, Xiaoze Liu, Xingyao Wang, Yangyi Chen, and
  Jing Gao. 2024.
\newblock Sayself: Teaching llms to express confidence with self-reflective
  rationales.
\newblock \emph{Proceedings of EMNLP}.

\bibitem[{Xu et~al.(2025)Xu, Liang, Mei, Gao, Tan, and Zhang}]{xu2025amem}
Wujiang Xu, Zujie Liang, Kai Mei, Hang Gao, Juntao Tan, and Yongfeng Zhang.
  2025.
\newblock A-mem: Agentic memory for llm agents.
\newblock \emph{Proceedings of NeurIPS}.

\bibitem[{Yan et~al.(2024)Yan, Mao, Ji, Stoica, Gonzalez, Zhang, and
  Patil}]{bfcl2024}
Fanjia Yan, Huanzhi Mao, Charlie Cheng-Jie Ji, Ion Stoica, Joseph~E. Gonzalez,
  Tianjun Zhang, and Shishir~G. Patil. 2024.
\newblock Berkeley function-calling leaderboard.
\newblock
  \url{https://gorilla.cs.berkeley.edu/blogs/8_berkeley_function_calling_leaderboard.html}.

\bibitem[{Yao et~al.(2024)Yao, Shinn, Razavi, and Narasimhan}]{yao2024taubench}
Shunyu Yao, Noah Shinn, Pedram Razavi, and Karthik Narasimhan. 2024.
\newblock $\\tau$-bench: A benchmark for tool-agent-user interaction in
  real-world domains.
\newblock \emph{arXiv preprint arXiv:2406.12045}.

\bibitem[{Yin et~al.(2024)Yin, Zhang, Ruan, and
  Wan}]{yin2024benchmarking_boundary}
Xunjian Yin, Xu~Zhang, Jie Ruan, and Xiaojun Wan. 2024.
\newblock Benchmarking knowledge boundary for large language models: A
  different perspective on model evaluation.
\newblock \emph{Proceedings of ACL}.

\bibitem[{Yin et~al.(2023)Yin, Sun, Guo, Wu, Qiu, and Huang}]{yin2023selfaware}
Zhangyue Yin, Qiushi Sun, Qipeng Guo, Jiawen Wu, Xipeng Qiu, and Xuanjing
  Huang. 2023.
\newblock Do large language models know what they don't know?
\newblock In \emph{Findings of ACL}.

\bibitem[{Zadrozny and Elkan(2001)}]{zadrozny2001obtaining}
Bianca Zadrozny and Charles Elkan. 2001.
\newblock Obtaining calibrated probability estimates from decision trees and
  naive bayesian classifiers.
\newblock In \emph{Proceedings of ICML}.

\bibitem[{Zhang et~al.(2024)Zhang, Diao, Lin, Fung, Lian, Wang, Chen, Ji, and
  Zhang}]{diao2024rtuning}
Hanning Zhang, Shizhe Diao, Yong Lin, Yi~Fung, Qing Lian, Xingyao Wang, Yangyi
  Chen, Heng Ji, and Tong Zhang. 2024.
\newblock \href {https://doi.org/10.18653/v1/2024.naacl-long.394} {R-tuning:
  Instructing large language models to say i don't know}.
\newblock \emph{Proceedings of NAACL}, pages 7113--7139.

\bibitem[{Zhou et~al.(2026{\natexlab{a}})Zhou, Sun, Ma
  et~al.}]{zhou2026sim2real}
Xuhui Zhou, Weiwei Sun, Qianou Ma, et~al. 2026{\natexlab{a}}.
\newblock \href {https://arxiv.org/abs/2603.11245} {Mind the sim2real gap in
  user simulation for agentic tasks}.
\newblock \emph{arXiv preprint}.

\bibitem[{Zhou et~al.(2026{\natexlab{b}})Zhou, Qu, Wu et~al.}]{mem1_2026}
Zijian Zhou, Ao~Qu, Zhaoxuan Wu, et~al. 2026{\natexlab{b}}.
\newblock \href {https://arxiv.org/abs/2506.15841} {Mem1: Learning to synergize
  memory and reasoning for efficient long-horizon agents}.
\newblock \emph{Proceedings of ICLR}.

\end{thebibliography}

\clearpage
\appendix
\renewcommand{\topfraction}{.9}
\renewcommand{\dbltopfraction}{.9}
\renewcommand{\textfraction}{.08}
\renewcommand{\floatpagefraction}{.65}
\renewcommand{\dblfloatpagefraction}{.65}
\makeatletter
\setlength{\@fptop}{0pt}
\setlength{\@fpsep}{12pt}
\setlength{\@fpbot}{0pt plus 1fil}
\setlength{\@dblfptop}{0pt}
\setlength{\@dblfpsep}{12pt}
\setlength{\@dblfpbot}{0pt plus 1fil}
\makeatother
\raggedbottom

\section{Datasets}\label{app:group_a}

The primary pool contains 500 questions from 20 MMLU subjects, with 25 questions per subject. Each model is evaluated in 50 sessions of 50 attempted turns. The interaction records contain the question, answer, confidence, simulated user feedback, and reference-based correctness label.

MMLU-Pro and BBH provide held-out subject and task comparisons. GSM8K, HumanEval, and BFCL supply additional task-level confidence comparisons, while ReAct records support prediction on held-out questions. Dataset-specific scoring and sample sizes accompany the corresponding results.

\section{Benchmark Details}
\label{app:group_b}

\subsection{Benchmark Details: Task, Protocol, Question Pool, Personas, Metrics}
\label{app:bench_details}

\begin{figure*}[tbp]
\centering
\includegraphics[width=\textwidth]{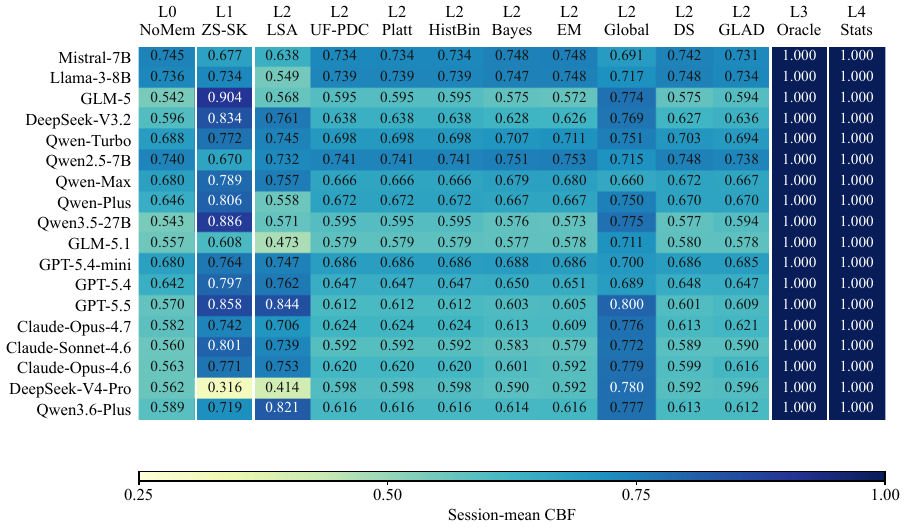}
\caption{\textbf{Full model-by-baseline CBF matrix.} All 18 models and the complete five-level design.}
\label{fig:heatmap}
\end{figure*}

\begin{figure*}[tbp]
\centering
\includegraphics[width=.90\textwidth]{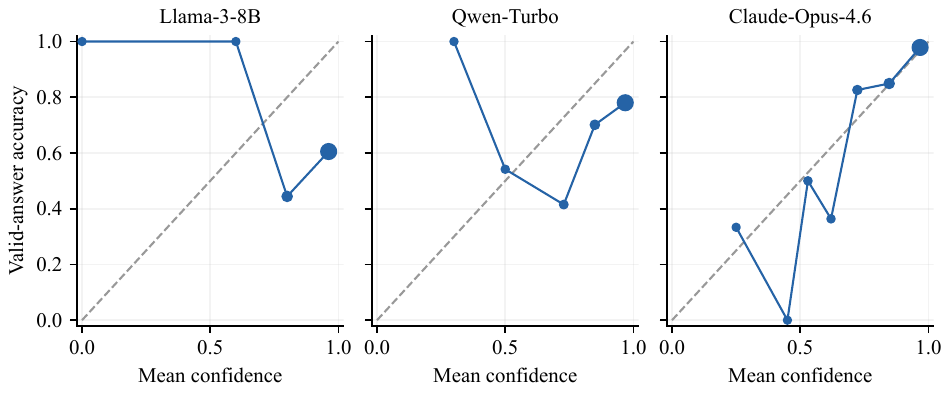}
\caption{\textbf{Recorded-confidence reliability curves.} Bin-level mean confidence and answer accuracy; the last bin includes confidence 1.}
\label{fig:reliability}
\end{figure*}

Figure~\ref{fig:spectrum} in the main text summarizes task accuracy and profile-estimation performance. The supplementary Figure~\ref{fig:heatmap} displays all five levels and their baselines; Figure~\ref{fig:reliability} summarizes per-answer confidence calibration.

\paragraph{Task formalisation.} A session contains $(q_t,a_t,c_t,f_t,d_t,y_t)$ records. Outcomes $y_t$ are used to score $p_d^*$ and are hidden from noisy-feedback estimators. The domain identifier $d_t$ groups observations for profile estimation; the answering prompt contains the question. Feedback is an utterance that may confirm, correct, or leave the answer's validity unclear.

\paragraph{Session protocol.} At each turn, the answering model receives a question and reports an answer and confidence. The simulated user then reacts according to its assigned persona and the current turn. Profile estimators consume the saved records; the next answer is not conditioned on their estimates. Each model contributes 50 sessions of 50 attempted turns.

\paragraph{Question pool and tiers.} The scoring pool contains 20 MMLU subjects. For descriptive tier summaries, subjects are ordered by each model's empirical accuracy and split into 6 strong, 7 medium, and 7 weak subjects. Threshold and random-partition analyses examine sensitivity to this grouping. The 232 numerical-answer turns in the additional math-word-problems domain of each Mistral/Llama archive are outside the 20-subject MCQ target and are excluded.

\paragraph{User personas.} Each persona has three expertise subjects, a response style, and a tolerance parameter between .20 and .40. Within expertise, the prompt requests confirmation of correct answers and correction of incorrect answers; the simulator receives the answer key. Outside expertise, the prompt requests acceptance of correct answers and either uncertainty or acceptance of incorrect answers according to the tolerance probability. Acceptance need not contain an explicit evaluative phrase. The measured channel is the output of the fixed rule decoder on these utterances.

\paragraph{Auxiliary metrics.} Beyond CBF (Eq.~\ref{eq:cbf}, computed once at the end of the 50-turn session), we report: (i) Hallucination Rate $\text{HR} = |\{t : c_t > 0.7 \wedge a_t \neq a_t^*\}| / T$; (ii) Expected Calibration Error following~\citet{guo2017calibration} with $M = 10$ equal-width bins, $\text{ECE} = \sum_{m=1}^M \frac{|B_m|}{T} | \text{acc}(B_m) - \text{conf}(B_m) |$; (iii) Brier Score $= \frac{1}{T} \sum_t (c_t - \mathbb{1}[a_t = a_t^*])^2$.

\subsection{MMLU Subject Selection}
\label{app:subjects}

The 20 subjects comprise 13 STEM subjects, four in MMLU's ``Other'' category, two in humanities, and one in social sciences. The following table illustrates the model-relative tiers.

\begin{table}[tbp]\centering\fontsize{9}{10.8}\selectfont\setlength{\tabcolsep}{4pt}

\fitTable{\begin{tabular}{@{}>{\raggedright\arraybackslash}p{.43\columnwidth}ccc@{}}\toprule
Subject & \makecell{Llama\\3-8B} & \makecell{Qwen\\Turbo} & \makecell{Qwen\\Max} \\\midrule
Abstract Algebra & W & W & M \\
Anatomy & S & S & M \\
Astronomy & M & S & S \\
Business Ethics & S & M & S \\
Clinical Knowledge & S & S & S \\
College Biology & S & S & S \\
College Chemistry & M & M & W \\
College Physics & M & W & W \\
Computer Security & S & M & W \\
Conceptual Physics & W & S & S \\
Econometrics & M & W & M \\
Electrical Engineering & M & M & S \\
Formal Logic & W & W & W \\
Global Facts & W & W & W \\
High School Biology & S & M & M \\
High School Chemistry & M & W & W \\
High School Mathematics & W & S & M \\
High School Physics & M & M & W \\
Machine Learning & W & M & M \\
Moral Scenarios & W & W & M \\\bottomrule\end{tabular}}
\caption{MMLU subjects in the scored records and empirical per-model tiers (S/M/W).}
\label{tab:subjects}
\end{table}

\subsection{User Persona Details}
\label{app:persona}

Each simulated user persona is defined by the following parameters:

\begin{itemize}[nosep,leftmargin=1em]
\item \textbf{Expertise}: three subjects stored with each persona.
\item \textbf{Feedback behavior}: confirmation, correction, or uncertainty, conditioned on expertise and answer correctness.
\item \textbf{Style}: natural-language presentation of the requested behavior.
\end{itemize}

The resulting feedback is decoded using the same rule classifier as the profile estimators. Appendix~\ref{app:feedback_channel} reports the observed channel, including missing labels.

\paragraph{Measured feedback channel.}\label{app:feedback_channel} Coverage is decoded labels / valid answers. Agreement is labels matching the answer key / decoded labels. Confirmation is positive labels on correct answers / all correct answers, and correction is negative labels on incorrect answers / all incorrect answers. Rates are percentages and measure rule-decoded stance. The 353 feedback API-error records remain in the valid-answer target with missing labels.

\begin{table*}[tbp]\centering\fontsize{9}{10.8}\selectfont\setlength{\tabcolsep}{4pt}

\fitTable{\begin{tabular}{llrrrrrr}\toprule
Model & State & Valid & Decoded & Coverage & Agreement & Confirm & Correct \\\midrule
Mistral-7B & in & 355 & 221 & 62.3 & 91.9 & 86.7 & 35.6 \\
Mistral-7B & out & 1908 & 30 & 1.6 & 56.7 & 0.0 & 1.5 \\
Llama-3-8B & in & 336 & 260 & 77.4 & 95.8 & 90.9 & 46.5 \\
Llama-3-8B & out & 1932 & 45 & 2.3 & 62.2 & 0.0 & 3.4 \\
GLM-5 & in & 246 & 218 & 88.6 & 99.5 & 89.5 & 50.0 \\
GLM-5 & out & 1359 & 20 & 1.5 & 5.0 & 0.1 & 0.0 \\
DeepSeek-V3.2 & in & 315 & 260 & 82.5 & 99.6 & 88.5 & 35.1 \\
DeepSeek-V3.2 & out & 1516 & 19 & 1.3 & 15.8 & 0.1 & 1.2 \\
Qwen-Turbo & in & 392 & 305 & 77.8 & 100.0 & 87.1 & 49.5 \\
Qwen-Turbo & out & 1966 & 41 & 2.1 & 17.1 & 0.0 & 1.5 \\
Qwen2.5-7B & in & 376 & 268 & 71.3 & 100.0 & 89.7 & 40.8 \\
Qwen2.5-7B & out & 2086 & 30 & 1.4 & 23.3 & 0.0 & 0.8 \\
Qwen-Max & in & 371 & 121 & 32.6 & 97.5 & 19.8 & 76.9 \\
Qwen-Max & out & 1997 & 37 & 1.9 & 62.2 & 0.8 & 2.4 \\
Qwen-Plus & in & 321 & 277 & 86.3 & 99.3 & 88.9 & 68.0 \\
Qwen-Plus & out & 1917 & 32 & 1.7 & 15.6 & 0.1 & 0.9 \\
Qwen3.5-27B & in & 269 & 235 & 87.4 & 100.0 & 88.8 & 54.5 \\
Qwen3.5-27B & out & 1402 & 24 & 1.7 & 4.2 & 0.1 & 0.0 \\
GLM-5.1 & in & 253 & 166 & 65.6 & 99.4 & 67.5 & 36.8 \\
GLM-5.1 & out & 1303 & 24 & 1.8 & 0.0 & 0.0 & 0.0 \\
GPT-5.4-mini & in & 389 & 109 & 28.0 & 100.0 & 29.0 & 25.2 \\
GPT-5.4-mini & out & 2078 & 13 & 0.6 & 15.4 & 0.0 & 0.4 \\
GPT-5.4 & in & 410 & 132 & 32.2 & 99.2 & 32.9 & 27.0 \\
GPT-5.4 & out & 2062 & 15 & 0.7 & 6.7 & 0.0 & 0.3 \\
GPT-5.5 & in & 417 & 357 & 85.6 & 100.0 & 88.1 & 54.8 \\
GPT-5.5 & out & 2083 & 34 & 1.6 & 20.6 & 0.1 & 4.0 \\
Claude-Opus-4.7 & in & 173 & 147 & 85.0 & 99.3 & 87.9 & 50.0 \\
Claude-Opus-4.7 & out & 994 & 22 & 2.2 & 31.8 & 0.3 & 4.2 \\
Claude-Sonnet-4.6 & in & 142 & 109 & 76.8 & 100.0 & 80.3 & 46.7 \\
Claude-Sonnet-4.6 & out & 772 & 10 & 1.3 & 0.0 & 0.0 & 0.0 \\
Claude-Opus-4.6 & in & 144 & 128 & 88.9 & 100.0 & 90.5 & 57.1 \\
Claude-Opus-4.6 & out & 805 & 14 & 1.7 & 0.0 & 0.0 & 0.0 \\
DeepSeek-V4-Pro & in & 352 & 296 & 84.1 & 100.0 & 87.1 & 46.2 \\
DeepSeek-V4-Pro & out & 2100 & 40 & 1.9 & 2.5 & 0.1 & 0.0 \\
Qwen3.6-Plus & in & 367 & 308 & 83.9 & 100.0 & 87.3 & 54.1 \\
Qwen3.6-Plus & out & 1990 & 45 & 2.3 & 15.6 & 0.2 & 1.6 \\\bottomrule\end{tabular}}
\caption{Observed feedback channel inside/outside persona expertise. Confirm = positive labels on correct answers / all correct answers; Correct = negative labels on incorrect answers / all incorrect answers. These are decoded stance rates. Outside expertise, neutral follow-up questions can express acceptance without a decoded positive label. Qwen-Max has 58/293 in-expertise correct answers with positive labels and 235/293 with neutral labels; 59 neutral replies contain \emph{that was accurate}, which the fixed rules miss. Its correction count is 60/78. The low confirmation rate therefore includes decoder misses; these records do not isolate a simulator-family effect.}
\label{tab:feedback_channel}
\end{table*}

\section{Baseline Implementations}
\label{app:group_c}

\subsection{Baseline Implementations}\label{app:baselines}

The original ten baselines retain their five-level organization. Global, DS, and GLAD-style add comparisons of overall-level estimation and explicit feedback-reliability modeling.

\paragraph{L0 --- NoMemory (B1).} Constant $\hat p_d=.5$ for domain CBF; recorded confidence for per-query metrics.
\paragraph{L1 --- ZS-SK (B11).} The model estimates domain accuracies without interaction history.
\paragraph{L2 --- LSA (B6).} The model estimates domain accuracies from its question and natural-language feedback history.
\paragraph{L2 --- UF-PDC (B3).} Domain proportions of strictly decoded positive feedback.
\paragraph{L2 --- Platt/HistBin (B7/B8).} Online Platt scaling~\citep{platt1999probabilistic} and histogram binning~\citep{zadrozny2001obtaining} recalibrate query confidence; their domain self-assessment interface returns the same decoded-label counts.
\paragraph{L2 --- Bayes-Domain (B9).} A Beta(1,1) prior gives $(s_d+1)/(n_d+2)$.
\paragraph{L2 --- EM-Joint (B10).} A symmetric reliability model with domain correctness prevalence and in/out expertise reliability.
\paragraph{L3 --- OraclePDC (B2).} Exact counts of verified outcomes on the scored domain support.
\paragraph{L4 --- GT-SelfStats (B12).} The model reports supplied empirical statistics.
\paragraph{Additional L2 controls.} Global assigns the pooled feedback Beta mean to every domain. DS uses state-specific confusion probabilities; GLAD-style uses additive ability and difficulty.

\paragraph{Information conditions.} The ladder compares information available for a domain-profile estimate. LSA performs in-context aggregation; Count, Beta, EM-Joint, DS, and GLAD-style instantiate alternative estimators. L1 and L4 provide prior and supplied-statistic references. This grouping does not constitute a direct evaluation of the memory architectures cited in Related Work.

\subsection{Baseline Implementation Details}
\label{app:baselines_impl}

\paragraph{B1: NoMemory.} Its domain self-assessment function returns .5. Answer confidence is a separate output and does not determine this profile.

\paragraph{B2: OraclePDC.} The evaluator maintains verified correct/total counts per recorded domain. Using the same outcomes and support as the target yields CBF 1 exactly.

\paragraph{B3: User-Feedback PDC.} The classifier maps explicit positive/negative feedback to 1/0. Uncertain and neutral responses are unobserved. Platt, HistBin, Bayes, EM-Joint, DS, and GLAD-style use this same stream.

\paragraph{B6: LSA.} At the session endpoint, the model reads domain-indexed question/feedback summaries and returns one probability per domain. The saved representation retains up to 80 question characters and 120 feedback characters per turn. Its complete response is parsed once; a valid profile is not selected by its CBF.

\paragraph{B10: EM-Joint.} The latent outcome $y_t$ has prior $p_{d_t}$ and symmetric agreement probability $r_{h_t}$ for expertise state $h_t$. With a decoded label $z_t$, the E-step computes its posterior outcome; smoothed M-steps update domain prevalence and reliability. Initialization uses $p_d=.5$ and $r_{\rm in}=.7,r_{\rm out}=.5$, followed by at most 30 iterations with tolerance $10^{-4}$. Missing-feedback domains retain .5.

\paragraph{B11: ZS-SK.} L1 estimates domain accuracy without session history. Each 20-domain prior is evaluated on the same eligible session targets as the other levels. The JSON response format appears in Appendix~\ref{app:group_g}.

\paragraph{B12: GT-SelfStats.} The empirical accuracy for each eligible domain of the current session is supplied directly, and the model reports the same values in a structured profile.

\section{Statistical Analysis}
\label{app:group_d}

\subsection{Sample Size and Paired Comparisons}\label{app:power}

\paragraph{Statistical unit.} A model's session-level score averages its eligible sessions. Model-level comparisons pair methods on the same records; pooled comparisons refit each method on all valid records of that model.

\paragraph{Uncertainty.} Session resampling quantifies variation among the observed sessions. Question-cluster resampling groups repeated questions within domain and refits the estimators, preserving the dependence induced by reused questions. The model inventory is a fixed selection rather than a random sample from a model population.

\begin{table}[tbp]\centering\fontsize{9}{10.8}\selectfont\setlength{\tabcolsep}{4pt}

\fitTable{\begin{tabular}{lrr}\toprule
Model & DS$-$Beta & GLAD-style$-$Beta \\\midrule
Qwen2.5-7B & -0.083 & -0.151 \\
GPT-5.4-mini & -0.010 & -0.048 \\
GPT-5.4 & -0.071 & 0.006 \\
GPT-5.5 & -0.056 & 0.036 \\
DeepSeek-V4-Pro & -0.041 & 0.053 \\\bottomrule\end{tabular}}
\caption{Paired pooled CBF differences on shared labels and domains. Question-clustered intervals appear in Table~\ref{tab:paired_shape}.}
\label{tab:effect_sizes}
\end{table}

\paragraph{Sample selection.} The 18-model panel includes an eight-model frontier subset. Method comparisons are descriptive across this fixed inventory; changing the subset changes the target population.

\paragraph{Detection limits.} Sparse per-domain outcomes limit the precision of session profiles. Larger held-out task sets, rather than post-hoc power estimates from observed effects, are needed to resolve small predictive differences.

\subsection{Statistical Robustness}
\label{app:stat_rob}

We report NoMemory CBF with session-bootstrap standard errors (2,000 draws, seed 20261011) for all 18 models. Confidence-based metrics are point estimates over valid answer records and include both endpoints of $[0,1]$.

\begin{table}[tbp]\centering\fontsize{9}{10.8}\selectfont\setlength{\tabcolsep}{4pt}

\fitTable{\begin{tabular}{lrrrr}\toprule
Model & NoMem CBF & ECE & HR & Brier \\\midrule
Mistral-7B & 0.745$\pm$0.008 & 0.582 & 0.583 & 0.581 \\
Llama-3-8B & 0.736$\pm$0.007 & 0.357 & 0.423 & 0.368 \\
GLM-5 & 0.542$\pm$0.005 & 0.039 & 0.042 & 0.042 \\
DeepSeek-V3.2 & 0.596$\pm$0.006 & 0.023 & 0.107 & 0.095 \\
Qwen-Turbo & 0.688$\pm$0.006 & 0.188 & 0.230 & 0.214 \\
Qwen2.5-7B & 0.740$\pm$0.007 & 0.259 & 0.374 & 0.297 \\
Qwen-Max & 0.680$\pm$0.007 & 0.167 & 0.209 & 0.190 \\
Qwen-Plus & 0.646$\pm$0.008 & 0.144 & 0.177 & 0.161 \\
Qwen3.5-27B & 0.543$\pm$0.005 & 0.033 & 0.041 & 0.040 \\
GLM-5.1 & 0.557$\pm$0.011 & 0.040 & 0.056 & 0.053 \\
GPT-5.4-mini & 0.680$\pm$0.006 & 0.199 & 0.246 & 0.215 \\
GPT-5.4 & 0.642$\pm$0.006 & 0.132 & 0.158 & 0.148 \\
GPT-5.5 & 0.570$\pm$0.005 & 0.032 & 0.056 & 0.056 \\
Claude-Opus-4.7 & 0.582$\pm$0.008 & 0.022 & 0.060 & 0.078 \\
Claude-Sonnet-4.6 & 0.560$\pm$0.010 & 0.052 & 0.048 & 0.064 \\
Claude-Opus-4.6 & 0.563$\pm$0.007 & 0.020 & 0.051 & 0.054 \\
DeepSeek-V4-Pro & 0.562$\pm$0.004 & 0.047 & 0.069 & 0.064 \\
Qwen3.6-Plus & 0.589$\pm$0.005 & 0.068 & 0.089 & 0.080 \\\bottomrule\end{tabular}}
\caption{NoMemory CBF and its session-bootstrap standard error; valid-answer confidence metrics.}
\label{tab:bootstrap_stability}
\end{table}

The errors reflect uncertainty conditional on the retained session pool. Paired differences should use their own resampled differences, rather than a threshold based on one method's standard error.

\paragraph{Cross-model summaries.} Table~\ref{tab:boot_pairs} averages paired CBF differences only where both methods are available. Table~\ref{tab:boot_frontier} reports the fixed frontier/non-frontier descriptive contrast with its sample counts.

\begin{table}[tbp]\centering\fontsize{9}{10.8}\selectfont\setlength{\tabcolsep}{4pt}

\fitTable{\begin{tabular}{lrr}\toprule
Comparison & $n$ & Mean difference \\\midrule
ZS $-$ NoMem & 18 & 0.124 \\
ZS $-$ LSA & 18 & 0.073 \\
LSA $-$ NoMem & 18 & 0.051 \\
EM $-$ NoMem & 18 & 0.020 \\
Platt $-$ NoMem & 18 & 0.024 \\\bottomrule\end{tabular}}
\caption{Paired mean differences over models with both estimates.}
\label{tab:boot_pairs}
\end{table}

\begin{table}[tbp]\centering\fontsize{9}{10.8}\selectfont\setlength{\tabcolsep}{4pt}

\fitTable{\begin{tabular}{lrr}\toprule
Method & $n_F/n_O$ & Frontier $-$ Other \\\midrule
NoMem & 8/10 & -0.054 \\
ZS & 8/10 & -0.047 \\
LSA & 8/10 & 0.088 \\
UF & 8/10 & -0.041 \\
Platt & 8/10 & -0.041 \\
HistBin & 8/10 & -0.041 \\
Bayes & 8/10 & -0.048 \\
EM & 8/10 & -0.049 \\\bottomrule\end{tabular}}
\caption{Descriptive raw-CBF contrasts for the fixed model subsets.}
\label{tab:boot_frontier}
\end{table}

Cross-model averages summarize different empirical targets. A larger raw CBF can result from a target profile closer to the constant reference, so these contrasts are not an ordering of model competence.

Normalized CBF is computed from the main table\textquotesingle s session means using Eq.~\ref{eq:adj_cbf}. NoMem is 0 and exact L3 counting is 1 for every model. It reports error reduction relative to the .5 profile.

\begin{table*}[tbp]\centering\fontsize{9}{10.8}\selectfont\setlength{\tabcolsep}{4pt}

\fitTable{\begin{tabular}{lrrrrrrrrr}\toprule
Model & L1 & LSA & Count & Beta & EM & Global & DS & GLAD-style & L4 \\\midrule
Mistral-7B & -0.267 & -0.421 & -0.042 & 0.011 & 0.012 & -0.213 & -0.011 & -0.055 & 1.000 \\
Llama-3-8B & -0.008 & -0.712 & 0.009 & 0.039 & 0.043 & -0.072 & 0.043 & -0.009 & 1.000 \\
GLM-5 & 0.791 & 0.057 & 0.117 & 0.073 & 0.066 & 0.508 & 0.072 & 0.113 & 1.000 \\
DeepSeek-V3.2 & 0.590 & 0.407 & 0.103 & 0.079 & 0.073 & 0.427 & 0.077 & 0.099 & 1.000 \\
Qwen-Turbo & 0.269 & 0.180 & 0.029 & 0.058 & 0.072 & 0.202 & 0.047 & 0.017 & 1.000 \\
Qwen2.5-7B & -0.266 & -0.031 & 0.006 & 0.042 & 0.051 & -0.094 & 0.031 & -0.006 & 1.000 \\
Qwen-Max & 0.342 & 0.240 & -0.041 & -0.002 & 0.003 & -0.062 & -0.025 & -0.040 & 1.000 \\
Qwen-Plus & 0.453 & -0.247 & 0.076 & 0.061 & 0.062 & 0.295 & 0.070 & 0.069 & 1.000 \\
Qwen3.5-27B & 0.751 & 0.061 & 0.115 & 0.073 & 0.067 & 0.507 & 0.075 & 0.112 & 1.000 \\
GLM-5.1 & 0.115 & -0.190 & 0.050 & 0.044 & 0.047 & 0.347 & 0.052 & 0.048 & 1.000 \\
GPT-5.4-mini & 0.263 & 0.208 & 0.018 & 0.023 & 0.017 & 0.062 & 0.019 & 0.016 & 1.000 \\
GPT-5.4 & 0.434 & 0.335 & 0.016 & 0.022 & 0.025 & 0.131 & 0.017 & 0.014 & 1.000 \\
GPT-5.5 & 0.670 & 0.637 & 0.099 & 0.076 & 0.081 & 0.535 & 0.073 & 0.091 & 1.000 \\
Claude-Opus-4.7 & 0.381 & 0.296 & 0.100 & 0.073 & 0.064 & 0.464 & 0.074 & 0.091 & 1.000 \\
Claude-Sonnet-4.6 & 0.549 & 0.407 & 0.073 & 0.051 & 0.044 & 0.482 & 0.066 & 0.067 & 1.000 \\
Claude-Opus-4.6 & 0.477 & 0.435 & 0.130 & 0.087 & 0.068 & 0.495 & 0.082 & 0.123 & 1.000 \\
DeepSeek-V4-Pro & -0.561 & -0.338 & 0.081 & 0.063 & 0.068 & 0.498 & 0.068 & 0.077 & 1.000 \\
Qwen3.6-Plus & 0.316 & 0.565 & 0.066 & 0.061 & 0.068 & 0.459 & 0.059 & 0.058 & 1.000 \\\bottomrule\end{tabular}}
\caption{Baseline-adjusted CBF on the main session target.}
\label{tab:adjusted_cbf}
\end{table*}

\subsection{Normalized Hallucination Rate}
\label{app:norm_hr}

We define normalized HR as $\Pr(c>.7\mid y=0)=\mathrm{HR}/\Pr(y=0)$. It is the fraction of errors assigned high confidence; a value of 1 means every error is high-confidence.

\begin{table}[tbp]\centering\fontsize{9}{10.8}\selectfont\setlength{\tabcolsep}{4pt}

\fitTable{\begin{tabular}{lrrr}\toprule
Model & Error rate & HR & Normalized HR \\\midrule
Mistral-7B & 0.584 & 0.583 & 0.999 \\
Llama-3-8B & 0.423 & 0.423 & 1.000 \\
Qwen-Turbo & 0.245 & 0.230 & 0.939 \\
Qwen-Max & 0.212 & 0.209 & 0.990 \\\bottomrule\end{tabular}}
\caption{High-confidence error prevalence on valid answers.}
\label{tab:norm_hr_table}
\end{table}

This conditional error fraction does not alone determine confidence discrimination: comparing confidence on correct answers is also necessary. Reliability curves in Figure~\ref{fig:reliability} provide the complementary view.

\subsection{CBF vs.\ Rank Correlation}
\label{app:cbf_rank}
\label{app:tau}

\begin{figure}[!htbp]
\centering
\includegraphics[width=\columnwidth]{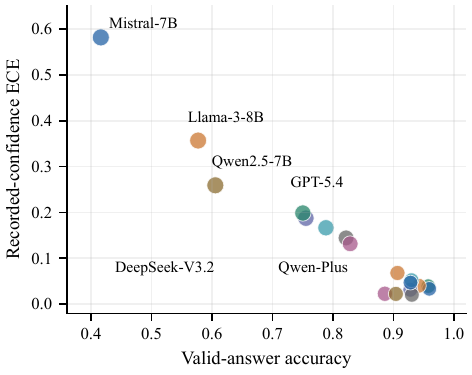}
\caption{\textbf{Accuracy, confidence calibration, and domain-profile error.} Answer accuracy versus ECE; bubble area represents NoMemory CBF.}
\label{fig:bubble}
\end{figure}

A natural alternative to CBF is Kendall's $\tau$ over domain accuracy rankings. However, $\tau$ is undefined when the baseline produces constant predictions; e.g., NoMemory estimates 0.5 for all domains. CBF avoids this by measuring absolute deviation rather than rank correlation, making it applicable to all baselines including the constant-prediction case. We recommend reporting both CBF and $\tau$ when applicable. Figure~\ref{fig:bubble} visualizes the three-way relationship between MMLU accuracy, recorded-confidence ECE, and NoMemory CBF across the full $n{=}18$ cross-baseline panel, illustrating why CBF and ECE are non-redundant.

\begin{table*}[tbp]\centering\fontsize{9}{10.8}\selectfont\setlength{\tabcolsep}{4pt}

\fitTable{\begin{tabular}{lrrrrrr}\toprule
Model & L1 bias & L1 shape & Global bias & Global shape & GLAD-style bias & GLAD-style shape \\\midrule
Mistral-7B & 0.204 & 0.138 & 0.184 & 0.118 & 0.217 & 0.226 \\
Llama-3-8B & -0.011 & 0.129 & 0.107 & 0.130 & 0.164 & 0.197 \\
GLM-5 & -0.032 & 0.034 & -0.060 & 0.041 & -0.014 & 0.098 \\
DeepSeek-V3.2 & -0.026 & 0.073 & -0.006 & 0.074 & 0.063 & 0.093 \\
Qwen-Turbo & 0.171 & 0.097 & 0.004 & 0.104 & 0.178 & 0.112 \\
Qwen2.5-7B & 0.245 & 0.134 & 0.100 & 0.129 & 0.229 & 0.167 \\
Qwen-Max & 0.080 & 0.096 & -0.268 & 0.071 & -0.231 & 0.340 \\
Qwen-Plus & 0.094 & 0.120 & -0.041 & 0.101 & 0.106 & 0.107 \\
Qwen3.5-27B & -0.054 & 0.047 & -0.077 & 0.033 & 0.003 & 0.060 \\
GLM-5.1 & -0.381 & 0.133 & -0.118 & 0.050 & 0.014 & 0.077 \\
GPT-5.4-mini & 0.070 & 0.148 & -0.031 & 0.132 & 0.044 & 0.167 \\
GPT-5.4 & 0.010 & 0.104 & -0.023 & 0.087 & 0.030 & 0.136 \\
GPT-5.5 & -0.056 & 0.042 & -0.056 & 0.051 & 0.039 & 0.076 \\
Claude-Opus-4.7 & -0.220 & 0.075 & -0.077 & 0.059 & 0.012 & 0.104 \\
Claude-Sonnet-4.6 & -0.156 & 0.050 & -0.087 & 0.049 & -0.013 & 0.088 \\
Claude-Opus-4.6 & -0.200 & 0.089 & -0.071 & 0.061 & 0.026 & 0.080 \\
DeepSeek-V4-Pro & -0.680 & 0.056 & -0.084 & 0.056 & 0.035 & 0.076 \\
Qwen3.6-Plus & -0.241 & 0.091 & -0.087 & 0.070 & 0.043 & 0.085 \\\bottomrule\end{tabular}}
\caption{Signed bias and centered domain-shape MAE on the pooled target; lower shape error is better.}
\label{tab:signed_shape}
\end{table*}

\begin{figure*}[tbp]\centering\includegraphics[width=.90\textwidth]{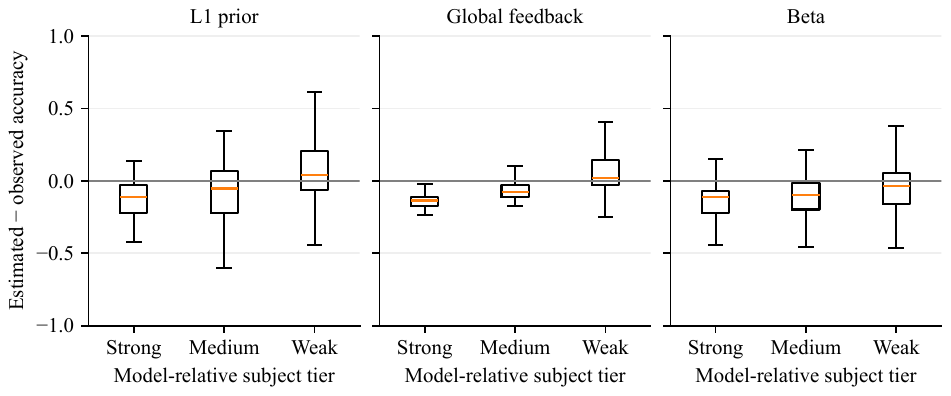}\caption{Signed domain errors for L1, Global, and Beta across all 18 models on pooled targets. Each model contributes six strong, seven medium, and seven weak subjects ranked by observed accuracy. Boxes show quartiles and medians; whiskers extend to 1.5 IQR. Positive errors denote overestimation.}\label{fig:signed_domains}\end{figure*}

\subsection{Cross-Baseline Bootstrap Significance}
\label{app:cross_boot}

Paired question-cluster intervals are used for the matched pooled comparisons in Table~\ref{tab:paired_shape}. All estimators are refitted within each of 400 resamples. The intervals describe uncertainty conditional on the recorded questions; no blanket significance threshold is assigned across models, metrics, and baselines.

\begin{table*}[tbp]\centering\fontsize{9}{10.8}\selectfont\setlength{\tabcolsep}{4pt}

\fitTable{\begin{tabular}{lrrr}\toprule
Model & DS$-$Beta & GLAD-style$-$Beta & Global / Beta / GLAD-style shape MAE \\\midrule
Qwen2.5-7B & -0.083 [-0.116, -0.038] & -0.151 [-0.170, -0.084] & 0.129 / 0.120 / 0.167 \\
GPT-5.4-mini & -0.010 [-0.125, +0.008] & -0.048 [-0.108, -0.006] & 0.132 / 0.105 / 0.167 \\
GPT-5.4 & -0.071 [-0.157, +0.018] & +0.006 [-0.063, +0.035] & 0.087 / 0.107 / 0.136 \\
GPT-5.5 & -0.056 [-0.075, +0.034] & +0.036 [+0.011, +0.092] & 0.051 / 0.089 / 0.076 \\
DeepSeek-V4-Pro & -0.041 [-0.086, +0.051] & +0.053 [+0.014, +0.098] & 0.056 / 0.121 / 0.076 \\\bottomrule\end{tabular}}
\caption{Pooled paired CBF differences and domain-shape error under strict decoding. Brackets give question-clustered intervals from 400 refitted resamples; lower shape error is better. Global is a constant profile.}
\label{tab:paired_shape}
\end{table*}

\section{Robustness Ablations}
\label{app:group_e}

\subsection{Additional Robustness Ablations}
\label{app:friend_or_foe}
\label{app:gpt55_ablation}
\label{app:detail_ablations}

This appendix contains additional ablation experiments orthogonal to the main diagnostic ladder.

\paragraph{A1: Information and estimator controls.} Tables~\ref{tab:main} and~\ref{tab:matched_l2} compare the common endpoint target. UF-PDC, Platt, and HistBin have identical domain outputs, while their query-confidence interfaces differ.

\paragraph{A2: Feedback noise and coverage.} Appendix~\ref{app:noise_sessions} reports five noise channels at 12.5\%, 50\%, and 100\% coverage for five models, with 100 retained repetitions per condition.

\paragraph{A3: Session length.} Prefix estimates are scored against the fixed end-of-session target, allowing the effect of additional feedback to be separated from changes in the target itself. Figure~\ref{fig:learning_curve} follows this comparison over the 50-turn sessions.

\paragraph{A4: Prior assessment on MMLU-Pro.}\label{app:heldout}
For each of eight frontier models, we sampled 15 questions in each of ten MMLU-Pro categories (150 questions/model) and queried domain priors on the same category names. Table~\ref{tab:mmlupro_categories} reports actual and estimated accuracy. These categories broaden the evaluation beyond the primary question pool.

\begin{table*}[tbp]
\centering\fontsize{9}{10.8}\selectfont

\setlength{\tabcolsep}{2pt}

\fitTable{%
\begin{tabular}{l rr rr rr rr rr rr rr rr}
\toprule
& \multicolumn{2}{c}{\textbf{GPT-5.4}} & \multicolumn{2}{c}{\textbf{GPT-5.4-mini}} & \multicolumn{2}{c}{\textbf{GPT-5.5}} & \multicolumn{2}{c}{\textbf{Cl-Op-4-6}} & \multicolumn{2}{c}{\textbf{Cl-So-4-6}} & \multicolumn{2}{c}{\textbf{Cl-Op-4-7}} & \multicolumn{2}{c}{\textbf{DSv4-Pro}} & \multicolumn{2}{c}{\textbf{Q3.6-Plus}} \\
\cmidrule(lr){2-3} \cmidrule(lr){4-5} \cmidrule(lr){6-7} \cmidrule(lr){8-9} \cmidrule(lr){10-11} \cmidrule(lr){12-13} \cmidrule(lr){14-15} \cmidrule(lr){16-17}
Category & acc & ZS & acc & ZS & acc & ZS & acc & ZS & acc & ZS & acc & ZS & acc & ZS & acc & ZS \\
\midrule
math         & 0.40 & 0.85 & 0.33 & 0.78 & 0.93 & 0.72 & 0.80 & 0.78 & 0.93 & 0.72 & 0.87 & 0.78 & 0.93 & 0.90 & 0.93 & 0.85 \\
physics      & 0.40 & 0.84 & 0.47 & 0.72 & 1.00 & 0.70 & 1.00 & 0.80 & 1.00 & 0.74 & 0.73 & 0.80 & 1.00 & 0.95 & 1.00 & 0.90 \\
chemistry    & 0.60 & 0.82 & 0.20 & 0.70 & 0.87 & 0.72 & 0.93 & 0.75 & 0.92 & 0.73 & 0.73 & 0.72 & 0.87 & 0.90 & 0.87 & 0.82 \\
law          & 0.60 & 0.81 & 0.53 & 0.74 & 0.80 & 0.76 & 0.73 & 0.70 & 0.73 & 0.67 & 0.80 & 0.60 & 0.79 & 0.75 & 0.73 & 0.80 \\
engineering  & 0.53 & 0.80 & 0.47 & 0.73 & 0.80 & 0.68 & 0.93 & 0.72 & 1.00 & 0.70 & 0.67 & 0.65 & 0.85 & 0.85 & 0.93 & 0.85 \\
economics    & 0.80 & 0.83 & 0.73 & 0.79 & 0.93 & 0.78 & 1.00 & 0.84 & 0.87 & 0.76 & 0.87 & 0.82 & 0.93 & 0.80 & 0.93 & 0.90 \\
health       & 0.67 & 0.86 & 0.73 & 0.77 & 0.87 & 0.76 & 0.79 & 0.79 & 0.80 & 0.75 & 0.71 & 0.72 & 0.87 & 0.90 & 0.87 & 0.90 \\
psychology   & 0.80 & 0.88 & 0.73 & 0.82 & 0.80 & 0.82 & 0.87 & 0.87 & 0.87 & 0.78 & 0.73 & 0.85 & 0.93 & 0.85 & 0.80 & 0.85 \\
business     & 0.47 & 0.84 & 0.20 & 0.80 & 0.80 & 0.82 & 0.93 & 0.81 & 0.80 & 0.75 & 0.80 & 0.78 & 0.73 & 0.90 & 0.87 & 0.90 \\
history      & 0.73 & 0.87 & 0.67 & 0.84 & 0.80 & 0.80 & 0.80 & 0.76 & 0.73 & 0.79 & 0.73 & 0.75 & 0.73 & 0.85 & 0.80 & 0.90 \\
\midrule
mean         & 0.60 & 0.84 & 0.51 & 0.77 & 0.86 & 0.76 & 0.88 & 0.78 & 0.86 & 0.74 & 0.76 & 0.75 & 0.86 & 0.86 & 0.87 & 0.87 \\
\midrule
unparsed/150 & \multicolumn{2}{c}{0} & \multicolumn{2}{c}{0} & \multicolumn{2}{c}{0} & \multicolumn{2}{c}{1} & \multicolumn{2}{c}{8} & \multicolumn{2}{c}{1} & \multicolumn{2}{c}{3} & \multicolumn{2}{c}{0} \\
\midrule
\textbf{CBF} & \multicolumn{2}{c}{\textbf{0.760}} & \multicolumn{2}{c}{\textbf{0.738}} & \multicolumn{2}{c}{\textbf{0.888}} & \multicolumn{2}{c}{\textbf{0.902}} & \multicolumn{2}{c}{\textbf{0.863}} & \multicolumn{2}{c}{\textbf{0.941}} & \multicolumn{2}{c}{\textbf{0.931}} & \multicolumn{2}{c}{\textbf{0.937}} \\
\bottomrule
\end{tabular}}
\caption{\textbf{MMLU-Pro per-category accuracy and ZS-SK estimates across 8 frontier models} (15 questions per category $\times$ 10 categories); bottom row = per-model CBF on this held-out set. Cl-Op = Claude-Opus, Cl-So = Claude-Sonnet, DSv4-Pro = DeepSeek-V4-Pro, Q3.6 = Qwen3.6.}
\label{tab:mmlupro_categories}
\end{table*}

Mean MMLU-Pro CBF is 0.870 across the eight models. GPT-5.5 and Claude-Opus-4.7 report conservative estimates on several categories, whereas individual categories exhibit larger errors. The table therefore supports a subject-level description beyond the primary MMLU pool.

\paragraph{A4b: Prior assessment on BBH.}\label{app:heldout_bbh}
For the same eight models, we sampled 15 questions from each of ten BIG-Bench Hard tasks~\citep{suzgun2022challenging} and queried priors on those task names. Table~\ref{tab:bbh_tasks} reports the complete task-level comparison. The stored scorer accepts a target string or its cleaned form when it occurs in the generated text.

\begin{table*}[tbp]
\centering\fontsize{9}{10.8}\selectfont

\setlength{\tabcolsep}{2pt}

\fitTable{%
\begin{tabular}{l cc cc cc cc cc cc cc cc}
\toprule
& \multicolumn{2}{c}{\textbf{GPT-5.4}} & \multicolumn{2}{c}{\textbf{GPT-5.4-mini}} & \multicolumn{2}{c}{\textbf{GPT-5.5}} & \multicolumn{2}{c}{\textbf{Cl-Op-4-6}} & \multicolumn{2}{c}{\textbf{Cl-So-4-6}} & \multicolumn{2}{c}{\textbf{Cl-Op-4-7}} & \multicolumn{2}{c}{\textbf{DSv4-Pro}} & \multicolumn{2}{c}{\textbf{Q3.6-Plus}} \\
\cmidrule(lr){2-3} \cmidrule(lr){4-5} \cmidrule(lr){6-7} \cmidrule(lr){8-9} \cmidrule(lr){10-11} \cmidrule(lr){12-13} \cmidrule(lr){14-15} \cmidrule(lr){16-17}
\textbf{Task} & $p^*$ & ZS & $p^*$ & ZS & $p^*$ & ZS & $p^*$ & ZS & $p^*$ & ZS & $p^*$ & ZS & $p^*$ & ZS & $p^*$ & ZS \\
\midrule
boolean\_expressions  & 0.87 & 0.94 & 0.87 & 0.82 & 1.00 & 0.95 & 0.93 & 0.96 & 1.00 & 0.88 & 0.87 & 0.95 & 1.00 & 0.85 & 1.00 & 0.95 \\
causal\_judgement     & 0.67 & 0.83 & 0.67 & 0.78 & 0.67 & 0.85 & 0.53 & 0.72 & 0.73 & 0.60 & 0.80 & 0.65 & 0.73 & 0.80 & 0.73 & 0.70 \\
date\_understanding   & 0.27 & 0.90 & 0.20 & 0.75 & 0.47 & 0.85 & 0.93 & 0.92 & 1.00 & 0.82 & 0.87 & 0.85 & 0.80 & 0.93 & 0.87 & 0.85 \\
logical\_deduction\_5 & 1.00 & 0.86 & 0.67 & 0.70 & 1.00 & 0.90 & 1.00 & 0.88 & 1.00 & 0.65 & 1.00 & 0.90 & 0.53 & 0.85 & 1.00 & 0.75 \\
navigate              & 0.80 & 0.88 & 0.53 & 0.68 & 1.00 & 0.85 & 0.93 & 0.95 & 1.00 & 0.76 & 0.93 & 0.90 & 1.00 & 0.90 & 1.00 & 0.70 \\
object\_counting      & 0.73 & 0.96 & 0.53 & 0.90 & 1.00 & 0.90 & 0.80 & 0.97 & 1.00 & 0.84 & 0.60 & 0.95 & 1.00 & 0.80 & 1.00 & 0.95 \\
ruin\_names           & 0.67 & 0.79 & 0.67 & 0.72 & 0.93 & 0.75 & 0.87 & 0.86 & 0.87 & 0.56 & 0.93 & 0.75 & 0.73 & 0.70 & 0.93 & 0.80 \\
sports\_understanding & 1.00 & 0.84 & 0.93 & 0.74 & 0.80 & 0.90 & 0.93 & 0.96 & 1.00 & 0.90 & 0.87 & 0.90 & 0.60 & 0.65 & 1.00 & 0.95 \\
tracking\_shuffled\_3 & 0.67 & 0.95 & 0.67 & 0.88 & 1.00 & 0.85 & 1.00 & 0.98 & 1.00 & 0.73 & 0.53 & 0.95 & 0.87 & 0.90 & 1.00 & 0.95 \\
word\_sorting         & 0.67 & 0.97 & 0.47 & 0.95 & 1.00 & 0.90 & 0.13 & 0.94 & 0.00 & 0.80 & 0.93 & 0.90 & 0.33 & 0.90 & 0.87 & 0.92 \\
\midrule
\textbf{CBF}          & \multicolumn{2}{c}{\textbf{0.781}} & \multicolumn{2}{c}{\textbf{0.780}} & \multicolumn{2}{c}{\textbf{0.850}} & \multicolumn{2}{c}{\textbf{0.861}} & \multicolumn{2}{c}{\textbf{0.734}} & \multicolumn{2}{c}{\textbf{0.860}} & \multicolumn{2}{c}{\textbf{0.835}} & \multicolumn{2}{c}{\textbf{0.901}} \\
\bottomrule
\end{tabular}}
\caption{\textbf{BBH per-task accuracy ($p^*$) and ZS-SK estimates across 8 frontier models} (15 questions per task $\times$ 10 BBH tasks); BBH has no subject-name overlap with MMLU. Bottom row = per-model CBF on this held-out set. Cl-Op = Claude-Opus, Cl-So = Claude-Sonnet, DSv4-Pro = DeepSeek-V4-Pro, Q3.6 = Qwen3.6.}
\label{tab:bbh_tasks}
\end{table*}

Mean BBH CBF is 0.825 across the same eight models, compared with 0.870 on MMLU-Pro. The full task table reveals errors obscured by these averages: for example, Claude-Sonnet-4.6 estimates .80 on word sorting but obtains .00. The following task comparisons describe this fixed evaluation sample.

Together, MMLU-Pro and BBH show that prior-profile errors vary with task content. Their distinct task labels broaden the evaluation, but public benchmark differences do not identify whether the underlying tasks were seen during training.

\paragraph{A5: Reliability estimators.}\label{app:ds_glad}
We compare adaptations of two established crowdsourcing models on the same decoded-label records.

\textbf{Asymmetric Dawid--Skene}: relaxes EM-Joint's symmetric reliability to per-state confusion matrices. For each rater state $s \in \{\text{in},\text{out}\}$, we estimate TPR$_s = P(\text{decoded}=1 \mid \text{correct}=1, s)$ and FPR$_s = P(\text{decoded}=1 \mid \text{correct}=0, s)$. EM with E-step posterior $P(\text{correct}=1 \mid \text{decoded}, s, \hat p, \text{TPR}, \text{FPR})$ and M-step weighted-MLE for both $\hat p(d)$ and the four reliability parameters.

\textbf{GLAD}-style: per-domain difficulty $\delta_d$ and per-state ability $\alpha_s$, with reliability $\sigma(\alpha_s - \delta_d)$ (logistic). EM updates over $\hat p(d)$, $\delta_d$, $\alpha_{\text{in}}$, $\alpha_{\text{out}}$.

\begin{table*}[tbp]\centering\fontsize{9}{10.8}\selectfont\setlength{\tabcolsep}{4pt}

\fitTable{\begin{tabular}{lcccc}\toprule
Model & EM-Joint & DS & GLAD-style & L1 \\\midrule
Mistral-7B & 0.748/0.755 & 0.742/0.799 & 0.731/0.712 & 0.677 \\
Llama-3-8B & 0.748/0.796 & 0.748/0.868 & 0.734/0.730 & 0.734 \\
GLM-5 & 0.572/0.909 & 0.575/0.883 & 0.594/0.908 & 0.904 \\
DeepSeek-V3.2 & 0.626/0.900 & 0.627/0.845 & 0.636/0.885 & 0.834 \\
Qwen-Turbo & 0.711/0.896 & 0.703/0.813 & 0.694/0.793 & 0.772 \\
Qwen2.5-7B & 0.753/0.835 & 0.748/0.791 & 0.738/0.724 & 0.670 \\
Qwen-Max & 0.680/0.711 & 0.672/0.676 & 0.667/0.643 & 0.789 \\
Qwen-Plus & 0.667/0.902 & 0.670/0.871 & 0.670/0.849 & 0.806 \\
Qwen3.5-27B & 0.573/0.909 & 0.577/0.820 & 0.594/0.940 & 0.886 \\
GLM-5.1 & 0.578/0.873 & 0.580/0.799 & 0.578/0.919 & 0.608 \\
GPT-5.4-mini & 0.686/0.895 & 0.686/0.849 & 0.685/0.811 & 0.764 \\
GPT-5.4 & 0.651/0.879 & 0.648/0.781 & 0.647/0.857 & 0.797 \\
GPT-5.5 & 0.605/0.938 & 0.601/0.818 & 0.609/0.909 & 0.858 \\
Claude-Opus-4.7 & 0.609/0.880 & 0.613/0.830 & 0.621/0.890 & 0.742 \\
Claude-Sonnet-4.6 & 0.579/0.844 & 0.589/0.838 & 0.590/0.915 & 0.801 \\
Claude-Opus-4.6 & 0.592/0.876 & 0.599/0.833 & 0.616/0.917 & 0.771 \\
DeepSeek-V4-Pro & 0.592/0.920 & 0.592/0.819 & 0.596/0.913 & 0.316 \\
Qwen3.6-Plus & 0.616/0.928 & 0.613/0.847 & 0.612/0.903 & 0.719 \\\bottomrule\end{tabular}}
\caption{Session/pooled CBF of reliability-aware methods; L1 uses the session target.}
\label{tab:second_order_full}
\end{table*}

DS and GLAD-style are adaptations of established crowdsourcing models~\citep{dawid1979maximum,whitehill2009glad}. DS fits state-specific true/false positive rates with 50 EM iterations, initialized at $(.75,.25)$ in-expertise and $(.55,.45)$ outside expertise. GLAD-style uses $r_{hd}=\sigma(\alpha_h-\delta_d)$, initialized at $\alpha=(1,1),\delta=0,p_d=.5$, with 80 generalized-EM iterations and step size .05. Ability/difficulty are clipped to $[-3,3]$. This additive parameterization differs from original GLAD. The same observed expertise state is supplied to both methods. Their gains depend on initialization and the information available in a single-label setting.

\paragraph{Likelihoods and updates.}
Dawid--Skene (DS)~\citep{dawid1979maximum} models asymmetric feedback errors. Let $h_t\in\{\mathrm{in},\mathrm{out}\}$ be the assigned expertise state, available to DS and GLAD-style, and let $\theta_{h1}$ and $\theta_{h0}$ be the true- and false-positive rates. Define $L_{tk}=\theta_{h_tk}^{z_t}(1-\theta_{h_tk})^{1-z_t}$. On observed feedback, the likelihood is
\begin{equation}
 \ell(p,\theta)=\sum_{t:z_t\ne\bot}
 \log\left[p_{d_t}L_{t1}+(1-p_{d_t})L_{t0}\right].
 \label{eq:likelihood}
\end{equation}
EM alternates posterior correctness and domain-prevalence updates:
\begin{align}
 \rho_t&=\frac{p_{d_t}L_{t1}}{p_{d_t}L_{t1}+(1-p_{d_t})L_{t0}},\label{eq:posterior}\\
 p_d&\leftarrow\frac{1}{M_d}\sum_{t\in\mathcal O_d}\rho_t.\label{eq:em_profile}
\end{align}
The confusion rates are updated from posterior-weighted label counts. Verified $y_t$ is excluded from these fits.

The evaluated GLAD-style variant uses ability $\alpha_h$ and domain difficulty $\delta_d$ to impose a symmetric channel:
\begin{equation}
 \begin{aligned}
 r_{hd}&=\sigma(\alpha_h-\delta_d),\\
 \theta_{hd,1}&=r_{hd},\qquad\theta_{hd,0}=1-r_{hd}.
 \end{aligned}
 \label{eq:glad}
\end{equation}
It shares the posterior/profile updates, with gradient updates for $\alpha$ and $\delta$. This additive parameterization is a variant of GLAD~\citep{whitehill2009glad}. DS uses 50 EM iterations; GLAD-style uses 80 generalized-EM iterations. Domain methods retain .5 when $M_d=0$; initialization and update parameters are given in Appendix~\ref{app:baselines}.

\paragraph{Ability initialization.} Both expertise states start at $\alpha_h=1$, with $\delta_d=0$ and $p_d=.5$, assuming positively oriented feedback in each state. This fixed initialization is used for every model, coverage, noise channel, and partition. Original GLAD~\citep{whitehill2009glad} uses $\sigma(\alpha_i\beta_j)$ and a positive ability prior. Our pair $(1,1)$ instead denotes two abilities in an additive adaptation. Table~\ref{tab:glad_init} compares this default with a mixed-sign sensitivity condition. The latter can orient a domain toward the complementary label, so its low clean-label CBF does not establish that reliable feedback is unhelpful.

\begin{table}[!htbp]\centering\fontsize{9}{10.8}\selectfont\setlength{\tabcolsep}{3pt}

\fitTable{\begin{tabular}{lrrrr}\toprule
Model & \makecell{Clean\\$(1,1)$} & \makecell{Clean\\$(1,-.3)$} & \makecell{30\%\\$(1,1)$} & \makecell{30\%\\$(1,-.3)$} \\\midrule
Qwen2.5-7B & 0.854 & 0.563 & 0.924 & 0.706 \\
GPT-5.4-mini & 0.879 & 0.411 & 0.922 & 0.563 \\
GPT-5.4 & 0.880 & 0.248 & 0.920 & 0.482 \\
GPT-5.5 & 0.934 & 0.127 & 0.912 & 0.296 \\
DeepSeek-V4-Pro & 0.938 & 0.080 & 0.911 & 0.283 \\\bottomrule\end{tabular}}
\caption{GLAD-style pooled CBF under two fixed ability initializations; 100 repetitions. The positive pair is the reported default.}
\label{tab:glad_init}
\end{table}

\paragraph{Optimization sensitivity.} The fixed 50-step DS and 80-step GLAD-style fits are finite-budget references. Table~\ref{tab:convergence} compares them with likelihood optimization on the same pooled labels and parameter bounds. Clean labels are deterministic; each 15\% and 30\% noise condition uses the same 100 saved draws. L-BFGS-B, followed by SLSQP when needed, stops at projected-gradient norm $\leq10^{-6}$, with a 20,000-iteration cap per optimizer. 2010/2010 fits meet this criterion; the largest norm is $1.01\times10^{-7}$. The per-label log-likelihood gains on clean labels are 0.000000--0.000061 for DS and 0.003--0.049 for GLAD-style. Final state reliabilities, domain parameters, and likelihood trajectories accompany the analysis materials.

\begin{table*}[tbp]\centering\fontsize{9}{10.8}\selectfont\setlength{\tabcolsep}{4pt}

\fitTable{\begin{tabular}{lrrrrrr}\toprule
Model & Noise & Beta & DS fixed & DS opt. & GLAD fixed & GLAD opt. \\\midrule
Qwen2.5-7B & 0 & 0.997 & 0.919 & 0.920 & 0.854 & 0.920 \\
Qwen2.5-7B & 15 & 0.943 & 0.906 & 0.906 & 0.909 & 0.909 \\
Qwen2.5-7B & 30 & 0.895 & 0.886 & 0.885 & 0.924 & 0.874 \\
GPT-5.4-mini & 0 & 0.996 & 0.838 & 0.838 & 0.879 & 0.961 \\
GPT-5.4-mini & 15 & 0.917 & 0.797 & 0.797 & 0.908 & 0.910 \\
GPT-5.4-mini & 30 & 0.843 & 0.772 & 0.771 & 0.922 & 0.849 \\
GPT-5.4 & 0 & 0.994 & 0.765 & 0.716 & 0.880 & 0.942 \\
GPT-5.4 & 15 & 0.897 & 0.723 & 0.719 & 0.917 & 0.915 \\
GPT-5.4 & 30 & 0.800 & 0.697 & 0.696 & 0.920 & 0.843 \\
GPT-5.5 & 0 & 0.992 & 0.713 & 0.702 & 0.934 & 0.982 \\
GPT-5.5 & 15 & 0.867 & 0.633 & 0.631 & 0.947 & 0.904 \\
GPT-5.5 & 30 & 0.741 & 0.606 & 0.597 & 0.912 & 0.800 \\
DeepSeek-V4-Pro & 0 & 0.991 & 0.765 & 0.765 & 0.938 & 0.986 \\
DeepSeek-V4-Pro & 15 & 0.865 & 0.639 & 0.632 & 0.952 & 0.907 \\
DeepSeek-V4-Pro & 30 & 0.739 & 0.607 & 0.592 & 0.911 & 0.805 \\\bottomrule\end{tabular}}
\caption{Pooled CBF with complete coverage. Fixed denotes the reported iteration budget; Opt. denotes likelihood optimization to the stated stationary-point criterion. No outcome labels enter either fit. Noise columns are percentages; noisy scores average 100 draws.}
\label{tab:convergence}
\end{table*}

Optimization raises clean-label GLAD-style CBF from .854--.938 to 0.920--0.986. Thus the fixed-budget score increase after adding noise does not establish a beneficial noise effect. DS improves its likelihood while retaining substantial calibration error. Its marginal observation model is $q_{hd}=f_h+p_d(t_h-f_h)$. An affine transformation of $p_d$ can be offset by a transformation of $t_h,f_h$ while preserving $q_{hd}$ whenever the parameters remain within their bounds. More single labels therefore need not uniquely determine prevalence and reliability. The pooled result cannot be explained solely by short sessions: both optimization budget and identifiability matter.

\paragraph{A6: Feedback decoding and coverage.}\label{app:decoder_ablation}
We compare domain counts under three channels: decoded observed labels, verified labels at those same observed positions, and all verified labels. Holding the positions fixed isolates label correction; adding the remaining positions isolates coverage.

\begin{table*}[tbp]\centering\fontsize{9}{10.8}\selectfont\setlength{\tabcolsep}{4pt}

\fitTable{\begin{tabular}{lrrrrr}\toprule
Model & Coverage & Decoded & Corrected positions & Complete & $\Delta$Acc (pp) \\\midrule
Mistral-7B & 11.1\% & 0.809 & 0.834 & 1.000 & +14.9 \\
Llama-3-8B & 13.4\% & 0.852 & 0.848 & 1.000 & +10.2 \\
GLM-5 & 14.8\% & 0.862 & 0.963 & 1.000 & +2.6 \\
DeepSeek-V3.2 & 15.2\% & 0.849 & 0.898 & 1.000 & +5.3 \\
Qwen-Turbo & 14.7\% & 0.892 & 0.870 & 1.000 & +7.4 \\
Qwen2.5-7B & 12.1\% & 0.846 & 0.822 & 1.000 & +17.0 \\
Qwen-Max & 6.7\% & 0.710 & 0.688 & 1.000 & -28.8 \\
Qwen-Plus & 13.8\% & 0.878 & 0.886 & 1.000 & +5.5 \\
Qwen3.5-27B & 15.5\% & 0.888 & 0.950 & 1.000 & +1.8 \\
GLM-5.1 & 12.2\% & 0.864 & 0.941 & 1.000 & +2.1 \\
GPT-5.4-mini & 4.9\% & 0.820 & 0.848 & 1.000 & -1.3 \\
GPT-5.4 & 5.9\% & 0.839 & 0.860 & 1.000 & +1.5 \\
GPT-5.5 & 15.6\% & 0.888 & 0.932 & 1.000 & +1.3 \\
Claude-Opus-4.7 & 14.5\% & 0.847 & 0.918 & 1.000 & +2.5 \\
Claude-Sonnet-4.6 & 13.0\% & 0.872 & 0.915 & 1.000 & +1.1 \\
Claude-Opus-4.6 & 15.0\% & 0.877 & 0.925 & 1.000 & +4.1 \\
DeepSeek-V4-Pro & 13.7\% & 0.870 & 0.924 & 1.000 & +3.6 \\
Qwen3.6-Plus & 15.0\% & 0.868 & 0.926 & 1.000 & +2.5 \\\bottomrule\end{tabular}}
\caption{Pooled CBF of Count under matched feedback-channel interventions. $\Delta$Acc (pp) is 100 times accuracy on decoded positions minus accuracy on all valid answers. For Qwen-Max, 79/158 decoded positions are correct (50.0\%), versus 1867/2368 valid answers (78.8\%), yielding $-28.8$ pp.}
\label{tab:decoder_ablation}
\end{table*}

Decoded feedback selects which answers supply labels. For Qwen-Max, correct answers are less likely to yield a decoded label than incorrect answers, consistent with the confirmation misses in Table~\ref{tab:feedback_channel}. This produces a low-accuracy observed subset. Correcting its labels alone cannot remove selection; all valid-answer domains remain in the scoring target.

\paragraph{Semantic-reference comparison.} We sample 358 nonempty feedback utterances on valid answers, stratified by model, expertise state, and frozen decoder category, with at most three randomly selected utterances per nonempty stratum. One LLM annotator assigns positive, negative, uncertain, or neutral stance from the utterance and submitted answer. Estimates weight each sample by its stratum population divided by sample size. Precision is correctly decoded stance divided by all positive/negative decoder outputs; recall is correctly decoded stance divided by all positive/negative reference labels. Table~\ref{tab:decoder_semantics} reports all 18 models. Per-expertise estimates and annotated utterances accompany the analysis materials.

The expanded rules recognize \emph{accurate}, \emph{well done}, and \emph{spot on}, and distinguish answer rejection from questions about incorrect options. A statement of ``the correct answer is A is compared with the submitted option: agreement is confirmation and a different option is correction. Outcome keys are not used by the decoder. The frozen rules remain the primary protocol; the expanded rules define a separate sensitivity condition.

\begin{table*}[tbp]\centering\fontsize{9}{10.8}\selectfont\setlength{\tabcolsep}{4pt}

\fitTable{\begin{tabular}{lrrrrrrr}\toprule
Model & $n$ & P(F) & R(F) & P(E) & R(E) & Cov.(F) & Cov.(E) \\\midrule
Mistral-7B & 19 & 0.884 & 0.230 & 1.000 & 0.950 & 0.111 & 0.191 \\
Llama-3-8B & 18 & 0.852 & 0.741 & 1.000 & 0.800 & 0.134 & 0.175 \\
GLM-5 & 19 & 0.916 & 0.190 & 0.999 & 0.588 & 0.148 & 0.360 \\
DeepSeek-V3.2 & 21 & 0.939 & 0.813 & 0.997 & 0.943 & 0.152 & 0.360 \\
Qwen-Turbo & 21 & 0.893 & 0.297 & 0.969 & 0.365 & 0.147 & 0.354 \\
Qwen2.5-7B & 20 & 0.899 & 0.253 & 0.989 & 0.966 & 0.121 & 0.314 \\
Qwen-Max & 21 & 0.766 & 0.120 & 0.964 & 0.834 & 0.067 & 0.318 \\
Qwen-Plus & 21 & 0.900 & 0.293 & 0.998 & 0.959 & 0.138 & 0.346 \\
Qwen3.5-27B & 19 & 0.907 & 0.143 & 0.999 & 0.709 & 0.155 & 0.364 \\
GLM-5.1 & 18 & 0.874 & 0.189 & 1.000 & 0.991 & 0.122 & 0.281 \\
GPT-5.4-mini & 22 & 0.904 & 0.137 & 0.997 & 0.974 & 0.049 & 0.337 \\
GPT-5.4 & 23 & 0.909 & 0.167 & 0.992 & 0.998 & 0.059 & 0.347 \\
GPT-5.5 & 19 & 0.913 & 0.145 & 0.989 & 0.444 & 0.156 & 0.356 \\
Claude-Opus-4.7 & 21 & 0.858 & 0.287 & 0.989 & 0.352 & 0.145 & 0.304 \\
Claude-Sonnet-4.6 & 18 & 0.916 & 0.273 & 0.974 & 0.317 & 0.130 & 0.323 \\
Claude-Opus-4.6 & 18 & 0.901 & 0.312 & 1.000 & 0.974 & 0.150 & 0.315 \\
DeepSeek-V4-Pro & 19 & 0.881 & 0.171 & 0.999 & 0.583 & 0.137 & 0.335 \\
Qwen3.6-Plus & 21 & 0.877 & 0.301 & 0.984 & 0.964 & 0.150 & 0.345 \\\bottomrule\end{tabular}}
\caption{Stratified semantic-reference comparison. P/R are weighted stance precision/recall; Cov. is coverage of all valid answers. F = frozen, E = expanded. The same sampled utterances support rule inspection and reference comparison.}
\label{tab:decoder_semantics}
\end{table*}

Across the weighted sample, precision/recall are 0.890/0.224 for frozen rules and 0.992/0.728 for expanded rules. Applying expanded rules to the complete valid-answer stream changes Global\textquotesingle s session-level lead from 14/18 to 11/18 (Table~\ref{tab:decoder_sensitivity}). Count becomes the best matched control for Qwen-Turbo, Qwen-Plus, GPT-5.4-mini, and GPT-5.4. The sensitivity confirms that the observed channel depends on language interpretation as well as the simulator; a single decoder cannot establish a decoder-independent estimator ranking.

\begin{table*}[tbp]\centering\fontsize{9}{10.8}\selectfont\setlength{\tabcolsep}{4pt}

\fitTable{\begin{tabular}{lrrrrr}\toprule
Model & Global & Count & Beta & DS & GLAD \\\midrule
Mistral-7B & 0.590 & 0.634 & 0.711 & 0.727 & 0.633 \\
Llama-3-8B & 0.679 & 0.693 & 0.735 & 0.738 & 0.690 \\
GLM-5 & 0.888 & 0.803 & 0.667 & 0.601 & 0.796 \\
DeepSeek-V3.2 & 0.836 & 0.788 & 0.700 & 0.657 & 0.783 \\
Qwen-Turbo & 0.774 & 0.797 & 0.771 & 0.729 & 0.789 \\
Qwen2.5-7B & 0.697 & 0.754 & 0.776 & 0.763 & 0.748 \\
Qwen-Max & 0.798 & 0.785 & 0.755 & 0.718 & 0.782 \\
Qwen-Plus & 0.804 & 0.813 & 0.742 & 0.692 & 0.806 \\
Qwen3.5-27B & 0.891 & 0.807 & 0.669 & 0.601 & 0.801 \\
GLM-5.1 & 0.801 & 0.745 & 0.647 & 0.607 & 0.740 \\
GPT-5.4-mini & 0.721 & 0.727 & 0.726 & 0.698 & 0.725 \\
GPT-5.4 & 0.725 & 0.739 & 0.713 & 0.672 & 0.735 \\
GPT-5.5 & 0.881 & 0.804 & 0.690 & 0.633 & 0.796 \\
Claude-Opus-4.7 & 0.859 & 0.784 & 0.689 & 0.648 & 0.777 \\
Claude-Sonnet-4.6 & 0.871 & 0.780 & 0.666 & 0.617 & 0.773 \\
Claude-Opus-4.6 & 0.872 & 0.781 & 0.670 & 0.629 & 0.773 \\
DeepSeek-V4-Pro & 0.880 & 0.791 & 0.680 & 0.626 & 0.785 \\
Qwen3.6-Plus & 0.860 & 0.800 & 0.704 & 0.645 & 0.792 \\\bottomrule\end{tabular}}
\caption{Session-mean CBF under expanded feedback rules, on the same valid-answer targets and domain support as the primary comparison.}
\label{tab:decoder_sensitivity}
\end{table*}

Exact counting reaches CBF 1 with complete verified labels by construction. That result does not imply that EM or a model aggregating the same labels will recover the target exactly. The matched-input study and synthetic-noise grid evaluate those estimators separately.

\subsection{Learning Dynamics}
\label{app:learn_dyn}

\begin{figure}[!htbp]
\centering
\includegraphics[width=\columnwidth]{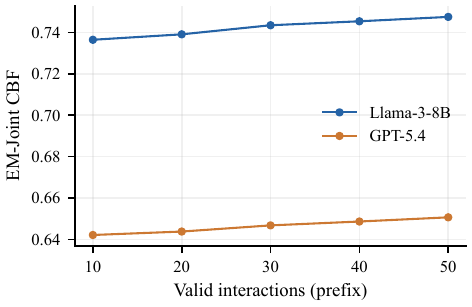}
\caption{\textbf{EM-Joint learning curves.} Prefixes of valid interactions are scored against the fixed final-session domain target.}
\label{fig:learning_curve}
\end{figure}

Figure~\ref{fig:learning_curve} follows EM-Joint estimates as more valid interactions become available. Each prefix is compared with the same final-session target. This distinguishes a changing estimate from a changing empirical reference.

\subsection{Prefix Estimation Curves}\label{app:lc_sat}

Table~\ref{tab:learn_sat} reports endpoint-target CBF at successive valid-turn prefixes. A session with fewer valid turns uses all of its available turns at later checkpoints. The final column matches the EM-Joint main-table score.

\begin{table}[tbp]\centering\fontsize{9}{10.8}\selectfont\setlength{\tabcolsep}{4pt}

\fitTable{\begin{tabular}{lrrrrr}\toprule
Model & 10 & 20 & 30 & 40 & 50 \\\midrule
Llama-3-8B & 0.736 & 0.739 & 0.743 & 0.745 & 0.748 \\
Qwen-Max & 0.680 & 0.679 & 0.679 & 0.680 & 0.680 \\
Qwen-Plus & 0.647 & 0.652 & 0.659 & 0.665 & 0.667 \\
GPT-5.4 & 0.642 & 0.644 & 0.647 & 0.649 & 0.651 \\
Claude-Opus-4.6 & 0.568 & 0.574 & 0.583 & 0.592 & 0.592 \\\bottomrule\end{tabular}}
\caption{EM-Joint CBF against a fixed final-session target at valid-turn prefixes.}
\label{tab:learn_sat}
\end{table}

These curves describe a 50-turn observation budget. They do not establish convergence with unlimited data; sparse informative feedback can leave domain estimates close to their initialization.

\subsection{Noise Sensitivity}
\label{app:noise}

We fix valid answers and apply five label-noise channels at 12.5\%, 50\%, and 100\% coverage on five models. The channels have false-positive/false-negative rates 0/0, .15/.15, .30/.30, .30/.10, and .10/.30. Each condition retains 100 repetitions and is scored per session and after pooling. The 12.5\% condition approximates the order of observed feedback coverage while using random rather than naturally selected missingness. Table~\ref{tab:noise_sens} gives full-coverage symmetric-noise results; Appendix~\ref{app:noise_sessions} contains the complete grid.

\begin{table*}[tbp]\centering\fontsize{9}{10.8}\selectfont\setlength{\tabcolsep}{4pt}

\fitTable{\begin{tabular}{llcccc}\toprule
Model & Noise & Global & Beta & DS & GLAD-style \\\midrule
Qwen2.5-7B & 0.00 & 0.749/0.871 & 0.895/0.997 & 0.977/0.919 & 0.934/0.854 \\
Qwen2.5-7B & 0.15 & 0.746/0.865 & 0.823/0.943 & 0.853/0.906 & 0.825/0.909 \\
Qwen2.5-7B & 0.30 & 0.739/0.854 & 0.764/0.895 & 0.757/0.886 & 0.727/0.924 \\
GPT-5.4-mini & 0.00 & 0.774/0.869 & 0.870/0.996 & 0.949/0.838 & 0.943/0.879 \\
GPT-5.4-mini & 0.15 & 0.756/0.854 & 0.795/0.917 & 0.834/0.797 & 0.835/0.908 \\
GPT-5.4-mini & 0.30 & 0.727/0.815 & 0.730/0.843 & 0.737/0.772 & 0.726/0.922 \\
GPT-5.4 & 0.00 & 0.803/0.913 & 0.856/0.994 & 0.914/0.765 & 0.953/0.880 \\
GPT-5.4 & 0.15 & 0.770/0.876 & 0.778/0.897 & 0.810/0.723 & 0.847/0.917 \\
GPT-5.4 & 0.30 & 0.721/0.803 & 0.707/0.800 & 0.717/0.697 & 0.731/0.920 \\
GPT-5.5 & 0.00 & 0.885/0.949 & 0.828/0.992 & 0.871/0.713 & 0.971/0.934 \\
GPT-5.5 & 0.15 & 0.807/0.873 & 0.744/0.867 & 0.770/0.633 & 0.863/0.947 \\
GPT-5.5 & 0.30 & 0.713/0.745 & 0.663/0.741 & 0.681/0.606 & 0.732/0.912 \\
DeepSeek-V4-Pro & 0.00 & 0.884/0.944 & 0.826/0.991 & 0.885/0.765 & 0.972/0.938 \\
DeepSeek-V4-Pro & 0.15 & 0.803/0.867 & 0.741/0.865 & 0.778/0.639 & 0.865/0.952 \\
DeepSeek-V4-Pro & 0.30 & 0.706/0.741 & 0.659/0.739 & 0.682/0.607 & 0.733/0.911 \\\bottomrule\end{tabular}}
\caption{Session/pooled CBF under complete-coverage symmetric label noise; means over 100 repetitions. This slice isolates label noise at fixed coverage; the full coverage comparison appears in Tables~\ref{tab:noise_1}--\ref{tab:noise_0125}.}
\label{tab:noise_sens}
\end{table*}

Lower label noise often improves reconstruction, but the ordering of estimators changes with coverage and scoring unit. With complete clean labels, exact Count equals 1. Beta approaches this ceiling as observations accumulate; latent reliability fits additionally depend on optimization and identifiability (Table~\ref{tab:convergence}).

\subsection{Feedback Controls on the Common Pool}\label{app:matched_control}
Each model receives the same 400 questions, 20 per subject. The answer budget is six attempts: the first correct answer is retained, otherwise the last valid answer. The target is domain performance on this fixed answer pool. All models are scored on the 398 questions with valid answers from each, giving 19--20 questions per subject. Symmetric noise flips either class; FA and FR flip labels only for incorrect and correct answers, respectively. Coverage is 50\% or 100\%. Profiles share the same domain targets and uniform domain weights. The table reports CBF for the fixed label draws.
\begin{table*}[tbp]\centering\fontsize{9}{10.8}\selectfont\setlength{\tabcolsep}{4pt}
\begin{tabular*}{\textwidth}{@{\extracolsep{\fill}}lllrrrrrr@{}}\toprule
Model & Channel & Coverage & LSA & Global & Count & Beta & DS & GLAD-style \\\midrule
GPT-5.5 & Clean & 100\% & 0.989 & 0.938 & 1.000 & 0.961 & 0.820 & 0.939 \\
GPT-5.5 & Clean & 50\% & 0.940 & 0.939 & 0.962 & 0.929 & 0.822 & 0.945 \\
GPT-5.5 & Sym. 15\% & 100\% & 0.844 & 0.852 & 0.846 & 0.826 & 0.723 & 0.914 \\
GPT-5.5 & Sym. 15\% & 50\% & 0.857 & 0.858 & 0.859 & 0.815 & 0.757 & 0.900 \\
GPT-5.5 & Sym. 30\% & 100\% & 0.715 & 0.723 & 0.718 & 0.707 & 0.639 & 0.796 \\
GPT-5.5 & Sym. 30\% & 50\% & 0.705 & 0.718 & 0.707 & 0.689 & 0.634 & 0.756 \\
GPT-5.5 & FA 30\% & 100\% & 0.884 & 0.891 & 0.884 & 0.863 & 0.744 & 0.932 \\
GPT-5.5 & FA 30\% & 50\% & 0.880 & 0.891 & 0.885 & 0.845 & 0.757 & 0.918 \\
GPT-5.5 & FR 30\% & 100\% & 0.710 & 0.713 & 0.713 & 0.700 & 0.644 & 0.797 \\
GPT-5.5 & FR 30\% & 50\% & 0.690 & 0.708 & 0.707 & 0.684 & 0.642 & 0.762 \\
DeepSeek-V4-Pro & Clean & 100\% & 0.995 & 0.898 & 1.000 & 0.968 & 0.845 & 0.905 \\
DeepSeek-V4-Pro & Clean & 50\% & 0.927 & 0.900 & 0.945 & 0.940 & 0.827 & 0.875 \\
DeepSeek-V4-Pro & Sym. 15\% & 100\% & 0.858 & 0.850 & 0.876 & 0.861 & 0.768 & 0.901 \\
DeepSeek-V4-Pro & Sym. 15\% & 50\% & 0.858 & 0.863 & 0.872 & 0.854 & 0.772 & 0.874 \\
DeepSeek-V4-Pro & Sym. 30\% & 100\% & 0.760 & 0.772 & 0.783 & 0.771 & 0.723 & 0.838 \\
DeepSeek-V4-Pro & Sym. 30\% & 50\% & 0.658 & 0.758 & 0.760 & 0.747 & 0.703 & 0.755 \\
DeepSeek-V4-Pro & FA 30\% & 100\% & 0.915 & 0.879 & 0.919 & 0.904 & 0.809 & 0.899 \\
DeepSeek-V4-Pro & FA 30\% & 50\% & 0.859 & 0.880 & 0.892 & 0.877 & 0.794 & 0.892 \\
DeepSeek-V4-Pro & FR 30\% & 100\% & 0.735 & 0.748 & 0.753 & 0.744 & 0.713 & 0.817 \\
DeepSeek-V4-Pro & FR 30\% & 50\% & 0.685 & 0.746 & 0.750 & 0.734 & 0.702 & 0.765 \\
Qwen2.5-7B & Clean & 100\% & 0.835 & 0.867 & 1.000 & 0.985 & 0.938 & 0.855 \\
Qwen2.5-7B & Clean & 50\% & 0.885 & 0.871 & 0.913 & 0.946 & 0.887 & 0.784 \\
Qwen2.5-7B & Sym. 15\% & 100\% & 0.884 & 0.860 & 0.902 & 0.900 & 0.876 & 0.864 \\
Qwen2.5-7B & Sym. 15\% & 50\% & 0.874 & 0.869 & 0.917 & 0.914 & 0.879 & 0.828 \\
Qwen2.5-7B & Sym. 30\% & 100\% & 0.886 & 0.852 & 0.874 & 0.874 & 0.852 & 0.845 \\
Qwen2.5-7B & Sym. 30\% & 50\% & 0.860 & 0.858 & 0.867 & 0.869 & 0.830 & 0.760 \\
Qwen2.5-7B & FA 30\% & 100\% & 0.849 & 0.872 & 0.907 & 0.908 & 0.895 & 0.813 \\
Qwen2.5-7B & FA 30\% & 50\% & 0.888 & 0.866 & 0.904 & 0.922 & 0.910 & 0.787 \\
Qwen2.5-7B & FR 30\% & 100\% & 0.889 & 0.809 & 0.846 & 0.846 & 0.862 & 0.803 \\
Qwen2.5-7B & FR 30\% & 50\% & 0.855 & 0.833 & 0.851 & 0.854 & 0.818 & 0.752 \\
\bottomrule\end{tabular*}
\caption{Complete L2 comparisons on the common question pool (Appendix~\ref{app:matched_control}). CBF is higher-is-better. Clean denotes correct labels; Sym. flips both classes, FA only incorrect-answer labels, and FR only correct-answer labels. Coverage is the fraction of available labels. All methods use the same answer targets and observed labels within each model.}
\label{tab:common_pool_control}
\end{table*}

Table~\ref{tab:common_pool_control} separates history aggregation from direct counting. With complete clean labels, LSA reaches .989 on GPT-5.5 and .995 on DeepSeek-V4-Pro, but .835 on Qwen2.5-7B, while Count is 1.000 for all three. Thus, even when the feedback contains the exact outcomes, reconstruction accuracy depends on how the history is aggregated.

\subsection{Domain-Partition Sensitivity}\label{app:partition_sensitivity}
We use the same answers and strict feedback labels for five models. Tiers are formed at $\pm10/15/20$ percentage points around the model mean. We also retain 100 random four-group partitions and 100 random five-subject subsets. Each draw preserves a common target and domain support across methods. Grouping affects both the target dispersion and CBF, so absolute values across different partitions have different difficulty.

\begin{table*}[tbp]\centering\fontsize{9}{10.8}\selectfont\setlength{\tabcolsep}{4pt}

\fitTable{\begin{tabular}{llrrrr}\toprule
Model & Partition & Global & Beta & DS & GLAD-style \\\midrule
Qwen2.5-7B & random\_five\_subjects & 0.857 & 0.881 & 0.781 & 0.724 \\
Qwen2.5-7B & random\_four\_groups & 0.892 & 0.897 & 0.865 & 0.653 \\
Qwen2.5-7B & threshold\_0.10 & 0.841 & 0.915 & 0.949 & 0.683 \\
Qwen2.5-7B & threshold\_0.15 & 0.815 & 0.911 & 0.898 & 0.753 \\
Qwen2.5-7B & threshold\_0.20 & 0.779 & 0.894 & 0.889 & 0.757 \\
GPT-5.4-mini & random\_five\_subjects & 0.864 & 0.865 & 0.763 & 0.817 \\
GPT-5.4-mini & random\_four\_groups & 0.940 & 0.933 & 0.791 & 0.824 \\
GPT-5.4-mini & threshold\_0.10 & 0.849 & 0.946 & 0.860 & 0.804 \\
GPT-5.4-mini & threshold\_0.15 & 0.818 & 0.933 & 0.836 & 0.835 \\
GPT-5.4-mini & threshold\_0.20 & 0.810 & 0.947 & 0.841 & 0.879 \\
GPT-5.4 & random\_five\_subjects & 0.903 & 0.854 & 0.741 & 0.860 \\
GPT-5.4 & random\_four\_groups & 0.959 & 0.946 & 0.782 & 0.851 \\
GPT-5.4 & threshold\_0.10 & 0.880 & 0.953 & 0.839 & 0.885 \\
GPT-5.4 & threshold\_0.15 & 0.869 & 0.896 & 0.836 & 0.882 \\
GPT-5.4 & threshold\_0.20 & 0.977 & 0.977 & 0.835 & 0.838 \\
GPT-5.5 & random\_five\_subjects & 0.922 & 0.873 & 0.792 & 0.910 \\
GPT-5.5 & random\_four\_groups & 0.942 & 0.933 & 0.790 & 0.938 \\
GPT-5.5 & threshold\_0.10 & 0.939 & 0.919 & 0.769 & 0.885 \\
GPT-5.5 & threshold\_0.15 & 0.942 & 0.942 & 0.776 & 0.938 \\
GPT-5.5 & threshold\_0.20 & 0.942 & 0.942 & 0.776 & 0.938 \\
DeepSeek-V4-Pro & random\_five\_subjects & 0.896 & 0.858 & 0.778 & 0.911 \\
DeepSeek-V4-Pro & random\_four\_groups & 0.918 & 0.911 & 0.796 & 0.938 \\
DeepSeek-V4-Pro & threshold\_0.10 & 0.919 & 0.900 & 0.632 & 0.878 \\
DeepSeek-V4-Pro & threshold\_0.15 & 0.918 & 0.918 & 0.770 & 0.938 \\
DeepSeek-V4-Pro & threshold\_0.20 & 0.918 & 0.918 & 0.770 & 0.938 \\\bottomrule\end{tabular}}
\caption{Fixed-answer domain-partition sensitivity: pooled CBF, higher-is-better. Random conditions average 100 draws; threshold conditions use deviations of 10, 15, or 20 percentage points from model mean accuracy. Methods are compared within each partition; see Appendix~\ref{app:partition_sensitivity}.}
\label{tab:partition_sensitivity}
\end{table*}

\begin{table}[!htbp]\centering\fontsize{9}{10.8}\selectfont\setlength{\tabcolsep}{3pt}

\fitTable{\begin{tabular}{lrrrrr}\toprule
Model & Groups & Global & Beta & DS & GLAD-style \\\midrule
Qwen2.5-7B & 20 & 0.857 & 0.874 & 0.791 & 0.724 \\
Qwen2.5-7B & 11 & 0.864 & 0.877 & 0.831 & 0.700 \\
Qwen2.5-7B & 4 & 0.878 & 0.849 & 0.900 & 0.636 \\
GPT-5.4-mini & 20 & 0.861 & 0.859 & 0.849 & 0.811 \\
GPT-5.4-mini & 11 & 0.873 & 0.868 & 0.747 & 0.827 \\
GPT-5.4-mini & 4 & 0.940 & 0.920 & 0.836 & 0.781 \\
GPT-5.4 & 20 & 0.908 & 0.851 & 0.781 & 0.857 \\
GPT-5.4 & 11 & 0.908 & 0.879 & 0.829 & 0.853 \\
GPT-5.4 & 4 & 0.965 & 0.856 & 0.831 & 0.850 \\
GPT-5.5 & 20 & 0.931 & 0.873 & 0.818 & 0.909 \\
GPT-5.5 & 11 & 0.951 & 0.911 & 0.795 & 0.920 \\
GPT-5.5 & 4 & 0.956 & 0.895 & 0.828 & 0.924 \\
DeepSeek-V4-Pro & 20 & 0.904 & 0.860 & 0.819 & 0.913 \\
DeepSeek-V4-Pro & 11 & 0.923 & 0.884 & 0.797 & 0.922 \\
DeepSeek-V4-Pro & 4 & 0.945 & 0.915 & 0.799 & 0.911 \\\bottomrule\end{tabular}}
\caption{Pooled CBF after refitting on 20 subjects, 11 MMLU subcategories, and 4 MMLU categories. This semantic grouping complements the random and threshold partitions in Table~\ref{tab:partition_sensitivity}; joint interpretation in Appendix~\ref{app:partition_sensitivity}.}
\label{tab:semantic_granularity}
\end{table}

Semantic and random partitions both change method rankings. In the four-category partition, Global leads four of five models; Qwen2.5-7B favors DS. Grouping changes the prediction target and its dispersion, so a higher CBF on coarse groups is not evidence of better fine-domain discrimination.

\subsection{Noise and Coverage: Complete Grid}\label{app:noise_sessions}

Tables~\ref{tab:noise_1}, \ref{tab:noise_05}, and~\ref{tab:noise_0125} compare 100\%, 50\%, and 12.5\% label coverage on the same five-model pool. Within each table, rows vary label error rates; across tables, corresponding rows isolate the amount of feedback. Each cell lists the session mean first and the pooled score second.

\begin{table*}[tbp]\centering\fontsize{9}{10.8}\selectfont\setlength{\tabcolsep}{4pt}

\fitTable{\begin{tabular}{llcccc}\toprule
Model & FPR/FNR & Global & Beta & DS & GLAD-style \\\midrule
Qwen2.5-7B & 0.00/0.00 & 0.749/0.871 & 0.895/0.997 & 0.977/0.919 & 0.934/0.854 \\
Qwen2.5-7B & 0.15/0.15 & 0.746/0.865 & 0.823/0.943 & 0.853/0.906 & 0.825/0.909 \\
Qwen2.5-7B & 0.30/0.30 & 0.739/0.854 & 0.764/0.895 & 0.757/0.886 & 0.727/0.924 \\
Qwen2.5-7B & 0.30/0.10 & 0.741/0.868 & 0.812/0.929 & 0.827/0.915 & 0.803/0.794 \\
Qwen2.5-7B & 0.10/0.30 & 0.723/0.815 & 0.784/0.856 & 0.795/0.886 & 0.762/0.819 \\
GPT-5.4-mini & 0.00/0.00 & 0.774/0.869 & 0.870/0.996 & 0.949/0.838 & 0.943/0.879 \\
GPT-5.4-mini & 0.15/0.15 & 0.756/0.854 & 0.795/0.917 & 0.834/0.797 & 0.835/0.908 \\
GPT-5.4-mini & 0.30/0.30 & 0.727/0.815 & 0.730/0.843 & 0.737/0.772 & 0.726/0.922 \\
GPT-5.4-mini & 0.30/0.10 & 0.770/0.869 & 0.803/0.942 & 0.829/0.813 & 0.840/0.835 \\
GPT-5.4-mini & 0.10/0.30 & 0.705/0.782 & 0.736/0.804 & 0.756/0.766 & 0.739/0.852 \\
GPT-5.4 & 0.00/0.00 & 0.803/0.913 & 0.856/0.994 & 0.914/0.765 & 0.953/0.880 \\
GPT-5.4 & 0.15/0.15 & 0.770/0.876 & 0.778/0.897 & 0.810/0.723 & 0.847/0.917 \\
GPT-5.4 & 0.30/0.30 & 0.721/0.803 & 0.707/0.800 & 0.717/0.697 & 0.731/0.920 \\
GPT-5.4 & 0.30/0.10 & 0.792/0.907 & 0.793/0.946 & 0.815/0.738 & 0.863/0.867 \\
GPT-5.4 & 0.10/0.30 & 0.702/0.768 & 0.709/0.767 & 0.730/0.692 & 0.738/0.872 \\
GPT-5.5 & 0.00/0.00 & 0.885/0.949 & 0.828/0.992 & 0.871/0.713 & 0.971/0.934 \\
GPT-5.5 & 0.15/0.15 & 0.807/0.873 & 0.744/0.867 & 0.770/0.633 & 0.863/0.947 \\
GPT-5.5 & 0.30/0.30 & 0.713/0.745 & 0.663/0.741 & 0.681/0.606 & 0.732/0.912 \\
GPT-5.5 & 0.30/0.10 & 0.842/0.923 & 0.770/0.922 & 0.797/0.645 & 0.896/0.935 \\
GPT-5.5 & 0.10/0.30 & 0.702/0.730 & 0.661/0.727 & 0.684/0.603 & 0.732/0.893 \\
DeepSeek-V4-Pro & 0.00/0.00 & 0.884/0.944 & 0.826/0.991 & 0.885/0.765 & 0.972/0.938 \\
DeepSeek-V4-Pro & 0.15/0.15 & 0.803/0.867 & 0.741/0.865 & 0.778/0.639 & 0.865/0.952 \\
DeepSeek-V4-Pro & 0.30/0.30 & 0.706/0.741 & 0.659/0.739 & 0.682/0.607 & 0.733/0.911 \\
DeepSeek-V4-Pro & 0.30/0.10 & 0.839/0.913 & 0.767/0.920 & 0.800/0.655 & 0.896/0.939 \\
DeepSeek-V4-Pro & 0.10/0.30 & 0.696/0.727 & 0.658/0.725 & 0.686/0.605 & 0.733/0.892 \\\bottomrule\end{tabular}}
\caption{CBF at 100\% label coverage. Each cell gives session/pooled scores averaged over 100 repetitions; higher is better. FPR/FNR are false-positive/false-negative label rates. Compare Tables~\ref{tab:noise_05} and~\ref{tab:noise_0125} for reduced coverage; analysis in Appendix~\ref{app:noise_sessions}.}
\label{tab:noise_1}
\end{table*}

\begin{table*}[tbp]\centering\fontsize{9}{10.8}\selectfont\setlength{\tabcolsep}{4pt}

\fitTable{\begin{tabular}{llcccc}\toprule
Model & FPR/FNR & Global & Beta & DS & GLAD-style \\\midrule
Qwen2.5-7B & 0.00/0.00 & 0.744/0.870 & 0.813/0.965 & 0.808/0.903 & 0.804/0.848 \\
Qwen2.5-7B & 0.15/0.15 & 0.740/0.865 & 0.777/0.931 & 0.743/0.889 & 0.738/0.888 \\
Qwen2.5-7B & 0.30/0.30 & 0.734/0.854 & 0.745/0.888 & 0.681/0.859 & 0.674/0.889 \\
Qwen2.5-7B & 0.30/0.10 & 0.738/0.868 & 0.775/0.921 & 0.734/0.891 & 0.730/0.797 \\
Qwen2.5-7B & 0.10/0.30 & 0.718/0.814 & 0.756/0.854 & 0.703/0.868 & 0.696/0.809 \\
GPT-5.4-mini & 0.00/0.00 & 0.769/0.869 & 0.785/0.969 & 0.819/0.843 & 0.829/0.873 \\
GPT-5.4-mini & 0.15/0.15 & 0.749/0.854 & 0.744/0.908 & 0.746/0.801 & 0.750/0.896 \\
GPT-5.4-mini & 0.30/0.30 & 0.721/0.815 & 0.705/0.838 & 0.674/0.769 & 0.673/0.893 \\
GPT-5.4-mini & 0.30/0.10 & 0.765/0.869 & 0.754/0.933 & 0.755/0.815 & 0.765/0.839 \\
GPT-5.4-mini & 0.10/0.30 & 0.700/0.782 & 0.707/0.802 & 0.682/0.766 & 0.679/0.838 \\
GPT-5.4 & 0.00/0.00 & 0.797/0.913 & 0.767/0.972 & 0.822/0.772 & 0.850/0.884 \\
GPT-5.4 & 0.15/0.15 & 0.763/0.876 & 0.722/0.890 & 0.747/0.726 & 0.764/0.911 \\
GPT-5.4 & 0.30/0.30 & 0.715/0.803 & 0.679/0.798 & 0.671/0.699 & 0.676/0.892 \\
GPT-5.4 & 0.30/0.10 & 0.786/0.907 & 0.736/0.935 & 0.761/0.744 & 0.786/0.871 \\
GPT-5.4 & 0.10/0.30 & 0.697/0.768 & 0.678/0.765 & 0.674/0.693 & 0.678/0.856 \\
GPT-5.5 & 0.00/0.00 & 0.875/0.949 & 0.733/0.977 & 0.781/0.725 & 0.887/0.935 \\
GPT-5.5 & 0.15/0.15 & 0.797/0.872 & 0.680/0.861 & 0.744/0.645 & 0.783/0.947 \\
GPT-5.5 & 0.30/0.30 & 0.706/0.744 & 0.629/0.738 & 0.658/0.612 & 0.675/0.895 \\
GPT-5.5 & 0.30/0.10 & 0.832/0.922 & 0.700/0.914 & 0.764/0.658 & 0.818/0.937 \\
GPT-5.5 & 0.10/0.30 & 0.695/0.730 & 0.626/0.724 & 0.657/0.610 & 0.673/0.879 \\
DeepSeek-V4-Pro & 0.00/0.00 & 0.873/0.943 & 0.730/0.976 & 0.788/0.770 & 0.889/0.941 \\
DeepSeek-V4-Pro & 0.15/0.15 & 0.793/0.866 & 0.677/0.858 & 0.744/0.647 & 0.784/0.951 \\
DeepSeek-V4-Pro & 0.30/0.30 & 0.699/0.741 & 0.624/0.735 & 0.657/0.615 & 0.676/0.891 \\
DeepSeek-V4-Pro & 0.30/0.10 & 0.829/0.912 & 0.696/0.911 & 0.765/0.661 & 0.818/0.942 \\
DeepSeek-V4-Pro & 0.10/0.30 & 0.688/0.726 & 0.622/0.721 & 0.657/0.612 & 0.674/0.874 \\\bottomrule\end{tabular}}
\caption{CBF at 50\% random label coverage, using the same models and noise channels as Table~\ref{tab:noise_1}. Each cell gives session/pooled scores averaged over 100 repetitions. Joint coverage analysis: Appendix~\ref{app:noise_sessions}.}
\label{tab:noise_05}
\end{table*}

\begin{table*}[tbp]\centering\fontsize{9}{10.8}\selectfont\setlength{\tabcolsep}{4pt}

\fitTable{\begin{tabular}{llccccc}\toprule
Model & FPR/FNR & Global & Count & Beta & DS & GLAD-style \\\midrule
Qwen2.5-7B & 0.00/0.00 & 0.725/0.869 & 0.735/0.905 & 0.759/0.917 & 0.736/0.851 & 0.736/0.808 \\
Qwen2.5-7B & 0.15/0.15 & 0.720/0.864 & 0.713/0.889 & 0.749/0.895 & 0.718/0.817 & 0.715/0.816 \\
Qwen2.5-7B & 0.30/0.30 & 0.714/0.853 & 0.692/0.853 & 0.740/0.860 & 0.700/0.759 & 0.693/0.791 \\
Qwen2.5-7B & 0.30/0.10 & 0.724/0.868 & 0.713/0.880 & 0.749/0.894 & 0.718/0.815 & 0.714/0.782 \\
Qwen2.5-7B & 0.10/0.30 & 0.701/0.814 & 0.699/0.827 & 0.743/0.837 & 0.707/0.780 & 0.700/0.750 \\
GPT-5.4-mini & 0.00/0.00 & 0.745/0.869 & 0.714/0.915 & 0.711/0.923 & 0.702/0.827 & 0.714/0.846 \\
GPT-5.4-mini & 0.15/0.15 & 0.724/0.853 & 0.687/0.880 & 0.699/0.876 & 0.682/0.778 & 0.687/0.844 \\
GPT-5.4-mini & 0.30/0.30 & 0.699/0.815 & 0.659/0.821 & 0.687/0.818 & 0.661/0.728 & 0.660/0.810 \\
GPT-5.4-mini & 0.30/0.10 & 0.743/0.869 & 0.695/0.900 & 0.703/0.902 & 0.687/0.790 & 0.695/0.836 \\
GPT-5.4-mini & 0.10/0.30 & 0.681/0.782 & 0.660/0.788 & 0.687/0.789 & 0.663/0.725 & 0.661/0.771 \\
GPT-5.4 & 0.00/0.00 & 0.768/0.912 & 0.706/0.929 & 0.679/0.928 & 0.677/0.779 & 0.706/0.880 \\
GPT-5.4 & 0.15/0.15 & 0.732/0.875 & 0.673/0.871 & 0.666/0.859 & 0.659/0.725 & 0.673/0.872 \\
GPT-5.4 & 0.30/0.30 & 0.690/0.802 & 0.640/0.788 & 0.653/0.779 & 0.637/0.685 & 0.641/0.814 \\
GPT-5.4 & 0.30/0.10 & 0.756/0.906 & 0.684/0.904 & 0.671/0.898 & 0.663/0.736 & 0.684/0.871 \\
GPT-5.4 & 0.10/0.30 & 0.675/0.768 & 0.640/0.761 & 0.652/0.754 & 0.637/0.685 & 0.640/0.786 \\
GPT-5.5 & 0.00/0.00 & 0.821/0.948 & 0.688/0.954 & 0.621/0.935 & 0.610/0.761 & 0.686/0.937 \\
GPT-5.5 & 0.15/0.15 & 0.750/0.870 & 0.645/0.860 & 0.605/0.831 & 0.608/0.672 & 0.644/0.922 \\
GPT-5.5 & 0.30/0.30 & 0.673/0.743 & 0.603/0.741 & 0.588/0.722 & 0.589/0.627 & 0.603/0.833 \\
GPT-5.5 & 0.30/0.10 & 0.782/0.921 & 0.661/0.905 & 0.611/0.878 & 0.611/0.689 & 0.660/0.933 \\
GPT-5.5 & 0.10/0.30 & 0.664/0.729 & 0.601/0.728 & 0.587/0.709 & 0.588/0.624 & 0.601/0.817 \\
DeepSeek-V4-Pro & 0.00/0.00 & 0.817/0.942 & 0.687/0.952 & 0.615/0.927 & 0.607/0.776 & 0.685/0.942 \\
DeepSeek-V4-Pro & 0.15/0.15 & 0.746/0.865 & 0.643/0.857 & 0.599/0.824 & 0.605/0.663 & 0.642/0.920 \\
DeepSeek-V4-Pro & 0.30/0.30 & 0.667/0.740 & 0.600/0.738 & 0.582/0.717 & 0.584/0.639 & 0.600/0.829 \\
DeepSeek-V4-Pro & 0.30/0.10 & 0.777/0.911 & 0.658/0.901 & 0.605/0.869 & 0.607/0.683 & 0.657/0.933 \\
DeepSeek-V4-Pro & 0.10/0.30 & 0.658/0.725 & 0.599/0.724 & 0.581/0.704 & 0.584/0.634 & 0.599/0.815 \\\bottomrule\end{tabular}}
\caption{CBF at 12.5\% random label coverage, including direct Count. Each cell gives session/pooled scores averaged over 100 repetitions. Together with Tables~\ref{tab:noise_1} and~\ref{tab:noise_05}, this tests the effect of sparse feedback (Appendix~\ref{app:noise_sessions}).}
\label{tab:noise_0125}
\end{table*}

Tables~\ref{tab:noise_1}--\ref{tab:noise_0125} show that coverage can change the relative advantage of reliability estimation. On GPT-5.5 with clean labels, GLAD-style exceeds Global at full coverage (.971 vs.\ .885 session CBF), but falls below it at 12.5\% coverage (.686 vs.\ .821). Sparse feedback therefore changes the ranking even without label errors. Session and pooled scores report this effect at different aggregation scales.

\subsection{Dual-Role Bias}
\label{app:dual_role}

The Qwen-Max agent--simulator pairing produces few explicit corrections of incorrect answers. Utterances such as ``I believe there might be a small misunderstanding'' illustrate why hedged language can remain undecoded by a strict binary classifier. The comparison varies both the generated feedback and its linguistic form; the measured effect is feedback coverage.

The Qwen-Max/Qwen-Turbo simulator comparison varies both generated utterances and their decodability. It is a within-family intervention, so it cannot separate shared-family effects from model-specific response style.

Simulator identity and classifier coverage are therefore reported as distinct properties of each trace.

\subsection{Within-Family Simulator Comparison}\label{app:cross_sim}

We compare a Qwen-Max answering model under Qwen-Max and Qwen-Turbo feedback simulators. Both conditions contain ten sessions of 50 attempted turns. Table~\ref{tab:cross_sim_ablation} is a descriptive comparison within the Qwen family; the 0.6 percentage-point answer-accuracy difference also reflects different recorded answer samples.

\begin{table}[!htbp]
\centering\fontsize{9}{10.8}\selectfont\setlength{\tabcolsep}{3pt}

\fitTable{\begin{tabular}{lccccc}
\toprule
\textbf{Simulator} & \textbf{Acc} & \textbf{ECE} & \textbf{HR} & \textbf{CBF} \\
\midrule
user = Qwen-Max    & 78.2\% & 0.165 & 20.2\% & 0.717 \\
user = Qwen-Turbo  & 77.6\% & 0.165 & 20.0\% & 0.722 \\
\midrule
$|\Delta|$ & 0.6pp & 0.001 & 0.2pp & 0.005 \\
\bottomrule
\end{tabular}}
\caption{\textbf{Within-family simulator comparison for Qwen-Max.} Recorded 500-turn runs with Qwen-Max and Qwen-Turbo simulators.}
\label{tab:cross_sim_ablation}
\end{table}

The displayed runs differ by .6 percentage points in task accuracy and .005 in CBF. They provide a descriptive within-family comparison. Because answers precede user feedback, similar raw answer confidence does not establish that the feedback is equally informative.

\subsection{R4 Adaptive Selection and Second-Order Estimators: Full Details}
\label{app:r4}

\begin{figure*}[tbp]
\centering
\includegraphics[width=\textwidth]{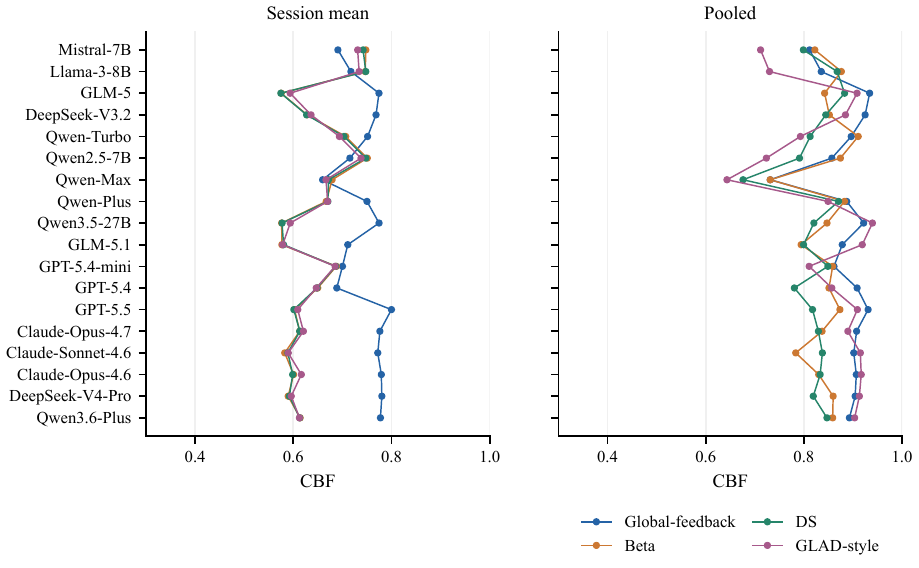}
\caption{\textbf{Reliability-aware estimation at two scoring units.} Session and pooled CBF for all 18 models under a shared label stream.}
\label{fig:second_order}
\end{figure*}

R4 chooses ZS when its measured CBF is higher than NoMem and LSA otherwise. We evaluate this descriptive selection on the 18 models for which both self-assessment panels are complete. Figure~\ref{fig:second_order} separately compares the reliability-aware estimators under the matched session and pooled units.

\paragraph{Observed selections.} R4 selects ZS on 14 of 18 complete models. Its mean CBF is 0.744, compared with 0.747 for always-ZS and 0.674 for always-LSA.

\paragraph{Selection target.} The score used to select a branch depends on the empirical domain outcomes. This makes R4 an outcome-informed diagnostic, even though its threshold is fixed.

\paragraph{Prediction.} A deployment rule must choose without the test outcomes. The held-out ReAct comparison below evaluates profiles formed only from the training half of each subject.

\paragraph{Best-of-two reference.} Choosing the higher observed score of ZS and LSA gives mean CBF 0.762 on these 18 models. This is a selection reference, not an implementable predictor without outcome information.

\section{Cross-Task Heterogeneity}
\label{app:group_f}

\subsection{Cross-Task Heterogeneity: Full Per-Task Comparison}
\label{app:fourtask}

Table~\ref{tab:fourtask_overview} includes all 18 models and the three extension tasks. MMLU uses valid-answer accuracy and ECE. GSM8K reports the archived numeric-answer score; HumanEval reports the fraction passing the custom test harness, and BFCL reports function-name/required-key matches. Gaps compare recorded confidence with each task's score. For Qwen3.6-Plus, the archived per-item flags reproduce 182/300 on GSM8K and 101/164 on HumanEval. These lower scores alone do not identify an API or parser failure. The extension archives use fixed task denominators; failed or unparsed cases cannot be consistently separated without the missing full responses. Their scores are therefore reported under their own archived scoring protocol, separately from valid-answer primary QA accuracy.

\begin{table*}[tbp]
\centering\fontsize{9}{10.8}\selectfont

\setlength{\tabcolsep}{4pt}
\fitTable{\begin{tabular}{l rr rr rr rr}
\toprule
& \multicolumn{2}{c}{\textbf{MMLU}} & \multicolumn{2}{c}{\textbf{GSM8K}} & \multicolumn{2}{c}{\textbf{HumanEval}} & \multicolumn{2}{c}{\textbf{BFCL name/key}} \\
\cmidrule(lr){2-3} \cmidrule(lr){4-5} \cmidrule(lr){6-7} \cmidrule(lr){8-9}
\textbf{Model} & Acc & ECE & Acc & Gap & Custom tests & Gap & Acc & Gap \\
\midrule
Mistral-7B & 41.6 & 0.582 & 55 & $+$31 & 31 & $+$52 & 71 & $+$15 \\
Llama-3-8B & 57.7 & 0.357 & 84 & $-$3  & 33 & $+$44 & 46 & $+$36 \\
GLM-5 & 95.8 & 0.039 & 96 & $+$1  & 91 & $-$49 & 64 & $+$32 \\
DeepSeek-V3.2 & 88.6 & 0.023 & 94 & $-$21 & 87 & $-$30 & 75 & $+$5  \\
Qwen-Turbo & 75.5 & 0.188 & 95 & $-$33 & 74 & $-$1  & 72 & $+$22 \\
Qwen2.5-7B & 60.6 & 0.259 & 90 & $-$18 & 64 & $-$4  & 73 & $+$18 \\
Qwen-Max & 78.8 & 0.167 & 97 & $-$18 & 82 & $-$31 & 74 & $+$17 \\
Qwen-Plus & 82.2 & 0.144 & 90 & $-$11 & 92 & $-$48 & 80 & $+$10 \\
Qwen3.5-27B & 95.9 & 0.033 & 98 & $-$1  & 77 & $-$35 & 72 & $+$24 \\
\midrule
GPT-5.4-mini & 75.0 & 0.199 & 95 & $-$2  & 74 & $-$25 & 76 & $+$15 \\
GPT-5.4 & 82.8 & 0.132 & 97 & $+$0.3 & 81 & $-$36 & 70 & $+$25 \\
Claude-Sonnet-4.6 & 93.0 & 0.052 & 88 & $-$4  & 79 & $-$40 & 72 & $+$17 \\
Claude-Opus-4.6 & 93.0 & 0.020 & 99 & $-$5  & 74 & $-$17 & 78 & $+$9  \\
GPT-5.5 & 92.8 & 0.032 & 97 & $-$1  & 85 & $-$31 & 97 & $-$4  \\
Claude-Opus-4.7 & 90.4 & 0.022 & 99 & $-$10 & 95 & $-$34 & 99 & $-$9  \\
\midrule
GLM-5.1 & 94.2 & 0.040 & 96 & --- & 79 & --- & 67 & --- \\
DeepSeek-V4-Pro & 92.8 & 0.047 & 97 & --- & 94 & --- & 70 & --- \\
Qwen3.6-Plus & 90.7 & 0.068 & 61 & $+$18 & 62 & $+$15 & 70 & $+$27 \\
\bottomrule
\end{tabular}}
\caption{Four-task results for all 18 models. MMLU: valid-answer accuracy (\%) and ECE; extension tasks: recorded task scores (\%) and confidence minus score (percentage points), with BFCL using function-name/argument-key checks. Dashes denote unavailable confidence. Scoring and cross-task interpretation: Appendix~\ref{app:fourtask} and~\ref{app:proto_controlled}.}
\label{tab:fourtask_overview}
\end{table*}

\subsection{Recorded Cross-Task Scores}\label{app:lite}\label{app:proto_controlled}

\begin{figure}[!htbp]
\centering
\includegraphics[width=\columnwidth]{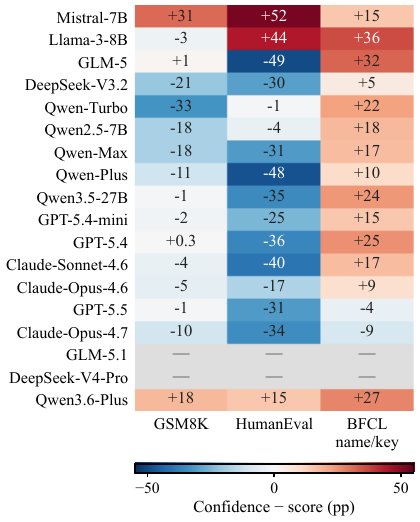}
\caption{\textbf{Task-dependent recorded confidence gaps.} Confidence minus task score; gray cells indicate unavailable confidence. BFCL is the name/key control.}
\label{fig:signflip}
\end{figure}

A sign change indicates opposite directions of the recorded confidence--score gap across tasks. We display all 18 rows, with unavailable confidence gaps left blank. The comparison describes task dependence under the extension scorers.

\begin{table}[!htbp]
\centering\fontsize{9}{10.8}\selectfont\setlength{\tabcolsep}{3pt}

\fitTable{\begin{tabular}{l c c c c}
\toprule
\textbf{Model} & \makecell{GSM8K\\Gap} & \makecell{HumanEval\\Gap} & \makecell{BFCL\\Gap} & \makecell{Sign-flip\\in subset?} \\
\midrule
Mistral-7B        & $+$31 & $+$52 & $+$15 & no (all $+$) \\
Llama-3-8B        & $-$3  & $+$44 & $+$36 & \textbf{yes} \\
GLM-5             & $+$1  & $-$49 & $+$32 & \textbf{yes} \\
DeepSeek-V3.2     & $-$21 & $-$30 & $+$5  & \textbf{yes} \\
Qwen-Turbo        & $-$33 & $-$1  & $+$22 & \textbf{yes} \\
Qwen2.5-7B        & $-$18 & $-$4  & $+$18 & \textbf{yes} \\
Qwen-Max          & $-$18 & $-$31 & $+$17 & \textbf{yes} \\
Qwen-Plus         & $-$11 & $-$48 & $+$10 & \textbf{yes} \\
Qwen3.5-27B       & $-$1  & $-$35 & $+$24 & \textbf{yes} \\
GPT-5.4-mini      & $-$2  & $-$25 & $+$15 & \textbf{yes} \\
GPT-5.4           & $+$0.3 & $-$36 & $+$25 & \textbf{yes} \\
Claude-Sonnet-4.6 & $-$4  & $-$40 & $+$17 & \textbf{yes} \\
Claude-Opus-4.6   & $-$5  & $-$17 & $+$9  & \textbf{yes} \\
GPT-5.5           & $-$1  & $-$31 & $-$4  & no (all $-$) \\
Claude-Opus-4.7   & $-$10 & $-$34 & $-$9  & no (all $-$) \\
GLM-5.1            & --- & --- & --- & --- \\
DeepSeek-V4-Pro    & --- & --- & --- & --- \\
Qwen3.6-Plus      & $+$18 & $+$15 & $+$27 & no (all $+$) \\
\midrule
\multicolumn{4}{l}{Opposite recorded signs (available rows)} & 12/16 \\
\bottomrule
\end{tabular}}
\caption{\textbf{Recorded confidence--score gaps across the three extension tasks (pp).} Results for 18 models; unavailable confidence gaps are unscored.}
\label{tab:signflip_data}
\end{table}

Twelve of the 16 rows with available confidence gaps have opposite recorded signs across the three extension tasks. These signs describe the stored summaries; the task scores and confidence extraction remain part of what is being measured.

The full task matrix supports comparisons within each recorded task protocol. Further task-specific validation is needed before using these profiles to allocate tool-use tasks.

\subsection{Lite protocol: T=100 vs T=500 rankings}

The stored lite-protocol analysis compares the first 100 and 500 records across 23 run configurations, including repeated model variants. Table~\ref{tab:lite_protocol} reports its rank correlations as a descriptive budget-sensitivity comparison.

\begin{table}[tbp]
\centering\fontsize{9}{10.8}\selectfont

\fitTable{\begin{tabular}{lcc}
\toprule
\textbf{Metric} & \textbf{Pearson $r$ (100 vs 500)} & \textbf{Spearman $\rho$} \\
\midrule
ECE & 0.966 & 0.911 \\
CBF & 0.967 & 0.933 \\
\bottomrule
\end{tabular}}
\caption{\textbf{Lite/full budget correlations on 23 recorded run configurations.} Pearson and Spearman statistics compare 100 versus 500 records.}
\label{tab:lite_protocol}
\end{table}

The correlations show agreement between these finite-budget summaries. Correlation is not a percentage of information retained, and this run-configuration analysis does not establish convergence of the 18-model session protocol.

\subsection{Held-Out ReAct Prediction}\label{app:react_prediction}
The six configurations contribute 960 stored records. For each subject, the first four questions form the profile and the last four define the prediction target. Both direct and ReAct answers pass the same A--D parsing gate; a subject requires at least two valid answers in both halves and modes. Methods share test support within configuration. Count and Beta use training outcomes, while Global uses the pooled training accuracy. Lower MAE is better.

\begin{table}[tbp]\centering\fontsize{9}{10.8}\selectfont\setlength{\tabcolsep}{4pt}

\fitTable{\begin{tabular}{lrrrrr}\toprule
Config. & Global & ZS & \makecell{Direct\\Count} & \makecell{ReAct\\Count} & \makecell{ReAct\\Beta} \\\midrule
qwen25\_1\_5b & 0.190 & 0.446 & 0.258 & 0.308 & 0.229 \\
qwen25\_7b & 0.237 & 0.252 & 0.346 & 0.285 & 0.268 \\
qwen25\_14b & 0.233 & 0.248 & 0.267 & 0.254 & 0.231 \\
qwen3\_4b\_2507 & 0.242 & 0.216 & 0.384 & 0.190 & 0.247 \\
qwen3\_8b & 0.166 & 0.211 & 0.197 & 0.167 & 0.176 \\
qwen3\_14b & 0.206 & 0.217 & 0.267 & 0.171 & 0.191 \\\bottomrule\end{tabular}}
\caption{Predicting held-out ReAct domain accuracy: MAE, lower-is-better. Columns compare a global training mean, the zero-shot prior, and domain profiles from direct-answer or ReAct training records; see Appendix~\ref{app:react_prediction}.}
\label{tab:react_prediction}
\end{table}

Table~\ref{tab:react_prediction} gives Global the lowest MAE in three of six configurations, ReAct Count in two, and ReAct Beta in one. For Qwen3-14B, ReAct Count improves on Global (.171 vs.\ .206), whereas Qwen2.5-7B favors Global (.237 vs.\ .285). Domain profiles can therefore aid held-out prediction, but their advantage over a global estimate varies by configuration.

\section{Implementation Notes and Extended Material}
\label{app:group_g}

\subsection{Implementation Note: Algorithmic-Baseline self\_assess Specification}
\label{sec:bugfix}

\paragraph{Interface specification.} The algorithmic baselines return a domain probability through \texttt{self\_assess(domain)}. UF-PDC returns the positive-label fraction, Bayes-Domain a Beta posterior mean, and EM-Joint the fitted latent-correctness rate. Platt-Online and HistBin-Online retain per-answer recalibration but use the label-count profile for domain CBF, explaining their identical CBF columns. DS and GLAD-style use the likelihoods in Appendix~\ref{app:baselines}.

\paragraph{Per-baseline definition.}
\begin{itemize}[nosep,leftmargin=1.4em]
  \item Platt-Online / HistBin-Online: $\hat p(d) = \mathrm{dom\_correct}[d] / \mathrm{dom\_total}[d]$ (or 0.5 if no observations).
  \item Bayes-Domain: $\hat p(d) = \alpha[d]/(\alpha[d]+\beta[d])$.
  \item EM-Joint: $\hat p(d) = \mathrm{clip}(p^*[d], 10^{-3}, 1{-}10^{-3})$ from EM iterate.
  \item UF-PDC: $\hat p(d) = \mathrm{correct}[d]/\mathrm{total}[d]$ from decoded feedback.
\end{itemize}

\paragraph{Per-baseline summary.} UF-PDC, Platt, and HistBin return identical domain counts on matched labels, so the main table shares their endpoint column as Count*. Their query-confidence mechanisms remain distinct. Bayes adds smoothing and EM-Joint fits latent reliability.

\subsection{Extended Related Work Discussion}
\label{app:related_extended}

This section provides the detailed discussion that was compressed in the main text's Related Work section. We organize the literature into thematic clusters and position LADDER against each cluster's design space.

\paragraph{LLM Self-Knowledge and Metacognition.}
Prior work studies several forms of self-knowledge. \citet{kadavath2022language} examine answer confidence through P(True); \citet{jian2025metacognition} study monitoring and control of internal activations; and \citet{binder2025introspection} use fine-tuning to test models' predictions of their own behavior. \emph{Behavioral evaluation without self-reports}~\citep{ackerman2026metacognition} finds that frontier models exhibit limited, context-dependent metacognitive signals when not explicitly prompted. \emph{Self-improvement framings}~\citep{liu2025metacognitive_learning} argue that intrinsic metacognitive learning is a precondition for autonomous improvement. Most directly relevant, \citet{barkan2026capable} test LLMs on BigCodeBench and SWE-Bench under multi-step tool execution and find systematic overconfidence, which worsens during execution for several frontier models. LADDER differs by measuring whether the agent can convert \emph{noisy interactive feedback} into a stable per-domain self-model---an ability orthogonal to one-shot probing or clean execution traces.

\paragraph{Knowledge Boundaries and Calibration Benchmarks.}
Existing benchmarks evaluate whether LLMs know what they know on a \emph{static, per-query} basis. \emph{Boundary taxonomies}~\citep{li2025knowledge_boundary, yin2024benchmarking_boundary} categorize how prompts and model knowledge affect answerability. \emph{Static factuality benchmarks}~\citep{wei2024simpleqa, hendrycks2021measuring, lin2022truthfulqa} evaluate short-answer factuality, multi-task understanding, and adversarial truthfulness. \emph{Multi-dimensional aggregations}~\citep{liang2022helm} include ECE alongside other metrics, and surveys~\citep{geng2024survey} compile method comparisons. None evaluates whether the agent's self-model can be \emph{accumulated} over noisy interaction; LADDER targets this gap by making per-domain accuracy estimation the explicit measurement axis.

\paragraph{LLM agent memory systems.} Memory architectures and benchmarks~\citep{packer2023memgpt,mem02025,li2025memos,tan2025rmm,xu2025amem,mem1_2026,hu2026memoryagent} motivate persistent summaries of interaction. Domain reliability adds an evaluable summary whose target is the model's observed answer accuracy.

\paragraph{Learning from Implicit Feedback.}
Three regimes characterize this line. \emph{Coactive learning} formalizes the user-edit setting but assumes structured, reliable signals. \emph{Retrospective dialogue distillation}~\citep{liu2025implicit} mines real interaction logs (WildChat, LMSYS), finding follow-up utterances often contain feedback but distillation outcomes vary with dialogue length---helpful on short MTBench questions, less so on longer WildBench ones. \emph{RLHF-induced overconfidence}~\citep{leng2025taming} shows that the dominant feedback-learning paradigm itself produces miscalibrated models. LADDER targets the orthogonal regime of \emph{inference-time} feedback accumulation under noisy, domain-dependent reliability---a setting where neither training-time RLHF nor offline distillation applies.

\paragraph{Selective Prediction and Abstention.}
These studies relate uncertainty estimates to answering or abstaining. \emph{Fine-tuning for uncertainty}~\citep{kapoor2024taught, diao2024rtuning} produces generalizable abstention behavior with as few as 1k LoRA samples or via R-Tuning's meta-skill formulation. \emph{Self-reflective rationale generation}~\citep{xu2024sayself} optimizes per-query confidence without persistent memory. \emph{Selective classification}~\citep{traub2024augrc} introduces AUGRC; surveys~\citep{li2025abstention} map the methodological landscape. \emph{Reasoning-induced regression}~\citep{kirichenko2025abstentionbench} shows that reasoning fine-tuning \emph{degrades} abstention by an average of 24\% on AbstentionBench. \emph{Epistemic vs.\ aleatoric separation}~\citep{ahdritz2024knowable} uses linear probes on frozen embeddings. They focus on query-level uncertainty; LADDER examines domain profiles formed from accumulated feedback.

\paragraph{Simulated Users and Closest Competitors.}
The closest competitors share LADDER's goal of evaluating self-knowledge under realistic interaction but differ on simulator design or feedback channel. \emph{Simulator validity}~\citep{zhou2026sim2real} quantifies the Sim2Real gap and shows LLM simulators bias toward an ``easy mode'' with positivity bias. \emph{Calibration through interaction}~\citep{fang2025ccg} (CCG) studies game-theoretic confidence adjustment using feedback computed from verified correctness and reported confidence. \emph{Learning confidence from sampled answers}~\citep{han2024lepe} (LePe) uses correctness frequencies against ground truth as confidence targets and requires offline weight updates. \emph{Per-query confidence elicitation}~\citep{xu2024sayself} (SaySelf) lacks persistent memory. \emph{Capability boundary collapse in RL}~\citep{dong2025rlplus} formalizes a related failure mode under RL training. \emph{Cross-task calibration}~\citep{shen2024thermometer} seeks transferable confidence estimates, while \emph{group-dependent calibration}~\citep{chen2024reconfidencing,manggala2025qacal} addresses variation across question groups. LADDER occupies a position none covers: noisy interactive feedback, accumulation horizon, and second-order user-reliability inference, scored against an oracle-bracketed ladder rather than a clean ground-truth signal.

\subsection{Model Configurations}\label{app:api_versions}

The panel contains 18 models across seven families (Table~\ref{tab:main}). Local inference uses Mistral-7B-Instruct-v0.3, Meta-Llama-3-8B-Instruct, and Qwen2.5-7B-Instruct. Other models are accessed through their named API configurations. LSA and L4 return one domain-indexed JSON profile with a 1,100-token output budget; local decoding is greedy and API requests use temperature zero.

\subsection{Extended Results}
\label{app:extended}

Table~\ref{tab:tier_cbf} reports pooled CBF within each empirical subject tier for Llama-3-8B, Qwen-Turbo, and Qwen-Max. It compares the constant profile, prior, and Beta estimate on the same tier targets.

\begin{table}[tbp]\centering\fontsize{9}{10.8}\selectfont\setlength{\tabcolsep}{4pt}

\fitTable{\begin{tabular}{llrrr}\toprule
Model & Method & Strong & Medium & Weak \\\midrule
Llama-3-8B & NoMemory & 0.746 & 0.946 & 0.870 \\
Llama-3-8B & ZS-SK & 0.846 & 0.901 & 0.865 \\
Llama-3-8B & Beta & 0.926 & 0.831 & 0.879 \\
Qwen-Turbo & NoMemory & 0.615 & 0.726 & 0.890 \\
Qwen-Turbo & ZS-SK & 0.949 & 0.839 & 0.714 \\
Qwen-Turbo & Beta & 0.916 & 0.912 & 0.905 \\
Qwen-Max & NoMemory & 0.627 & 0.709 & 0.826 \\
Qwen-Max & ZS-SK & 0.948 & 0.927 & 0.803 \\
Qwen-Max & Beta & 0.542 & 0.739 & 0.887 \\\bottomrule\end{tabular}}
\caption{Per-tier pooled CBF on the observed subject partition.}
\label{tab:tier_cbf}
\end{table}

\subsection{Reproducibility Index}
\label{app:repro_index}

The reproducibility index links the principal outputs to their analysis inputs. The saved records and machine-readable results accompany the manuscript sources.

\begin{table}[!htbp]\centering\fontsize{9}{10.8}\selectfont\setlength{\tabcolsep}{3pt}

\fitTable{\begin{tabular}{@{}>{\raggedright\arraybackslash}p{.22\linewidth}>{\raggedright\arraybackslash}p{.37\linewidth}>{\raggedright\arraybackslash}p{.32\linewidth}@{}}\toprule
Output & Input & Operation \\\midrule
Tables~\ref{tab:main}, \ref{tab:matched_l2} & QA records; self-assessment profiles & Matched domain scoring \\
Table~\ref{tab:second_order_full} & Same decoded-label records & Session/pooled refitting \\
Table~\ref{tab:noise_sens} & Fixed answers; 100 noise repetitions & Session/pooled CBF \\
Table~\ref{tab:subjects} & Empirical subject outcomes & Per-model ranks \\
Table~\ref{tab:react_prediction} & 960 ReAct records & Within-subject held-out scoring \\
Figures~\ref{fig:spectrum}, \ref{fig:heatmap}, \ref{fig:ladder} & Main-table profiles & Shared score rendering \\\bottomrule\end{tabular}}
\caption{Reproducibility index for the primary analyses.}
\label{tab:reproducibility_index}
\end{table}

\paragraph{Supplementary materials.} The supplementary materials provide scoring code, analysis inputs, and repeated draws for reproducing the principal results.

\section{Reproducibility Checklist}
\label{app:group_h}

\subsection{Reproducibility Checklist}
\label{app:repro_checklist}

The following index summarizes the empirical reporting components of the study.

\paragraph{Claims and contributions.} Sections~\ref{sec:benchmark}--\ref{sec:analysis} connect the five information levels to measured domain-profile error. Limitations discusses the task scope and the use of simulated feedback.

\paragraph{Data and code.} The empirical material comprises 45,000 primary interaction records, per-model session summaries, baseline implementations, self-assessment replies, and the extension and sensitivity analyses. The scoring scripts reuse the stored answer and feedback streams. Model request configurations appear in Appendix~\ref{app:api_versions}.

\paragraph{Experimental protocol.} Main-study CBF reconstructs observed session accuracy. Held-out prediction uses a separate train/test split in the ReAct analysis. Appendix~\ref{app:baselines} specifies estimator updates; Appendix~\ref{app:stat_rob} reports resampling and paired contrasts. Noise and partition experiments retain the results of each repetition.

\paragraph{Repeatability.} Stored traces permit deterministic comparisons of estimators on the same observations. New LLM calls can vary with sampling and service versions. Simulator comparisons and held-out predictions are reported as separate experiments in Appendices~\ref{app:dual_role} and~\ref{app:react_prediction}.

\paragraph{Ethics and broader impact.} The primary interaction study uses simulated personas. Human-feedback evaluation is a separate study requiring its own documented procedures. The benchmark measures domain-estimation performance and does not certify safe deployment.

\section{Worked Examples and Prompt Templates}
\label{app:group_i}

This appendix gives a worked tour of the protocol so that readers can verify the data shape and reproduce the LLM calls. \S\ref{app:sample_sessions} shows two real session excerpts (one mid-tier model, one frontier model). \S\ref{app:failure_modes} catalogs the most common decoder and self-knowledge failure modes encountered in the released traces. \S\ref{app:prompts} provides the answer/simulator templates, the feedback-rule specification, and current self-assessment prompts.

\subsection{Sample session trajectories}
\label{app:sample_sessions}

We show five turns from each of two recorded sessions. Each excerpt contains the agent's answer, reported confidence, verified outcome, and simulated-user feedback. In-expertise and out-of-expertise denote the assigned user persona's relationship to the subject.

\paragraph{Example 1: Qwen-Plus, biology persona.} Persona expertise: college biology, high school biology, and anatomy; style \texttt{academic}, tolerance $0.2$, explicitness $0.9$.

\begin{itemize}[nosep,leftmargin=1.4em,labelsep=0.4em]
\item \textbf{T1 high school biology} (in-expertise). \emph{Q:} ``During which phase of the cell cycle does the quantity of DNA in a eukaryotic cell typically double?'' \emph{Agent:} D, conf $1.00$ ($\checkmark$). \emph{User:} ``That's correct, the quantity of DNA in a eukaryotic cell doubles during the S phase of the cell cycle.''
\item \textbf{T2 high school chemistry} (out). \emph{Q:} organic-acid functional group. \emph{Agent:} C, conf $1.00$ ($\checkmark$). \emph{User:} ``Could you elaborate on that answer? I'm familiar with the concept but I'd like to be sure I understand which specific functional group you're referring to.''
\item \textbf{T3 abstract algebra} (out). \emph{Q:} integral domain $R$ implies $R[x]$ integral domain. \emph{Agent:} C, conf $1.00$ ($\checkmark$). \emph{User:} ``The answer C seems reasonable, but could you elaborate on how both statements are related and why that leads to the conclusion?''
\item \textbf{T4 global facts} (out). \emph{Q:} \% of Americans valuing free media. \emph{Agent:} C, conf $0.85$ (\textbf{wrong}). \emph{User:} ``I see, thank you for the information. It's reassuring to know that a significant portion of Americans value the importance of a free press.''
\item \textbf{T5 high school chemistry} (out). \emph{Q:} pH of a weak monoprotic acid. \emph{Agent:} \texttt{None}, conf $0.50$ (excluded: no parsed answer). \emph{User:} ``Thank you for the information. I'll proceed with the assumption that the pKa value provided is correct\dots''
\end{itemize}

T4 contains a wrong answer followed by a polite acknowledgement, illustrating why conversational agreement need not be a correctness label. T5 has an unparsed answer and is excluded from the valid-answer accuracy target.

\paragraph{Example 2: GPT-5.4, biology persona.} Same persona as Example~1; agent is a frontier model.

\begin{itemize}[nosep,leftmargin=1.4em,labelsep=0.4em]
\item \textbf{T1 clinical knowledge} (out). \emph{Q:} ulnar nerve palsy. \emph{Agent:} C, conf $0.96$ ($\checkmark$). \emph{User:} ``That makes sense---could you elaborate on what specifically happens in ulnar nerve palsy that makes C the correct answer?''
\item \textbf{T2 clinical knowledge} (out). \emph{Q:} chain-of-survival ordering. \emph{Agent:} B, conf $0.99$ ($\checkmark$). \emph{User:} ``That makes sense---could you walk me through why that particular ordering is considered optimal\dots?''
\item \textbf{T3 conceptual physics} (out). \emph{Q:} Doppler effect. \emph{Agent:} C, conf $0.99$ ($\checkmark$). \emph{User:} ``That makes sense, though I'm curious what the full answer choice C said\dots''
\item \textbf{T4 high school chemistry} (out). \emph{Q:} para-dichlorobenzene IUPAC name. \emph{Agent:} D, conf $0.99$ ($\checkmark$). \emph{User:} ``That makes sense---could you walk me through the IUPAC naming logic\dots?''
\item \textbf{T5 astronomy} (out). \emph{Q:} when the next major impact will occur. \emph{Agent:} D, conf $0.94$ ($\checkmark$). \emph{User:} ``That's an interesting answer, though I'd love to hear more about the reasoning behind it\dots''
\end{itemize}

GPT-5.4 reports confidence of at least .94 on all five displayed turns. Several user replies begin with ``That makes sense'', but explicit agreement, follow-up questions, and correctness judgments play different roles in the feedback stream. Only utterances satisfying the decoder's positive/negative rules supply algorithmic labels.

\subsection{Failure mode catalog}
\label{app:failure_modes}

The examples illustrate several ways in which confidence, feedback, and domain estimates can diverge.

\begin{enumerate}[nosep,leftmargin=1.4em]
\item \textbf{Acknowledgment without verification.} Polite agreement need not certify answer correctness.
\item \textbf{Saturated confidence.} Little variation in reported confidence limits its usefulness for distinguishing outcomes.
\item \textbf{Unparsed answers.} Missing answer options are excluded from accuracy and retained in the attempt count.
\item \textbf{Ambiguous feedback.} Follow-up questions can leave a domain with few decoded labels.
\item \textbf{Prior--history disagreement.} A history-based estimate can be worse than a zero-shot profile, as on Qwen-Plus in Table~\ref{tab:main}.
\end{enumerate}

\subsection{Prompt templates}
\label{app:prompts}

The answer and simulator templates illustrate the interaction format. The self-assessment prompts below specify the inputs used in the complete-profile collection.

\paragraph{Agent answer prompt.} Used at every turn to elicit an answer and a confidence value.

\begin{small}
\begin{Verbatim}[fontsize=\small,breaklines=true,breakanywhere=true]
Answer the question and give your confidence 0-100%.

Question: {question}
Choices:
{choices_one_per_line}

Format your response exactly as:
Answer: [A/B/C/D]
Confidence: [0-100]
\end{Verbatim}
\end{small}

\paragraph{Simulated user prompt.} The persona receives the current question, the agent's answer, and a sampled stage direction.

\begin{small}
\begin{Verbatim}[fontsize=\small,breaklines=true,breakanywhere=true]
Your previous question: {question}

The assistant's answer: {agent_answer}

[Stage direction - do NOT quote]
{scenario}

Now write your natural next response (1-2 sentences):
\end{Verbatim}
\end{small}

\paragraph{Feedback decoding.} The primary rule classifier checks negative patterns before uncertainty and positive patterns; unmatched utterances are neutral. API error strings are missing feedback. Only positive and negative labels enter the estimators. The complete implementation is distributed with the reproducibility scripts. These labels measure rule-decoded stance, not human-annotated semantic correctness.

\paragraph{L1 JSON response format.}

\begin{small}\begin{Verbatim}[fontsize=\small,breaklines=true,breakanywhere=true]
Without any interaction history or task outcomes, estimate your expected multiple-choice accuracy in each of these MMLU subjects. Return only a JSON object mapping each listed domain to a probability in [0,1]. Domains: {domains}
\end{Verbatim}
\end{small}

\paragraph{L2 LSA and L4 supplied-statistic prompts.} Both return one domain profile at the session endpoint.

\begin{small}\begin{Verbatim}[fontsize=\small,breaklines=true,breakanywhere=true]
Estimate what fraction of your answers was correct in each domain of this session, using the following question and user-feedback history. Return only one JSON object mapping each listed domain to a probability in [0,1].
{history}
\end{Verbatim}
\end{small}

\begin{small}\begin{Verbatim}[fontsize=\small,breaklines=true,breakanywhere=true]
Report the following supplied per-domain accuracies exactly. Return only one JSON object mapping the same domain names to probabilities in [0,1].
{statistics}
\end{Verbatim}
\end{small}

\end{document}